\documentclass[10pt,twocolumn,letterpaper]{article}

\usepackage[pagenumbers]{arxiv}      
\definecolor{arxivblue}{rgb}{0.21,0.49,0.74}
\usepackage[pagebackref,breaklinks,colorlinks,allcolors=arxivblue]{hyperref}
\usepackage{siunitx}
\usepackage{xcolor}
\usepackage[table]{xcolor}
\usepackage{tcolorbox}
\usepackage{soul}
\usepackage{makecell}
\usepackage[misc]{ifsym}
\usepackage{enumitem}
\usepackage{amsmath}
\usepackage{amsthm}
\newtheorem{definition}{Definition}
\usepackage{longtable}
\usepackage{caption}
\usepackage{booktabs}
\usepackage{multirow}
\usepackage{titletoc}
\usepackage{tabularx}
\usepackage{iftex} % 用于检测编译环境

\ifdefined\latexml
    \newenvironment{strip}{\par\vspace{1mm}}{\par\vspace{1mm}}
\else
    \usepackage{cuted}
\fi
\usepackage{setspace}
\usepackage{tocloft}
\def\paperID{12558} % *** Enter the Paper ID here
\def\confName{Arxiv}
\def\confYear{2026}

\newcommand{\best}[1]{\cellcolor[HTML]{FF9898}{#1}}
\newcommand{\second}[1]{\cellcolor[HTML]{FFCB98}{#1}}
\newcommand{\third}[1]{\cellcolor[HTML]{FFFF98}{#1}}

\definecolor{Red}{HTML}{FF9898}
\definecolor{Orange}{HTML}{FFCB98}
\definecolor{Yellow}{HTML}{FFFF98}
\definecolor{PurpleBox}{HTML}{7030A0}
\definecolor{YellowBox}{HTML}{FFC000}
\definecolor{GreenBox}{HTML}{00B050}

\title{XYZCylinder: Towards Compatible Feed-Forward 3D Gaussian Splatting for Driving Scenes via Unified Cylinder Lifting Method}

\author{
Haochen Yu, Qiankun Liu \textsuperscript{\Letter}, Hongyuan Liu, Jianfei Jiang, \\ Juntao Lyu, Jiansheng Chen, Huimin Ma \textsuperscript{\Letter} \\
University of Science and Technology Beijing\\
{\tt\small \{haochen.yu,hongyuanliu,jiangjf,lyujuntao\}@xs.ustb.edu.cn} \\
{\tt\small \{liuqk3,jschen,mhmpub\}@ustb.edu.cn }
}

\begin{document}
\maketitle
% 用于三维重建的前馈模式已成为近期研究的焦点，这类模式学习隐式且固定的视角变换以生成单一的场景表示。然而，将其应用于复杂的驾驶场景时却暴露出显著的局限性。造成这种性能差距的两个核心问题是导致这一差距的原因。首先，对固定视角变换的依赖阻碍了对不同相机配置的适应性，严重限制了兼容性。其次，从稀疏的 360° 视角中学习复杂的驾驶场景并保持最小重叠的固有难度损害了最终的重建精度。为了解决这些难题，我们引入了 \textbf{XYZCylinder}，这是一种基于统一圆柱体提升方法的新方法，该方法将相机建模和特征提升整合在一起。为解决兼容性问题，我们设计了统一圆柱体相机建模（UCCM）策略。该策略明确地对投影参数进行建模，以统一各种相机设置，从而避免了学习与视角相关对应关系的需求。为了提高重建的准确性，我们提出了一种基于新设计的圆柱面特征组（CPFG）的混合表示方法，该方法包含多个专用模块，用于将二维图像特征提升到三维空间。大量的评估结果证实，XYZCylinder 不仅在不同的评估设置下达到了最先进的性能，而且在全新的场景中，无论相机设置如何变化，都能以零样本的方式展现出出色的兼容性。
\begin{abstract}
Feed-forward paradigms for 3D reconstruction have become a focus of recent research, which learn implicit, fixed view transformations to generate a single scene representation. However, their application to complex driving scenes reveals significant limitations. Two core challenges are responsible for this performance gap. First, the reliance on a fixed view transformation hinders compatibility to varying camera configurations. Second, the inherent difficulty of learning complex driving scenes from sparse 360° views with minimal overlap compromises the final reconstruction fidelity. To handle these difficulties, we introduce \textbf{XYZCylinder}, a novel method built upon a unified cylinder lifting method that integrates camera modeling and feature lifting. To tackle the compatibility problem, we design a Unified Cylinder Camera Modeling (UCCM) strategy. This strategy explicitly models projection parameters to unify diverse camera setups, thus bypassing the need for learning viewpoint-dependent correspondences. To improve the reconstruction accuracy, we propose a hybrid representation with several dedicated modules based on newly designed Cylinder Plane Feature Group (CPFG) to lift 2D image features to 3D space. Extensive evaluations confirm that XYZCylinder not only achieves state-of-the-art performance under different evaluation settings but also demonstrates remarkable compatibility in entirely new scenes with different camera settings in a zero-shot manner. Project page: \href{https://yuyuyu223.github.io/XYZCYlinder-projectpage/}{here}
\end{abstract}    
\section{Introduction}
\label{sec:intro}
% 3D 重建技术旨在通过有限的 2D 图像视角构建具有空间结构和视觉真实感的 3D 数字模型，这在计算机图形学和计算机视觉领域一直是一个热门话题，并且已被广泛应用于诸多任务中，比如自动驾驶。在本文中，我们专注于利用某一时间点的稀疏视角来重建驾驶场景。
3D reconstruction focuses on building a 3D digital model with spatial structure and visual fidelity from the limited views of 2D images, which has been a hot topic in computer graphics and computer vision, and has been widely used in many tasks, for example, autonomous driving. In this paper, we focus on the reconstruction of driving scenes using the sparse views of one timestamp.

% 以往的迭代重建方法虽然具有较高的精度，但对大规模 3D 资产采集而言计算成本过高。通常，这些方法 \citep{magicdrive3d,reconx，viewcrafter,splatfields} 是基于神经辐射场（NeRF）\citep{nerf} 或 3D 均值片绘（3DGS）\citep{3DGS} 的表示来重建场景的，需要针对不同的场景进行迭代优化。优化过程本身的高延迟和高计算成本阻碍了迭代重建方法在效率密集型任务中的应用。相比之下，前向式重建方法 \citep{drivingforward， Omniscene， 6imgto3d} 能够在单次前向传递中重建场景，并能推广到不同的场景，因此吸引了更多研究人员的关注。
Previous iterative reconstruction methods, while capable of high accuracy, are too computationally expensive for large-scale 3D asset collection. Generally, these methods \citep{magicdrive3d,reconx,viewcrafter,splatfields} reconstruct the scene based on Neural Radiance Fields (NeRF) \citep{nerf} or 3D Gaussian Splatting (3DGS) \citep{3DGS} representations, which need to be iteratively optimized for different scenes. The inherent high latency and computational cost of the optimization process hinder iterative reconstruction methods from being applicable to efficiency-intensive tasks. In contrast, feed-forward reconstruction methods \citep{drivingforward, Omniscene, 6imgto3d} 
reconstruct the scenes within a single forward pass and generalize to different scenes, making them attract more attention from researchers. 

\begin{figure}
    \centering
    \includegraphics[width=\linewidth]{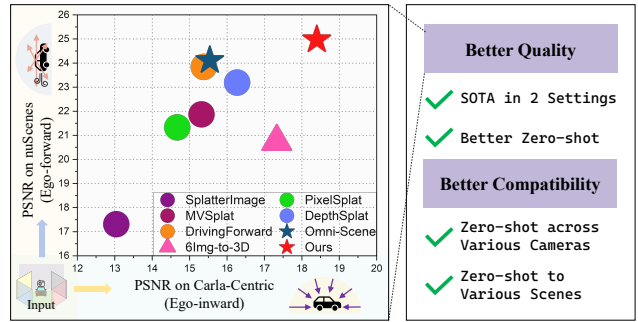}
    \caption{Reconstruction results of the proposed method under different evaluation settings. Our model achieves better reconstruction quality and better compatibility.}
     \label{psnr}
     \vspace{-15pt}
\end{figure}

% 现有的前馈重建方法 \citep{Flare, pixelsplat, mvsplat, depthsplat, MonoSplat, efreesplat, freesplat, dust3r, Mvdust3r, driv3r} 主要是在隐式方式下学习固定的视图变换，并使用单一的表示来重建场景。在重建驾驶场景时，它们的兼容性和重建精度受到限制。例如，不同的汽车可能具有不同的相机配置，包括相机的数量以及相机的外在和内在参数。视图依赖设计学习固定的视图变换，限制了它们对不同驾驶场景的泛化能力。在遇到各种相机配置时，它们会遭受严重的质量下降，需要对网络进行彻底的重新设计和重新训练。此外，它们的重建能力依赖于从不同视图之间高度重叠区域中学习到的基于像素的空间对应关系，当重叠区域变小时（例如，在常见的驾驶场景中，对于 360 度全景图，大约需要 6 个视图），这种对应关系可能会失效。此外，这些方法基于每个像素的深度和偏移量生成了“2.5D”场景表示，但在处理复杂的驾驶场景时会因尺度模糊和不完整几何结构等问题而出现变形、空洞和失真现象。尽管一些方法 \citep{Omniscene，6imgto3d} 通过紧凑的跨视图 3D 表示（例如 Triplane \citep{eg3d} 和 Tri-Perspective View (TPV) \citep{TPVFormer}）来重建驾驶场景，但它们在大角度视图变换和稀疏几何线索的情况下会面临性能下降的问题。
Existing feed-forward reconstruction methods \citep{Flare, pixelsplat, mvsplat, depthsplat, MonoSplat, efreesplat, freesplat, dust3r, Mvdust3r, driv3r} mainly learn a fixed view transformation implicitly and reconstruct the scene with a single representation. Their compatibility and reconstruction accuracy are limited when reconstructing driving scenes. For example, different cars may have different camera configurations, including the number of cameras and the extrinsic and intrinsic parameters of the cameras. The view-dependent design learns the fixed view transformation, limiting its generalization for different driving scenes. When meeting various camera configurations, they suffer from severe quality degradation, requiring a complete redesign and retraining of the networks. Besides, their reconstruction capability relies on the pixel-based spatial correspondence learned from the highly overlapping regions between different views, which may fail when the overlapping regions become small (\eg, $\sim$6 views for \SI{360}{\degree} panorama in common driving scenes). In addition, these methods produce the ``2.5D'' scene representation based on per-pixel depths and offsets, which suffers from the issues of deformation, holes, and distortion due to the scale ambiguity and incomplete geometry when meeting complex driving scenes. Though some methods \citep{Omniscene,6imgto3d} reconstruct driving scenes with compact cross-view 3D representations (\ie, Triplane \citep{eg3d} and Tri-Perspective View (TPV) \citep{TPVFormer}), they suffer from performance degradation with large-angle view transformation and sparse geometric cues.

% 为解决这些问题，我们提出了 \textbf{XYZCylinder} 这种基于统一圆柱体提升方法的前馈重建技术，用于驱动场景的生成。该方法包含相机建模策略和特征组。 （1）为了提高兼容性，我们设计了一种统一圆柱体相机建模策略（\Cref{feature_extraction}），其中基于学习的视图变换被替换为确定性和明确的映射，省去了空间对应关系的学习过程。通过在圆柱体平面的构建过程中调整无需训练的参数，所提出的方法能够在不同相机配置下实现零样本重建，展示了其兼容性。 （2）为了提高重建精度，我们将场景重建表述为由新设计的圆柱体平面特征组（CPFG）解码的混合表示形式。借助专用的占用感知、体积感知和像素感知模块，CPFG 将特征从二维空间提升到三维空间（\Cref{foreground_reconstruction}）。还采用了前景-背景解耦策略（有限深度和无限深度），以更好地进行建模（\Cref{background_reconstruction}），从而获得无漂移伪影的清晰场景。为了评估 XYZCylinder，采用了自车向前 \cite{Omniscene} 和自车向内 \citep{6imgto3d} 的设置，如 \Cref{psnr} 所示。大量的实验结果表明，XYZCylinder 在不同的评估设置下都取得了最先进的重建结果，并且能够以零样本的方式很好地推广到具有不同相机配置的其他驾驶场景中。
To handle these issues, we propose \textbf{XYZCylinder}, a feed-forward reconstruction method for driving scenes based on a unified cylinder lifting method. This method includes a camera modeling strategy and feature groups. (1) To improve the compatibility, we design a Unified Cylinder Camera Modeling (UCCM) strategy (\Cref{feature_extraction}), where the learning-based view transformation is replaced by a deterministic and explicit mapping, omitting the learning of spatial correspondence. By adjusting the training-free parameters during the construction of the cylinder planes, the proposed method can achieve zero-shot reconstruction with different camera configurations, showing its compatibility. (2) To improve the reconstruction accuracy, we formulate the scene reconstruction as a hybrid representation decoded by the newly designed Cylinder Plane Feature Group (CPFG). With the dedicated occupancy-aware, volume-aware, and pixel-aware modules, CPFG lifts the features from 2D space to 3D space (\Cref{foreground_reconstruction}). A foreground-background decoupling strategy (finite and infinite depth) is also adopted for better modeling (\Cref{background_reconstruction}) to obtain clean scenes without floating artifacts.
To evaluate XYZCylinder, both ego-forward \cite{Omniscene} and ego-inward \citep{6imgto3d} settings are adopted, as shown in \Cref{psnr}. Extensive experimental results show that XYZCylinder achieves SOTA reconstruction results across different evaluation settings, and can be generalized well to other driving scenes with different camera configurations in a zero-shot manner. The main contributions in this paper include:
\begin{itemize}[leftmargin=*, topsep=0pt, partopsep=0pt, parsep=0pt, itemsep=0pt, labelsep=0.5em]

% 我们设计了一种统一的圆柱摄像机建模策略，将不同的视角映射到一个具有可调节参数的统一圆柱平面上，从而提高了 XYZCylinder 在不同摄像机设置下的兼容性。
\item We design a unified cylinder camera modeling strategy to map different views to a unified cylinder plane with adjustable parameters, improving the compatibility of XYZCylinder across different camera settings.

% 我们采用一种混合表示方法来模拟驾驶场景，该方法基于圆柱面特征组构建了专用模块，从而提高了 XYZCylinder 的重建精度。
\item We model the driving scene using a hybrid representation with dedicated modules based on the cylinder plane feature group, enhancing the reconstruction accuracy of XYZCylinder. 

% 所提出的 XYZ 立体相机在不同的评估条件下均能实现最先进的重建效果，并且还能以零样本的方式应用于具有不同相机设置的场景中。
\item The proposed XYZCylinder achieves state-of-the-art reconstruction results under different evaluation settings and can also be generalized to scenes with different camera settings in a zero-shot manner.  

\end{itemize}
\section{Related Works}
\label{sec:relatedwork}
\noindent
\textbf{Feedforward Reconstruction Models.} 
% 前向重建方法是由其场景表示所定义的。体积方法 \citep{6imgto3d,scube,TGS} 能确保几何完整性并有效地处理遮挡问题，但它们往往生成的表面过于平滑，缺乏高频细节。相反，基于像素的模型 \citep{dust3r,Mvdust3r,VGGT,pixelsplat,mvsplat，depthsplat，Flare，driv3r,evolsplat,mvsgaussian,nexusgs,pixelgaussian,freesplat,freesplat++，MonoSplat,gssplat,smilesplat,efreesplat,splatt3r} 在前向场景中擅长生成密集的几何结构，但在面向外部的场景（如自动驾驶）中却存在空洞和遮挡问题。这种权衡促使了诸如 Omni-Scene 这样的混合方法的出现 \citep{Omniscene}。我们的工作在此范式基础上取得了进展，与 Omni-Scene 的基于像素的策略不同，我们通过为每种表示使用四个预测分支来实现区分，从而实现了更完整和更详细的几何重建。
Feedforward reconstruction methods are defined by their scene representation. Volumetric methods \citep{6imgto3d,scube,TGS} ensure geometric completeness and handle occlusions effectively, yet they often produce overly smooth surfaces lacking high-frequency detail. Conversely, pixel-based models \citep{dust3r,Mvdust3r,VGGT,pixelsplat,mvsplat,depthsplat,Flare,driv3r,evolsplat,mvsgaussian,nexusgs,pixelgaussian,freesplat,freesplat++,MonoSplat,gssplat,smilesplat,efreesplat,splatt3r} excel at generating dense geometry in forward-facing scenarios but suffer from voids and occlusions in outward-facing settings like autonomous driving. This trade-off has motivated hybrid methods like Omni-Scene \citep{Omniscene}. Our work advances this paradigm, distinguishing itself from Omni-Scene's pixel-guided strategy by employing four prediction branches for each representation, which yields a more complete and detailed geometry.

\noindent
\textbf{2D Feature Lifting.} 
% 将二维特征提升至三维表示形式是三维感知的基础。CaDDN~\citep{CaDDN} 利用预测的类别深度分布将二维特征投影到三维体积中。LSS~\citep{LSS} 通过使用预测的每像素深度分布来投影视锥特征，从而形成鸟瞰视图（BEV）地图。BEVFormer~\citep{bevformer} 利用空间交叉注意力将多摄像头的二维特征转换为统一的 BEV 表示。TPVFormer~\citep{TPVFormer} 使用变压器从二维输入构建三个相互垂直的平面。DFA3D~\citep{dfa3d} 将深度与三维可变形注意力机制相结合，将二维特征提升至三维形式。相比之下，我们的方法利用特征空间一致性和通道重排序，将二维特征重构为基于支柱的场，以供后续的三维任务使用。相关工作的更详细综述见补充材料。
Lifting 2D features to a 3D representation is fundamental to 3D perception. CaDDN~\citep{CaDDN} projects 2D features into a 3D volume using a predicted categorical depth distribution. LSS~\citep{LSS} forms a Bird's-Eye-View (BEV) map by projecting view frustum features with a predicted per-pixel depth distribution. BEVFormer~\citep{bevformer} employs spatial cross-attention to convert multi-camera 2D features into a unified BEV representation. TPVFormer~\citep{TPVFormer} uses a transformer to construct three orthogonal planes from 2D inputs. DFA3D~\citep{dfa3d} integrates depth with a 3D deformable attention mechanism to lift 2D features. In contrast, our approach leverages feature spatial consistency and channel reordering to restructure 2D features into a pillar-based field for subsequent 3D tasks. A more detailed review of related work is provided in supplementary materials.
\section{Methodology}
\label{Methodology}
\begin{figure*}[t]
	\centering
	\begin{minipage}{\linewidth}
		\centering
    	\includegraphics[width=\linewidth]{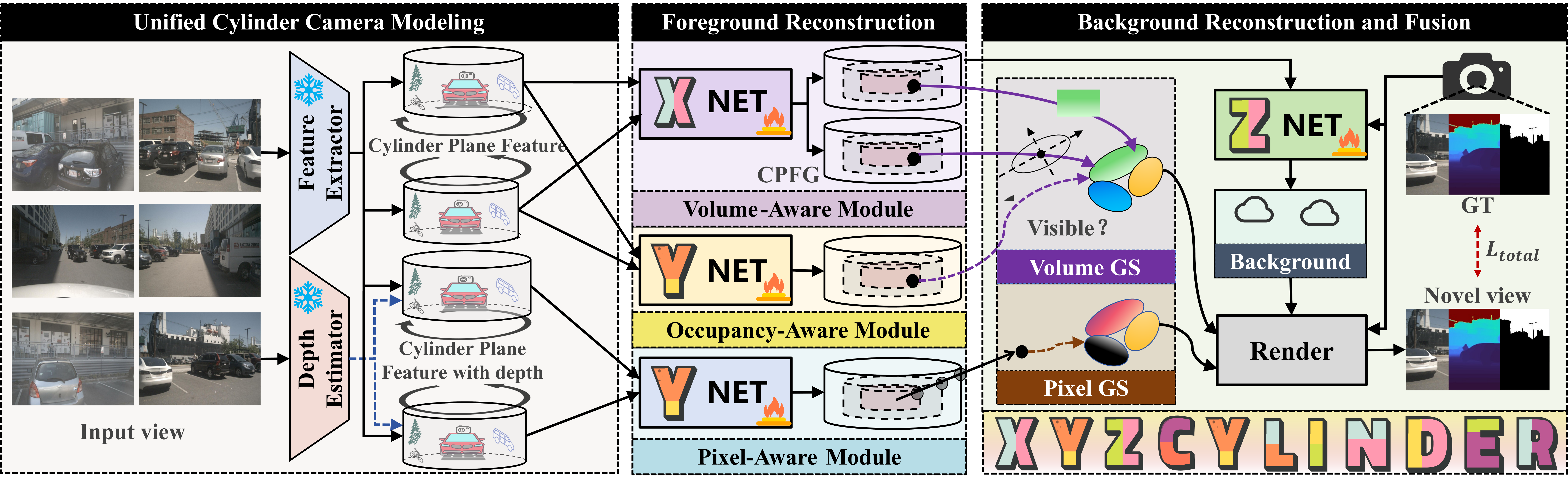}
     % 我们的模型使用六视角作为输入，以新视角GT作为监督 % 这是一种全新的稀疏视角驾驶重建模型
     \caption{\textbf{Overview of XYZCylinder.} The scene is reconstructed in three stages with the unified cylinder camera modeling for feature extraction and a hybrid representation with different dedicated modules for foreground and background reconstruction.}
    	\label{main_pipe}
	\end{minipage}
 \vspace{-10pt}
\end{figure*}

\begin{figure*}[htb]
	\centering
	\includegraphics[width=\linewidth]{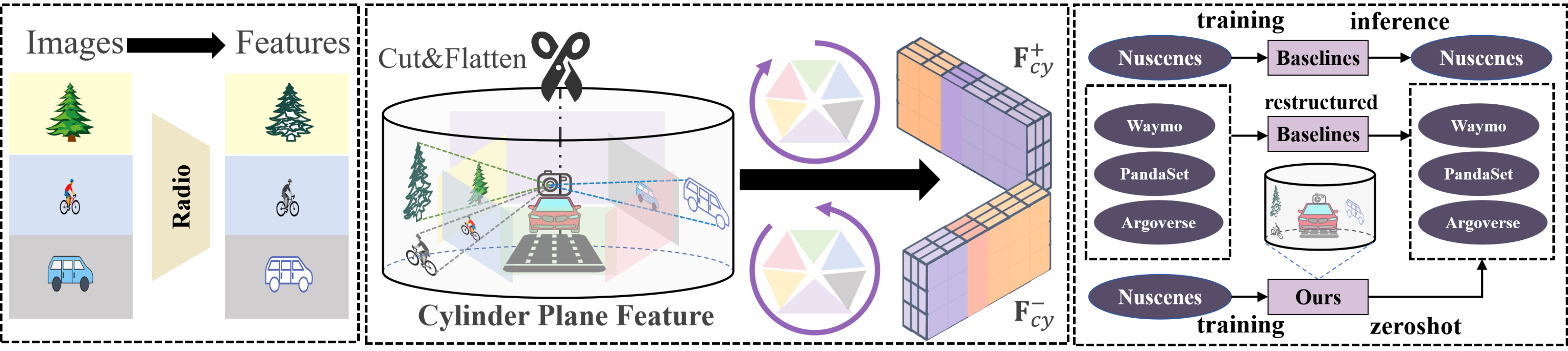}
     % Unified Cylinder Plane的构建方法，包含图像的特征提取、异向的投影
     % 得益于UCP的设计，使得我们的模型具有很强的零样本泛化能力，统一了几乎常用自动驾驶数据集。
	\caption{\textbf{Overview of the unified cylinder camera modeling (UCCM).} The design of UCCM empowers our model with zero-shot generalization across different datasets.}
	\label{UCP}
 \vspace{-12pt}
\end{figure*}

% 在某一时间点内，如果有 $N$ 个视角（例如，在常见的驾驶场景中，$N=6$），则目标是重建相应的三维场景。具体而言，第 $n$ 个视角用元组 $\mathcal{C}_n = \{\mathbf{I}_n， f_n， \mathbf{P}_n， \mathbf{W}_n^e， \mathbf{c}_n^{e} \}$ 来表示，其中 $\mathbf{I}_n \in \mathbb{R}^{3\times H\times W}$ 是以垂直视场 $f_n$ 拍摄的图像，而 $\mathbf{P}_n \in \mathbb{R}^{4\times 4}$、$\mathbf{W}_n^e \in \mathbb{R}^{4\times 4}$ 和 $\mathbf{c}_n^{e} \in \mathbb{R}^{3}$ 分别是相机的内参、从相机到自车坐标系的外参以及相机在自车坐标系中的位置。如 \Cref{main_pipe} 所示，重建流程分为三个阶段：（1）在特征提取阶段（\Cref{feature_extraction}），设计了一个统一圆柱相机建模（UCCM）来将不同的视角投影到一个统一的圆柱平面上；（2）在前景重建阶段（见\Cref{foreground_reconstruction}），构建感知模块、体积感知模块和像素感知模块，以生成圆柱面特征组（CPFG）；（3）在背景重建与融合阶段（见\Cref{background_reconstruction}），在二维空间中生成背景（即天空和云层），并与渲染的二维前景（即汽车、道路和建筑物）图像进行融合。
Given $N$ views (\eg, $N=6$ in common driving scenes) in one timestamp, the goal is to reconstruct the corresponding 3D scene. Formally, the $n$-th view is denoted by a tuple $\mathcal{C}_n = \{\mathbf{I}_n, f_n, \mathbf{P}_n, \mathbf{W}_n^e, \mathbf{c}_n^{e} \}$, where $\mathbf{I}_n \in \mathbb{R}^{3\times H\times W}$ is the image captured with the vertical Field of View (FoV) $f_n$, and $\mathbf{P}_n \in \mathbb{R}^{4\times 4}$, $\mathbf{W}_n^e \in \mathbb{R}^{4\times 4}$, and $\mathbf{c}_n^{e} \in \mathbb{R}^{3}$ are the intrinsic parameters of the camera, extrinsic parameters mapping from the camera to the ego coordinate system, and the position of the camera in the ego-vehicle coordinate system, respectively.
As shown in \Cref{main_pipe}, the reconstruction pipeline is divided into three stages: (1) In the feature extraction stage (\Cref{feature_extraction}), a Unified Cylinder Camera Modeling (UCCM) is designed to project different views into a unified cylinder plane; (2) In the foreground reconstruction stage (\Cref{foreground_reconstruction}), the Occupancy-Aware Module, Volume-Aware Module, and Pixel-Aware Module are designed to produce Cylinder Plane Feature Group (CPFG); (3) In the background reconstruction and fusion stage (\Cref{background_reconstruction}), the background (\ie, sky and clouds) is generated in the 2D space and fused with the rendered 2D foreground (\ie, cars, roads and buildings) image.

\begin{figure*}[t]
	\centering
	\includegraphics[width=0.9\linewidth]{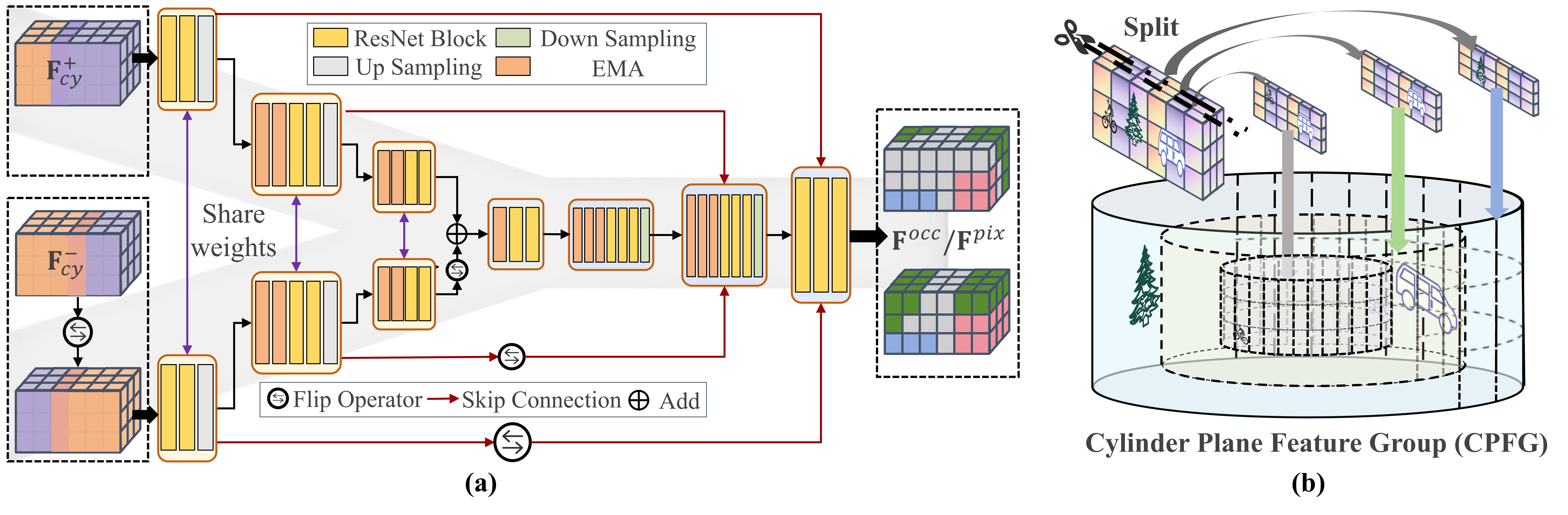}
         % （1）左侧：用于感知占用情况的 Y 形网络架构的 YNet$_{\rm occ}$ 模块和用于像素感知的 YNet$_{\rm pix}$ 模块。该网络主要基于 ResNet 块 \citep{he2016deep} 和 EMA \citep{EMAttention} 实现。 （2）右侧：圆柱面特征组是通过在通道维度上分割特征构建而成的。
		\caption{(a) \textbf{Architecture of Y-shaped network} for the occupancy-aware module YNet$_{\rm occ}$ and pixel-aware module YNet$_{\rm pix}$. The network is mainly implemented based on the ResNet Block \citep{he2016deep} and EMA \citep{EMAttention}. (b) The cylinder plane feature group is constructed by splitting the feature in the channel dimension.}
		\label{Ynet}
 \vspace{-12pt}
\end{figure*}

\subsection{Feature Extraction} 
\label{feature_extraction}
% 处理由 $N$ 个环绕式摄像机所获取的 360° 视角并非易事。为此，我们设计了一种统一圆柱摄像机建模（UCCM）策略，以便将这 $N$ 个视角有效地投影到一个圆柱面上，如图 \Cref{UCP} 所示。UCCM 使用五个关键参数来定义一个圆柱坐标系：中心点的坐标 $\mathbf{c}_{u}^x \in \mathbb{R}^3$、半径 $R_u \in \mathbb{R}$、高度 $Z_u \in \mathbb{R}$ 以及圆柱面的分辨率 $H_u \times W_u$。\textbf{}
Processing the 360° field of view from $N$-surrounded cameras is nontrivial. To do this, we design a Unified Cylinder Camera Modeling (UCCM) strategy to efficiently project these $N$ views into a cylinder plane as shown in \Cref{UCP}. UCCM defines a cylindrical coordinate system using five key parameters: the coordinates of the central point $\mathbf{c}_{u}^x \in \mathbb{R}^3$, radius $R_u \in \mathbb{R}$, height $Z_u \in \mathbb{R}$, and resolution $H_u \times W_u$ of the cylinder plane.

\noindent
\textbf{Construction of Cylinder Plane.}
% 中心点 $\mathbf{c}_{u}^x$ 是通过将 $N$ 个摄像机的三维位置进行平均计算得出的，其计算公式为 $\mathbf{c}_{u}^x = \sum_{n=0}^{N-1}\mathbf{c}_n^e / N$。而圆柱体的高度 $Z_u$ 则由场景的垂直范围决定。具体而言，对于一个三维化的占用网格 $\mathbf{O} \in \mathbb{R}^{L_o\times H_o\times W_o}$，我们首先提取其对应的点云 $\mathbf{X}_o \in \mathbb{R}^{3 \times L_o\times H_o\times W_o}$。高度 $Z_u$ 是沿垂直轴计算的最顶部和最底部点之间的距离。然后，半径的计算公式为：$R_u = \frac{Z_u / 2}{\tan{(\rho \cdot f_{min} / 2)}}$ 。
The central point $\mathbf{c}_{u}^x$ is computed by averaging the 3D positions of the $N$ cameras using $\mathbf{c}_{u}^x=\sum_{n=0}^{N-1}\mathbf{c}_n^e / N$, while the cylinder's height $Z_u$ is derived from the vertical extent of the scene. Specifically, given a voxelized occupancy grid $\mathbf{O} \in \mathbb{R}^{L_o\times H_o\times W_o}$, we first extract its corresponding point cloud $\mathbf{X}_o\in \mathbb{R}^{3 \times L_o\times H_o\times W_o}$. The height $Z_u$ is the distance between the top-most and bottom-most points along the vertical axis.
%  在理想情况下，如果摄像机位置集中且参数共享，原始视场角将与圆柱投影所需的垂直角度完全匹配。但在实际中，摄像机与 UCP 中心轴之间的空间偏移会导致不匹配，从而使圆柱体的顶部和底部区域无法投影，并产生边界伪影。为了缓解这一问题，我们调整了圆柱体的半径 $R_u$。有效垂直角度 $F_u \in \mathbb{R}$ 被设定为摄像机的原始视场角 $F_{base} \in \mathbb{R}$ 的最小值乘以一个小于 1 的系数 $\rho$，如 \Cref{eq-5} 中所表述的那样，从而减少了未投影的区域。
% In an ideal scenario where the $N$ cameras share the same location and intrinsic parameters, and only differ in the poses, the FoV would perfectly match the vertical angle required for cylindrical projection. Unfortunately, there exist inevitable camera-to-camera and camera-to-central axis offsets, leaving the top and bottom regions in the cylinder plane unprojected and introducing boundary artifacts. To mitigate this issue, the effective vertical angle is set to the minimum FoV $f_{min} \in \mathbb{R}$ of these $N$ views, which is further multiplied by a factor of $\rho$, and a small height offset $(0, 0, \Delta h)^T$ is applied to $\mathbf{c}_{u}^x$, thereby bmitigating the unprojected regions. 
Then, the radius is calculated by $R_u=\frac{Z_u / 2}{\tan{(\rho \cdot f_{min}/2)}}$. 

\noindent
\textbf{Production of Cylinder Plane Feature.}
% 根据第 n 个视图，图像 $\mathbf{I}_n$ 被输入到特征提取器（即 Radio-v2.5 \citep{Radio}）中，以生成图像特征 $\mathbf{F}_n$。接下来，我们需要基于 $\mathbf{F}_n$ 获取圆柱面的第 n 个视图特征。为此，我们将圆柱面上所有离散点的圆柱坐标投影到笛卡尔坐标系中，然后使用第 n 个相机的中心点坐标 $\mathbf{c}_u^x$、外在参数 $\mathbf{W}_n^e$ 和内在参数 $\mathbf{P}_n$ 将其进一步投影到 2D 相机像素坐标系中。投影到特征 $\mathbf{F}_n$ 之外的点将被舍弃，仅保留位于特征 $\mathbf{F}_n$ 内的点。通过对这些保留的点对 $\mathbf{F}_n$ 进行双线性插值，可以得到第 n 个视图的圆柱面特征。最后，通过将这些圆柱面特征的 N 个视图逐一叠加来生成完整的圆柱面特征。
Given the $n$-th view, the image $\mathbf{I}_n$ is fed into the feature extractor (\ie, Radio-v2.5 \citep{Radio}) to produce the image feature $\mathbf{F}_n$. Then we need to get the $n$-th view feature for the cylinder plane based on $\mathbf{F}_n$. To do this,  we project the cylindrical coordinates of all discrete points on the cylinder plane to Cartesian coordinates, which are further projected to 2D camera pixel coordinates using the coordinate of the central point $\mathbf{c}_u^x$, the extrinsic parameter $\mathbf{W}_n^e$, and the intrinsic parameter $\mathbf{P}_n$ of the $n$-th camera. The projected points outside the feature $\mathbf{F}_n$ are abandoned, and only the ones within the feature $\mathbf{F}_n$ are kept. The $n$-th view cylinder plane feature is obtained by bilinear interpolation of $\mathbf{F}_n$ for those kept points. Finally, the cylinder plane feature is produced by overlaying these $N$ views of the cylinder plane feature one-by-one.  

% 考虑到相邻视图之间存在一定的重叠部分，我们将这个圆柱面特征的 $N$ 个视图以两种方式进行叠加：（1）顺时针叠加。这 $N$ 个视图从第 0 个视图到第 $(N-1)$ 个视图依次叠加，最终的圆柱面特征表示为 $\mathbf{F}_{cy}^{+} \in \mathbb{R}^{D_{feat} \times H_u \times W_u}$；（2）逆时针叠加。这 $N$ 个视图从第 $(N-1)$ 个视图到第 0 个视图依次叠加，最终的圆柱面特征表示为 $\mathbf{F}_{cy}^{-} \in \mathbb{R}^{D_{feat} \times H_u \times W_u}$。在叠加过程中，重叠点的特征是通过用新输入的特征覆盖现有的特征而产生的。有关这两种叠加方式的示例，请参阅补充材料。
Taking the fact into consideration that there exists some overlap between adjacent views, we overlay these $N$ views of the cylinder plane feature in two ways: (1) Overlaying in a clockwise manner. These $N$ views are overlaid from the 0-th view to the $(N-1)$-th view, and the final cylinder plane feature is denoted as $\mathbf{F}_{cy}^{+} \in \mathbb{R}^{D_{feat} \times H_u \times W_u}$; (2) Overlaying in a counter-clockwise manner. These $N$ views are overlaid from the $(N-1)$-th view to the 0-th view, and the final cylinder plane feature is denoted as $\mathbf{F}_{cy}^{-} \in \mathbb{R}^{D_{feat} \times H_u \times W_u}$. During the overlaying procedure, the features for the overlapping points are produced by overwriting the existing ones with the incoming ones. Please refer to the supplementary materials for the illustration of these two overlay ways.
% 通过将由深度估计器生成的深度图和置信度图（即 Metric3D-v2\citep{metric3dv2}）添加到图像特征中，然后按照上述相同的流程对其进行处理，我们生成了具有深度信息的圆柱面特征，记为 $\bar{\mathbf{F}}_{cy}^{+}$、$\bar{\mathbf{F}}_{cy}^{-} \in \mathbb{R}^{(D_{feat}+2) \times H_u \times W_u}$ 。
By augmenting the image features with depth and confidence maps generated by the depth estimator (\ie, Metric3D-v2\citep{metric3dv2}), and then processing them through the identical pipeline mentioned above, we generate the cylinder plane feature with depth information, denoted as $\bar{\mathbf{F}}_{cy}^{+}, \bar{\mathbf{F}}_{cy}^{-} \in \mathbb{R}^{(D_{feat}+2) \times H_u \times W_u}$.

% Note that the parameters $\Delta h$ and $\rho$ are adjustable and are set to different values for occupancy-aware, volume-aware and pixel-aware modules, as well as different camera configurations, which means that the cylinder plane features are slightly different for different modules and different camera configureations. %since the cylinder planes are affected by these two parameters. 
% Please refer to \Cref{Detailedarch} for more details about the settings of $\Delta h$ and $\rho$.

\subsection{Foreground Reconstruction} 
\label{foreground_reconstruction}
% 基于圆柱面特征，我们可以重建驾驶场景。然而，从密集的体素中重建场景会耗费大量的内存和计算资源，因为大多数体素都是空的，可以进行修剪处理。为了解决这个问题，我们分别重建前景和背景部分，然后将它们融合起来。此外，前景部分是通过专门的占用感知、体积感知和像素感知模块进行稀疏重建的。
Based on the cylinder plane feature, we can reconstruct the driving scene. However, reconstructing the scene from dense voxels is memory- and computation-intensive, as most voxels are empty and can be pruned. To handle this, we reconstruct the foreground and background separately and then fuse them. In addition, the foreground is sparsely reconstructed with the dedicated occupancy-aware, volume-aware, and pixel-aware modules. 

% \textbf{占用感知模块} 旨在根据二维图像特征来区分占用区域和空闲区域，其灵感来源于文献 \cite{wei2023surroundocc} 中的研究。该占用感知模块主要通过一个 Y 形网络（称为 YNet$_{\rm occ}$）来实现，如 \Cref{Ynet} 所示。YNet$_{\rm occ}$ 以 $\mathbf{F}_{cy}^{+}$ 和 $\mathbf{F}_{cy}^{-}$ 作为输入，并使用权重共享的双分支编码器。这两个分支的输出特征被融合，然后由单分支解码器进一步处理。由于 $\mathbf{F}_{cy}^{+}$ 和 $\mathbf{F}_{cy}^{-}$ 中的大多数特征是相同的，因此它们之间存在信息冗余。为了解决这个问题， $\mathbf{F}_{cy}^{-}$ 在输入网络之前进行翻转，而编码器的输出在融合之前也进行翻转。设 YNet$_{\rm occ}$ 的输出特征为 $\mathbf{F}^{occ} \in \mathbb{R}^{K D_{occ}\times H_u \times W_u}$，其通道维度可以被 $K$ 整除。通过重塑 $\mathbf{F}^{occ}$ 可得到 CPFG $\mathbf{F}^{occ}_{cy} \in \mathbb{R}^{D_{occ}\times K \times H_u \times W_u}$，如 \Cref{Ynet} 所示。借助 CPFG 的辅助，经过充分训练的特征提取器的空间感知能力得到了提升。
The \textbf{Occupancy-Aware Module} is designed to distinguish the occupied and empty spaces based on the 2D image features, which is inspired by the work in \cite{wei2023surroundocc}. The occupancy-aware module is mainly implemented with a Y-shaped Network (termed as YNet$_{\rm occ}$), as shown in \Cref{Ynet}. YNet$_{\rm occ}$ takes $\mathbf{F}_{cy}^{+}$ and $\mathbf{F}_{cy}^{-}$ as input with a weight-shared dual-branch encoder.
 The output features of these two branches are fused to be further processed by a single branch decoder.
Since most of the features in $\mathbf{F}_{cy}^{+}$ and $\mathbf{F}_{cy}^{-}$ are the same, there exists information redundancy between them. To handle this, $\mathbf{F}_{cy}^{-}$ is flipped before being fed into the network, and the output of the encoder is flipped back before fusion. 
Let $\mathbf{F}^{occ} \in \mathbb{R}^{K D_{occ}\times H_u \times W_u}$  be the output feature of YNet$_{\rm occ}$, whose channel dimension can be evenly divided by $K$. A CPFG $\mathbf{F}^{occ}_{cy} \in \mathbb{R}^{D_{occ}\times K \times H_u \times W_u}$ is obtained by reshaping $\mathbf{F}^{occ}$, as illustrated in \Cref{Ynet}. The spatial awareness of the well-pretrained feature extractor is lifted with the help of CPFG.

% 为了判断占据网格 $\mathbf{O}$ 中的体素是否被占用，需要将 $\mathbf{X}_o$ 中所有点的笛卡尔坐标投影到圆柱坐标系中（有关此投影的详细信息，请参阅补充材料）。然后，通过对 $\mathbf{F}^{occ}_{cy}$ 进行三线性插值，获得占据网格 $\mathbf{O}$ 的占据特征 $\mathbf{F}^{occ}_{o} \in \mathbb{R}^{D_{occ} \times L_o\times H_o\times W_o}$。基于占据特征 $\mathbf{F}^{occ}_{o}$，通过两个基于轻量级多层感知机的头生成一个 3D 占据概率图 $\hat{\mathbf{P}}_{3D} \in \mathbb{R}^{2 \times L_o\times H_o\times W_o}$ 和一个 2D 视觉里程计（BEV）占据概率图 $\hat{\mathbf{P}}_{2D} \in \mathbb{R}^{H_o\times W_o}$。 在推理阶段，被占用的体素的索引由 $\mathcal{I}_v=\{\mathbf{i}=(l_v， h_v， w_v) | \hat{\mathbf{P}}_{3D}[0， \mathbf{i}] < \hat{\mathbf{P}}_{3D}[1， \mathbf{i}] \}$ 来确定。
To judge whether the voxels in the occupancy grid $\mathbf{O}$ are occupied or not, the Cartesian coordinates of all points in $\mathbf{X}_o$ are projected into the cylindrical coordinates (please refer to the supplementary materials for the details of this projection). Then the occupancy feature $\mathbf{F}^{occ}_{o}\in \mathbb{R}^{D_{occ} \times L_o\times H_o\times W_o}$ for the occupancy grid $\mathbf{O}$ is obtained by trilinear interpolation of $\mathbf{F}^{occ}_{cy}$. Based on the occupancy feature $\mathbf{F}^{occ}_{o}$, a 3D occupancy probability map $\hat{\mathbf{P}}_{3D} \in \mathbb{R}^{2 \times L_o\times H_o\times W_o}$ 
and a 2D BEV occupancy probability map $\hat{\mathbf{P}}_{2D} \in \mathbb{R}^{H_o\times W_o}$ 
are produced by two lightweight MLP-based heads.
In the inference stage, the indices of occupied voxels are determined by $\mathcal{I}_v=\{\mathbf{i}=(l_v, h_v, w_v) | \hat{\mathbf{P}}_{3D}[0, \mathbf{i}] < \hat{\mathbf{P}}_{3D}[1, \mathbf{i}]\}$.

\begin{figure*}[t]
	\centering
		\centering
		\includegraphics[width=0.9\linewidth]{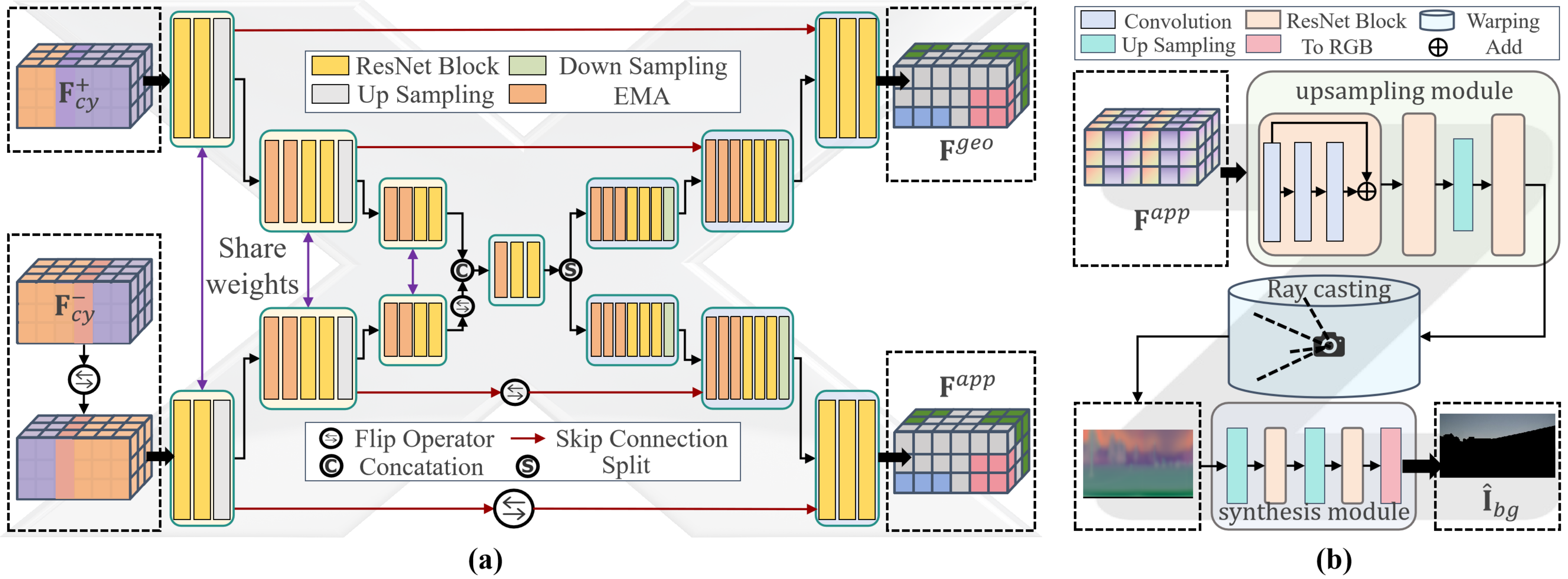}
		\caption{(a) \textbf{Architecture of XNet$_{\rm vol}$}, which is composed of a dual-branch encoder for downsampling and a dual-branch decoder for appearance and geometric feature upsampling. (b) \textbf{Overview of the Z-Net architecture.} It includes upsampling, ray casting and synthesis.}
		\label{Xnet}
 \vspace{-12pt}
\end{figure*}

% \textbf{体积感知模块} 用于为被占用的体素生成三维高斯分布，从而描绘出前景的轮廓。如图 \Cref{Xnet} 所示，该模块采用 X 形网络（称为 XNet$_{\rm vol}$）来实现。与 YNet$_{\rm occ}$ 类似，XNet$_{\rm vol}$ 以圆柱面特征 $\mathbf{F}_{cy}^{+}$ 和 $\mathbf{F}_{cy}^{-}$ 作为输入，并通过权重共享的双分支编码器进行处理。但不同之处在于，双分支的融合特征被输入到双分支解码器中，在此解码器中，两个分支分别生成几何特征 $\mathbf{F}^{geo} \in \mathbb{R}^{KD_{geo} \times H_u \times W_u}$ 和外观特征 $\mathbf{F}^{app} \in \mathbb{R}^{K D_{app} \times H_o \times W_o}$。这两个特征 $\mathbf{F}^{geo}$ 和 $\mathbf{F}^{app}$ 进一步重塑为几何 CPFG $\mathbf{F}^{geo}_{cy} \in \mathbb{R}^{D_{geo}\times K \times H_u \times W_u}$ 和外观 CPFG $\mathbf{F}^{app}_{cy} \in \mathbb{R}^{D_{app}\times K \times H_o \times W_o}$。与占用特征 $\mathbf{F}_{o}^{occ}$ 类似，对于占用网格的几何特征 $\mathbf{F}^{geo}_o \in \mathbb{R}^{D_{geo}\times L_o \times H_o \times W_o}$ 和外观特征 $\mathbf{F}^{app}_o \in \mathbb{R}^{D_{app}\times L_o \times H_o \times W_o}$，分别是通过对 $\mathbf{F}^{geo}_{cy}$ 和 $\mathbf{F}^{app}_{cy}$ 进行三线性插值而获得的。
The \textbf{Volume-Aware Module} is used to generate 3D Gaussians for the occupied voxels, which renders the outline of the foreground. As shown in \Cref{Xnet}, it is implemented with an X-shaped Network (termed as XNet$_{\rm vol}$). Similar to the YNet$_{\rm occ}$, XNet$_{\rm vol}$ takes the cylinder plane feature $\mathbf{F}_{cy}^{+}$ and $\mathbf{F}_{cy}^{-}$ as input with a weight-shared dual-branch encoder. But differently, the fused feature of the dual-branch is fed into a dual-branch decoder, where the two branches produce the geometry feature  $\mathbf{F}^{geo} \in \mathbb{R}^{KD_{geo} \times H_u \times W_u}$ and the appearance feature $\mathbf{F}^{app} \in \mathbb{R}^{K D_{app} \times H_o \times W_o}$ independently. The two features $\mathbf{F}^{geo}$ and $\mathbf{F}^{app}$ are further reshaped into the geometry CPFG  $\mathbf{F}^{geo}_{cy} \in \mathbb{R}^{D_{geo}\times K \times H_u \times W_u}$ and the appearance CPFG $\mathbf{F}^{app}_{cy} \in \mathbb{R}^{D_{app}\times K \times H_o \times W_o}$. 
Similar to the occupancy feature $\mathbf{F}_{o}^{occ}$,
the geometry feature $\mathbf{F}^{geo}_o \in \mathbb{R}^{D_{geo}\times L_o \times H_o \times W_o}$ and appearance feature $\mathbf{F}^{app}_o \in \mathbb{R}^{D_{app}\times L_o \times H_o \times W_o}$ for the occupancy grid are obtained by trilinear interpolation of $\mathbf{F}^{geo}_{cy}$ and $\mathbf{F}^{app}_{cy}$, respectively.

% 对于具有索引 $\mathbf{i} \in \mathcal{I}_v$ 的被占用体素，基于其几何和外观特征生成了 $G_v$ 个三维高斯分布 $\mathcal{G}_{\mathbf{i}} = \{\mathbf{X}_o[;\mathbf{i}]+\Delta \mathbf{x}_{\mathbf{i}}^g， \mathbf{c}_{\mathbf{i}}^g， \mathbf{\Sigma}_{\mathbf{i}}^g,\alpha_{\mathbf{i}}^g\}_{g=0}^{G_v-1}$，其中 $\mathbf{X}_o[;\mathbf{i}]+\Delta \mathbf{x}_{\mathbf{i}}^g$、$\mathbf{c}_{\mathbf{i}}^g$、$\mathbf{\Sigma}_{\mathbf{i}}^g$ 和 $\alpha_{\mathbf{i}}^g$ 分别是第 $g$ 个三维高斯的空间位置、球谐函数系数、各向异性协方差矩阵和不透明度。具体而言，空间偏移量 $\{\Delta\mathbf{x}_{\mathbf{i}}^g\}_{g=0}^{G_v-1}$ 是通过基于 $\mathbf{F}^{geo}_{o}[:，\mathbf{i}]$ 的轻量级多层感知机获得的，而其余高斯参数 $\{\mathbf{c}_{\mathbf{i}}^g， \mathbf{\Sigma}_{\mathbf{i}}^g,\alpha_{\mathbf{i}}^g\}_{g=0}^{G_v-1}$ 则是通过基于 $\mathbf{F}^{app}_{o}[:，\mathbf{i}]$ 的另一个轻量级多层感知机获得的。我们将所有被占用体素的三维高斯分布表示为 $\mathcal{G}_{\mathcal{I}_v} = \cup_{\mathbf{i} \in \mathcal{I}_v} \mathcal{G}_\mathbf{i}$ 。
For an occupied voxel with the index $\mathbf{i} \in \mathcal{I}_v$, $G_v$ 3D Gaussians $\mathcal{G}_{\mathbf{i}} = \{\mathbf{X}_o[;\mathbf{i}]+\Delta \mathbf{x}_{\mathbf{i}}^g, \mathbf{c}_{\mathbf{i}}^g, \mathbf{\Sigma}_{\mathbf{i}}^g,\alpha_{\mathbf{i}}^g\}_{g=0}^{G_v-1}$ are generated based on the geometry and appearance features, where $\mathbf{X}_o[;\mathbf{i}]+\Delta \mathbf{x}_{\mathbf{i}}^g$, $\mathbf{c}_{\mathbf{i}}^g$, $\mathbf{\Sigma}_{\mathbf{i}}^g$ and $\alpha_{\mathbf{i}}^g$ are the spatial position, spherical harmonics coefficients, anisotropic covariance matrix, and opacity of the $g$-th 3D Gaussian, respectively. Specifically, the spatial offset $\{\Delta\mathbf{x}_{\mathbf{i}}^g\}_{g=0}^{G_v-1}$ is obtained by a lightweight MLP based on $\mathbf{F}^{geo}_{o}[:,\mathbf{i}]$, while the rest Gaussian parameters $\{\mathbf{c}_{\mathbf{i}}^g, \mathbf{\Sigma}_{\mathbf{i}}^g,\alpha_{\mathbf{i}}^g\}_{g=0}^{G_v-1}$ are obtained by another lightweight MLP based on $\mathbf{F}^{app}_{o}[:,\mathbf{i}]$.
We denote 3D Gaussians for all occupied voxels as $\mathcal{G}_{\mathcal{I}_v} = \cup_{\mathbf{i} \in \mathcal{I}_v} \mathcal{G}_\mathbf{i}$.

% \textbf{像素感知模块} 用于为场景的纹理细化生成额外的 3D 傅里叶高斯函数。与占用感知模块类似，像素感知模块也是采用 Y 形网络实现的（称为 YNet$_{\rm pix}$）。与 YNet$_{\rm occ}$ 不同，YNet$_{\rm pix}$ 以 $\bar{\mathbf{F}}_{cy}^{+}$ 和 $\bar{\mathbf{F}}_{cy}^{-}$ 作为输入，并使用原始注意力机制而非 EMA。由于 YNet$_{\rm pix}$ 是为纹理细化而设计的，因此在经过 YNet$_{\rm pix}$ 处理之前/之后，其输入/输出都会通过因子 $k_i$/$k_o$ 进行上采样，以使生成的特征具有更高的空间分辨率（$k_i \cdot k_o$ = 4）。
The \textbf{Pixel-Aware Module} is used to generate additional 3D Gaussians for the texture refinement of the scene. Similar to the occupancy-aware module, the pixel-aware module is also implemented with a Y-shaped network (termed as YNet$_{\rm pix}$). Different from YNet$_{\rm occ}$, YNet$_{\rm pix}$ takes $\bar{\mathbf{F}}_{cy}^{+}$ and $\bar{\mathbf{F}}_{cy}^{-}$ as input and utilizes the original attention instead of EMA. Since YNet$_{\rm pix}$ is designed for the texture refinement, the input/output of YNet$_{\rm pix}$ is upsampled with the factor $k_i$/$k_o$ before/after being processed by YNet$_{\rm pix}$ to make the produced feature have higher spatial resolution ($k_i \cdot k_o$ = 4). 

% 设 $\mathbf{F}^{pix} \in \mathbb{R}^{D_{pix} \times 4H_u \times 4W_u}$ 为 YNet$_{\rm pix}$ 的上采样输出，并且 $\mathcal{I}_p = \{\mathbf{i}=(h_u， w_u) | 0 \leq h_u < 4H_u，  0 \leq w_u < 4W_u， \}$ 表示 $\mathbf{F}^{pix}$ 中所有像素的索引，对于 $\mathbf{i} \in \mathcal{I}_p$ 的每个像素，我们生成 $G_p$ 个三维高斯分布 $\mathcal{G}_{\mathbf{i}} = \{(\mathbf{x}_{\mathbf{i}}+\Delta \mathbf{x}_{\mathbf{i}}^g， \mathbf{c}_{\mathbf{i}}^g， \mathbf{\Sigma}_{\mathbf{i}}^g,\alpha_{\mathbf{i}}^g)\}_{g=0}^{G_p-1}$，其中 $\mathbf{x}_{\mathbf{i}}$ 是借助深度估计器估计出的深度来投影得到的像素 $\mathbf{i}$ 的笛卡尔坐标。高斯参数 $\{ \Delta \mathbf{x}_{\mathbf{i}}^g， \mathbf{c}_{\mathbf{i}}^g， \mathbf{\Sigma}_{\mathbf{i}}^g， \alpha_{\mathbf{i}}^g \}_{g=0}^{G_p-1}$ 是基于 $\mathbf{F}^{pix}[:， \mathbf{i}]$ 由一个轻量级的多层感知机估计得到的。所有像素的三维高斯分布表示为 $\mathcal{G}_{\mathcal{I}_p} = \cup_{\mathbf{i} \in \mathcal{I}_p} \mathcal{G}_\mathbf{i}$。
Let $\mathbf{F}^{pix} \in \mathbb{R}^{D_{pix} \times 4H_u \times 4W_u}$ be the upsampled output of YNet$_{\rm pix}$, and $\mathcal{I}_p = \{\mathbf{i}=(h_u, w_u) | 0 \leq h_u < 4H_u,  0 \leq w_u < 4W_u, \}$ be the indices of all pixels in $\mathbf{F}^{pix}$, we generate $G_p$ 3D Gaussians $\mathcal{G}_{\mathbf{i}} = \{(\mathbf{x}_{\mathbf{i}}+\Delta \mathbf{x}_{\mathbf{i}}^g, \mathbf{c}_{\mathbf{i}}^g, \mathbf{\Sigma}_{\mathbf{i}}^g,\alpha_{\mathbf{i}}^g)\}_{g=0}^{G_p-1}$ for each pixel $\mathbf{i} \in \mathcal{I}_p$, where $\mathbf{x}_{\mathbf{i}}$ is the projected Cartesian coordinates of pixel $\mathbf{i}$ with the help of the estimated depth from the depth estimator. The Gaussian parameters $\{ \Delta \mathbf{x}_{\mathbf{i}}^g, \mathbf{c}_{\mathbf{i}}^g, \mathbf{\Sigma}_{\mathbf{i}}^g, \alpha_{\mathbf{i}}^g \}_{g=0}^{G_p-1}$ are estimated by a lightweight MLP based on $\mathbf{F}^{pix}[:, \mathbf{i}]$. The 3D Gaussians for all pixels are denoted as $\mathcal{G}_{\mathcal{I}_p} = \cup_{\mathbf{i} \in \mathcal{I}_p} \mathcal{G}_\mathbf{i}$.

% 前景部分由三维高斯分布表示为 $\mathcal{G}_{fg} = \mathcal{G}_{\mathcal{I}_v} \cup \mathcal{G}_{\mathcal{I}_p}$。给定具有相机参数 $\mathbf{W}^{e}_{t}$ 和 $\mathbf{P}_{t}$ 以及空间分辨率 $H_t \times W_t$ 的目标视图，我们可以通过对 $\mathcal{G}_{fg}$ 进行栅格化渲染来获得渲染后的前景图像 $\hat{\mathbf{I}}_{fg} \in \mathbb{R}^{3 \times H_t \times W_t}$、$\alpha$ 映射 $\hat{\mathbf{A}}_{fg} \in \mathbb{R}^{H_t \times W_t}$ 和深度图 $\hat{\mathbf{D}}_{fg} \in \mathbb{R}^{H_t \times W_t}$。
The foreground is represented by 3D Gaussians as $\mathcal{G}_{fg} = \mathcal{G}_{\mathcal{I}_v} \cup \mathcal{G}_{\mathcal{I}_p}$. Given a target view with the camera parameters $\mathbf{W}^{e}_{t}$ and $\mathbf{P}_{t}$, and spatial resolution $H_t \times W_t$, we can obtain the rendered foreground image $\hat{\mathbf{I}}_{fg} \in \mathbb{R}^{3 \times H_t \times W_t}$, $\alpha$-map $\hat{\mathbf{A}}_{fg} \in \mathbb{R}^{H_t \times W_t} $ and depth map $\hat{\mathbf{D}}_{fg} \in \mathbb{R}^{H_t \times W_t}$ with the rasterization rendering of $\mathcal{G}_{fg}$.

\subsection{Background Reconstruction and Fusion} 
\label{background_reconstruction}
% 背景是在二维空间中通过一个 Z 形网络（称为 ZNet$_{\rm bg}$）生成的，并与渲染的前景图像 $\hat{\mathbf{I}}_{fg}$ 进行融合。如 \Cref{Xnet} 所示，ZNet$_{\rm bg}$ 以由 XNet$_{\rm vol}$ 生成的外观特征 $\mathbf{F}^{app}$ 作为输入，并生成生成的背景图像。与像素感知模块 YNet$_{\rm pix}$ 类似，ZNet$_{\rm bg}$ 也需要生成背景的详细纹理（即天空和云朵）。考虑到这一点，背景图像以更高的分辨率（具体而言，$4H_t \times 4 W_t$）生成，然后降采样到目标分辨率（即 $H_t \times W_t$），并与前景图像融合。
The background is generated in the 2D space with a Z-shaped Network (termed as ZNet$_{\rm bg}$) and fused with the rendered foreground image $\hat{\mathbf{I}}_{fg}$. 
As shown in \Cref{Xnet}, ZNet$_{\rm bg}$ takes the appearance feature $\mathbf{F}^{app}$ produced by XNet$_{\rm vol}$ as input, and produces the generated background image. Similar to the pixel-aware module YNet$_{\rm pix}$, ZNet$_{\rm bg}$ also needs to generate the detailed texture of the background (\ie, the sky and clouds).  Taking this into consideration, the background image is generated at a higher resolution (in detail, $4H_t \times 4 W_t$) and downsampled to the target resolution (\ie, $H_t \times W_t$) and fused with the foreground image. 

% 在 ZNet$_{\rm bg}$ 中，背景生成过程分为三个步骤。 （1）使用基于 ResNet 的 \citep{he2016deep} 增强模块将输入特征放大两倍。 （2）从目标视图的相机中心向目标分辨率 $H_t \times W_t$ 的每个像素发射光线。 计算这些光线与步骤（1）中放大后的特征的交点（请参考补充材料），这些交点用于对目标特征（分辨率 $H_t \times W_t$）进行双线性采样，以生成目标视图的背景图像 $\hat{\mathbf{I}}_{bg} \in \mathbb{R}^{3 \times 4H_t \times 4W_t}$。 （3）基于步骤（2）中采样的目标特征，使用基于 StyleGAN 的 \citep{stylegan} 合成模块生成背景图像 $\hat{\mathbf{I}}_{bg} \in \mathbb{R}^{3 \times 4H_t \times 4W_t}$。
The generation of background is divided into three steps within ZNet$_{\rm bg}$. (1) Upsample the input feature by $2\times$ with a ResNet-based \citep{he2016deep} module. (2) Cast rays from the camera center of the target view through each pixel of the target resolution $H_t \times W_t$. The intersection points of these rays and the upsampled feature in step (1) are computed (refer to the supplementary materials), which are used to bilinearly sample the target feature (with the resolution of $H_t \times W_t$) for the target view. (3) Generate the background image $\hat{\mathbf{I}}_{bg} \in \mathbb{R}^{3 \times 4H_t \times 4W_t}$ using a StyleGAN-based \citep{stylegan} synthesis module based on the target feature sampled in step (2).

% 为了将背景图像与前景图像融合在一起，将背景图像 $\hat{\mathbf{I}}_{bg}$ 下采样至目标分辨率 $\hat{\mathbf{I}}^{'}_{bg} \in \mathbb{R}^{3 \times H_t \times W_t}$。然后，采用 $\alpha$ 混合方法将下采样的背景图像与前景图像 $\hat{\mathbf{I}} = \hat{\mathbf{I}_{fg}} + (1 - \hat{\mathbf{A}}_{fg}) \otimes \hat{\mathbf{I}}_{bg}^{'}$ 进行融合。
To fuse the background image with the foreground image, the background image $\hat{\mathbf{I}}_{bg}$ is downsampled to the target resolution  $\hat{\mathbf{I}}^{'}_{bg} \in \mathbb{R}^{3 \times H_t \times W_t}$. Then, $\alpha$-blending is adopted to fuse the downsampled background image with the foreground image $\hat{\mathbf{I}} = \hat{\mathbf{I}_{fg}} + (1- \hat{\mathbf{A}}_{fg}) \otimes \hat{\mathbf{I}}_{bg}^{'}$. 

% % 
\begin{figure*}[t]
	\centering
	\includegraphics[width=0.9\linewidth]{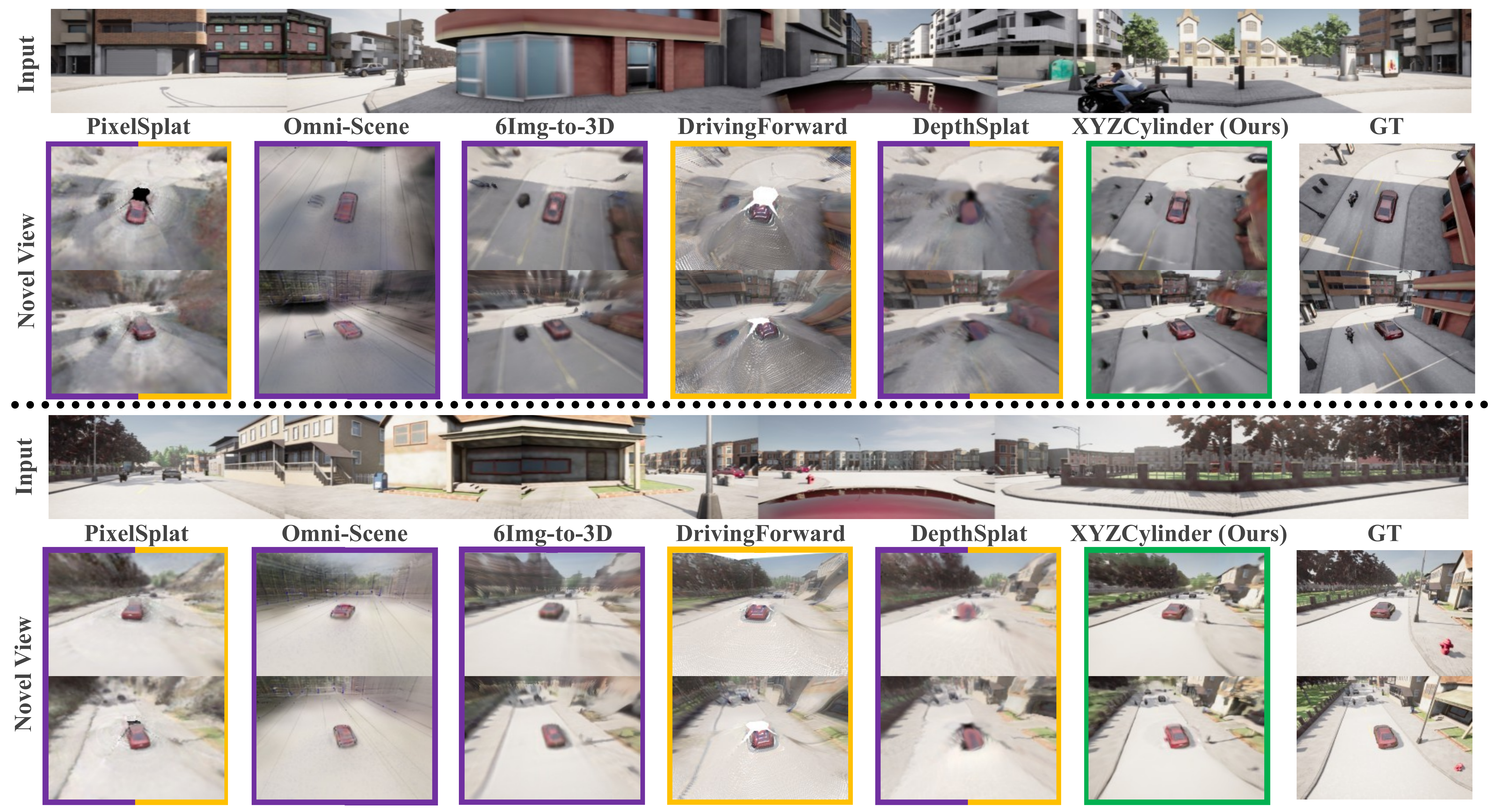}
 % \textbf{不同方法在 Carla-Centric 上的定性比较。} \fcolorbox{黄色框}{白色}{黄色框} 表示场景空洞现象（即白色区域），\fcolorbox{紫色框}{白色}{紫色框} 表示模糊伪影，而 \fcolorbox{绿色框}{白色}{绿色框} 则表明我们的 XYZCylinder 成功规避了上述两个问题。
	\caption{\textbf{Qualitative comparison of different methods on Carla-Centric.} \fcolorbox{YellowBox}{white}{Yellow boxes} indicate scene void phenomena (\ie, white regions), \fcolorbox{PurpleBox}{white}{purple boxes} denote blurring artifacts, and \fcolorbox{GreenBox}{white}{green boxes} signify that our XYZCylinder successfully circumvents both aforementioned issues.}
	\label{carla}
 \vspace{-12pt}
\end{figure*}

\subsection{Optimization of XYZCylinder}
% XYZ 立体缸体的优化分为两个阶段。（1）通过监督 3D 和 2D 占据概率图 $\hat{\mathbf{P}}_{3D}$ 和 $\hat{\mathbf{P}}_{2D}$ 来训练占位感知模块。对于 $\hat{\mathbf{P}}_{3D}$，使用交叉熵损失、语义损失 \citep{monoscene} 和几何损失 \citep{monoscene}，而对于 $\hat{\mathbf{P}}_{2D}$，则采用 BEV 损失 \citep{FastOcc}。通过使用泊松重建聚合语义点云，并随后进行体素化操作来生成真实占位图。（2）使用冻结的占位感知模块来训练其余模块，通过监督渲染图像 $\hat{\mathbf{I}}$、$\alpha$ 图 $\hat{\mathbf{A}}_{fg}$ 和深度 $\hat{\mathbf{D}}_{fg}$ 来进行训练。具体而言，$\hat{\mathbf{I}}$ 由重建 L1 损失和感知相似性损失 \citep{lpips} 来监督。$\hat{\mathbf{A}}_{fg}$ 仅由重建 L1 损失来监督，而 $\alpha$ 图的真实值则通过 LISA \citep{LISA} 获取。$\hat{\mathbf{D}}_{fg}$ 由重建的 L1 损失和皮尔逊深度损失进行监督 \citep{sparsegs}，而真实深度图则是由深度估计器获取的。更多关于数据预处理和详细损失的详情，请参阅补充材料。
The optimization of XYZCylinder is divided into two stages. (1) Training the occupancy-aware module by supervising the 3D and 2D occupancy probability maps $\hat{\mathbf{P}}_{3D}$ and $\hat{\mathbf{P}}_{2D}$. The cross-entropy loss, semantic loss \citep{monoscene}, and geometric loss \citep{monoscene} are used for $\hat{\mathbf{P}}_{3D}$, while the BEV loss \citep{FastOcc} is adopted for $\hat{\mathbf{P}}_{2D}$. The ground-truth occupancy map is generated by aggregating semantic point clouds using Poisson reconstruction, followed by a voxelization operation. (2) Training the rest modules with the frozen occupancy-aware module by supervising the rendered image $\hat{\mathbf{I}}$, $\alpha$-map $\hat{\mathbf{A}}_{fg}$, and depth $\hat{\mathbf{D}}_{fg}$. Specifically,  $\hat{\mathbf{I}}$ is supervised by the reconstruction L1 loss and perceptual similarity loss \citep{lpips}. 
$\hat{\mathbf{A}}_{fg}$ is only supervised by the reconstruction L1 loss, and the ground-truth of the $\alpha$-map is obtained by LISA \citep{LISA}. 
 $\hat{\mathbf{D}}_{fg}$ is supervised by the reconstruction L1 loss and Pearson depth loss \citep{sparsegs}, and the ground-truth depth map is obtained by the depth estimator. 
Please refer to the supplementary materials for more details of data preprocessing and detailed losses.

\section{Experiments}
\label{experiments}

\textbf{Baselines.} We select recent works, including SplatterImage \citep{szymanowicz2024splatter}, PixelSplat \citep{pixelsplat}, MVSplat \citep{mvsplat}, DepthSplat \citep{depthsplat}, DrivingForward \citep{drivingforward}, Omni-Scene \citep{Omniscene}, and 6Img-to-3D \citep{6imgto3d} as the baselines for ego-forward and ego-inward tasks. Implementation details and modifications for the baselines on both datasets are provided in the supplementary materials.

\noindent
\textbf{Datasets.}
We follow Omni-Scene \citep{Omniscene} and 6Img-to-3D \citep{6imgto3d} to evaluate ego-forward and ego-inward reconstruction on nuScenes and Carla-Centric, respectively. In both evaluation settings, 6 views are used without specification. In addition, Waymo \citep{waymo}, Pandaset \citep{pandaset}, ONCE \citep{once}, and Argoverse \citep{argoverse} are used for zero-shot evaluation.
For more details, please refer to the supplementary materials.

\noindent
\textbf{Metrics.} We evaluate our method's photometric quality using the standard metrics of PSNR, SSIM, and LPIPS. To assess geometric accuracy, we compute the PCC between the rendered depth maps and predicted depth maps. The detailed formulas for these metrics can be found in the supplementary materials.

\noindent
\textbf{Comparison with Baselines.} \Cref{tab:main-table} presents the quantitative comparison between the proposed XYZCylinder and several baselines. We also introduce vanilla 3DGS, and calculate metrics on 100 random scenes from the two datasets after per-tile training, demonstrating the failure of iterative methods for sparse-view driving reconstruction.  Overall, XYZCylinder achieves all the best metrics on both datasets.
Beyond its superior hybrid representation and advanced architecture, the superior performance of our method also stems from two aspects. First, by decoupling the reconstruction of foreground and background, we prevent the foreground 3D Gaussians $\mathcal{G}_{fg}$ from being misplaced into the background region. Second, the frozen occupancy-aware module can constrain the optimization space of the volume-aware module, benefiting the generation of 3D Gaussians $\mathcal{G}_{\mathcal{I}_v}$ which are vital to the geometric accuracy.

\begin{table}[t]
\begin{center}

            \captionof{table}{\textbf{Quantitative comparison of our model against the baselines.} The \sethlcolor{Red}\hl{best}, \sethlcolor{Orange}\hl{second-best}, and \sethlcolor{Yellow}\hl{third-best} results are marked with colors.}

            \setlength{\tabcolsep}{1.3pt}
            \scriptsize
            \label{tab:main-table}
        	\begin{tabular}{c|c|c|c|c|c|c|c|c}
                \Xhline{1.2pt}
        		                  & \multicolumn{4}{c|}{\textbf{nuScenes} (ego-forward)}             & \multicolumn{4}{c}{\textbf{Carla-Centric} (ego-inward)}          \\
                \hline
        		\textbf{Models}            & PSNR$\uparrow$      & LPIPS$\downarrow$    & SSIM$\uparrow$     & PCC$\uparrow$      & PSNR$\uparrow$    & LPIPS$\downarrow$   & SSIM$\uparrow$    & PCC$\uparrow$     \\
                \hline 
                3DGS &     12.30    &     0.633    &   0.317      &    -0.054      &  8.49       &   0.686      &    0.337     &   -0.078       \\
                \hline 
        		SplatterImage     & 17.31        & 0.661        & 0.442        & 0.027         & 13.04        & 0.708        & 0.448        & 0.180         \\
        		PixelSplat        & 21.33        & 0.376        & 0.607        & 0.077         & 14.67        & 0.565        & 0.412        & 0.554         \\
        		MVSplat           & 21.87        & 0.342        & 0.621        & 0.201         & 15.32        & 0.507        & 0.457        & 0.566            \\
        		DepthSplat        & 23.19        & 0.339        & 0.675        & 0.431         & \third16.27  & \third0.503  & \third0.508  & 0.581         \\
        		DrivingForward    & \third23.84  & \third0.280  & \second0.739 & 0.437         & 15.38        & 0.541        & 0.445        & \third0.696  \\
        	    6Img-to-3D        & 20.74        & 0.650        & 0.560        & \third0.570   & \second17.33 & \second0.485 & \second0.620 & \second0.765      \\	
                  Omni-Scene        & \second24.11 & \second0.242 & \third0.734  & \second0.816  & 15.54        & 0.558        & 0.462        & 0.551            \\
              \hline
        		XYZCylinder              & \best24.97   & \best0.231   & \best0.750   & \best0.887    & \best18.40   & \best0.359   & \best0.622   & \best0.817     \\
                \Xhline{1.2pt}
        	\end{tabular}
        	\end{center}
\vspace{-16pt}
\end{table}

\begin{figure*}[t]
	\centering
	\includegraphics[width=0.9\linewidth]{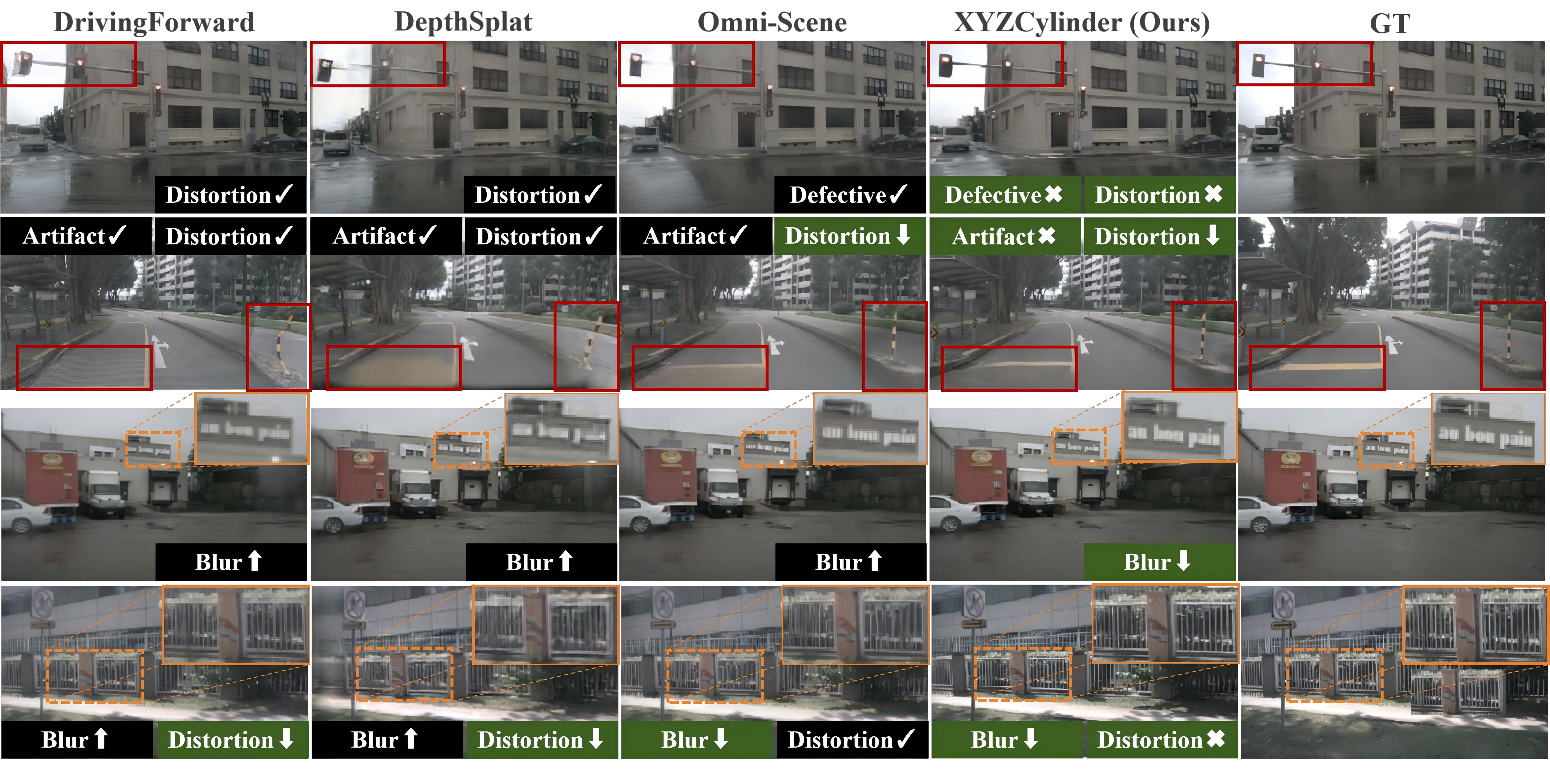}

	\caption{\textbf{Qualitative comparison of different methods on nuScenes.} The baseline method exhibits significant geometric distortions (warping artifacts), structural defects (incomplete geometries), texture blurring, and visual artifacts. In contrast, our XYZCylinder produces superior visual quality with much reduced artifacts.}
	\label{nuScenes}
 \vspace{-12pt}
\end{figure*}

Interestingly, all methods achieve substantially better performance on nuScenes than on Carla-Centric. The reason is that the viewpoint shifts between the training and test sets of nuScenes are smaller than those of Carla-Centric, reducing the demand for geometric reconstruction capability. It demonstrates the necessity of introducing the Carla-Centric dataset. Taking the PSNR on nuScenes for example, some pixel-based methods (\eg, DepthSplat and DrivingForward) achieve better PSNR than the volume-based method (\ie, 6Img-to-3D), and achieve considerable PSNR with volume-pixel-based methods (\ie, Omni-Scene and XYZCylinder). Nevertheless, the geometric accuracy of pixel-based methods, particularly in depth estimation (\ie, PCC), remains limited. Although both our XYZCylinder and Omni-Scene are volume-pixel-based methods, Omni-Scene suffers from suboptimal performance. This is primarily because its volumetric branch relies on feature injection from the pixel branch and employs volumetric sampling lacking geometric constraints.

The qualitative comparisons of different methods further highlight the superiority of XYZCylinder. As shown in \Cref{carla} and \Cref{nuScenes}, XYZCylinder produces significantly better visual quality on Carla-Centric with finer texture details and more accurate geometry, while other methods are plagued by the artificial holes (\ie, white regions in rendered images of DrivingForward). On nuScenes, XYZCylinder further demonstrates its advantages through enhanced geometric completeness (\eg, reconstructing a complete traffic light), superior texture fidelity with fewer artifacts and sharper text, and higher geometric accuracy, free from ghosting or distortion. Extended analyses and further discussions are available in the supplementary materials.

\noindent
\textbf{Ablation on the Importance of Main Designs.} 
Here, we show the importance of the separate reconstruction of foreground and background, as well as the necessity of the occupancy/volume/pixel-aware module for foreground reconstruction. Results are shown in \Cref{tab:ab1}. When the background (\ie, sky and clouds) is treated as the foreground and reconstructed by 3D Gaussians (w/o. Separate Reconstruction), or the volume-aware and pixel-aware modules are removed (w/o. Volume and w/o. Pixel), all metrics on two datasets are reduced, indicating the necessity of the key designs. In addition, XYZCylinder runs out of memory if the occupancy-aware module is removed by treating all voxels as occupied (w/o. Occupancy), which means that the occupancy-aware module can save memory efficiently. The full model's superior performance validates our hybrid representation's effectiveness on reconstruction accuracy.

\begin{table}
	\begin{center}
    \setlength{\tabcolsep}{1pt}
    \caption{\textbf{Ablation study on our main designs.}}
    \scriptsize
    \label{tab:ab1}
	\begin{tabular}{c|c|c|c|c|c|c|c|c}
        \Xhline{1.2pt}
		                  & \multicolumn{4}{c|}{\textbf{nuScenes} (ego-forward)}             & \multicolumn{4}{c}{\textbf{Carla-Centric} (ego-inward)}          \\
        \hline
		\textbf{Models}            & PSNR$\uparrow$      & LPIPS$\downarrow$    & SSIM$\uparrow$     & PCC$\uparrow$      & PSNR$\uparrow$      & LPIPS$\downarrow$    & SSIM$\uparrow$     & PCC$\uparrow$     \\
        \hline 
        w/o. Separate   & \third23.25   & \third0.280   & \second0.725   & \third0.854       & \third17.58     & \third0.371     & \third0.611     & \second0.797      \\
        \hline
        w/o. Occupancy & \multicolumn{8}{c}{\cellcolor[HTML]{F5F5F5} Out of Memory} \\
        w/o. Volume     & \second23.92   & \second0.256   & \third0.722   & \second0.886     & 15.67     & 0.589     & 0.455     & \third0.694      \\
        w/o. Pixel    & 22.41   & 0.338   & 0.661   & 0.667  & \second18.19     & \second0.367     & \second0.616     & 0.521  \\  
        \hline
         XYZCylinder & \best{24.97}   & \best{0.231}   & \best{0.750}   & \best{0.887}  & \best{18.40}     & \best{0.359}     & \best{0.622}     & \best{0.817}    \\
        \Xhline{1.2pt}
	\end{tabular}
	\end{center}
 \vspace{-16pt}
\end{table}

\begin{table}
        	\begin{center}
            \setlength{\tabcolsep}{1.3pt}
            \caption{\textbf{Ablation study on the design of X/Y-shape network.}}
            \scriptsize
            \label{tab:xy}
        	\begin{tabular}{c|c|c|c|c|c|c|c|c}
                \Xhline{1.2pt}
        		                  & \multicolumn{4}{c|}{\makecell{\textbf{nuScenes}(ego-forward)}}             & \multicolumn{4}{c}{\makecell{\textbf{Carla-Centric}(ego-inward)}}          \\
                \hline
        		 \quad \textbf{Models}             & PSNR$\uparrow$      & LPIPS$\downarrow$    & SSIM$\uparrow$     & PCC$\uparrow$      & PSNR$\uparrow$      & LPIPS$\downarrow$    & SSIM$\uparrow$     & PCC$\uparrow$     \\
                \hline 
                w/o. X-flip      & \second{24.88}   & \third0.240   & \second{0.747}   & \best0.890      & 18.28         & 0.381        & 0.603         & \third0.802    \\  
                w/o. Y-flip      & 24.33      & 0.245         & 0.734            & 0.881           & \third18.36  & 0.364  & \second0.620  & 0.793          \\  
                w/o. X-dual enc     & 24.16            & 0.260         & 0.721            & 0.876           & 18.21         & 0.377        & 0.607         & 0.793          \\
                w/o. Y-dual enc     & 24.04            & 0.256         & 0.726            & \second0.887           & 18.34   & \second0.360 & 0.618         & 0.796          \\
                w/o. X-dual dec      & 24.31            & \second0.239  & \third0.741      & 0.885   & 18.33         & 0.367        & \second0.620   & \second{0.807} \\
                w/o. Y-depth      & \third24.46            & 0.242  & 0.740      & 0.875   & \second18.37         & \third0.361        & 0.618   & 0.800 \\
                \hline 
        		XYZCylinder    & \best{24.97}     & \best{0.231}  & \best{0.750}     & \second{0.887}  & \best{18.40}  & \best{0.359} & \best{0.622}  & \best{0.817}   \\
                \Xhline{1.2pt}
        	\end{tabular}
        	\end{center}
        	
\vspace{-8mm}
\end{table}

\begin{figure*}[t]
	\centering
	\includegraphics[width=\linewidth]{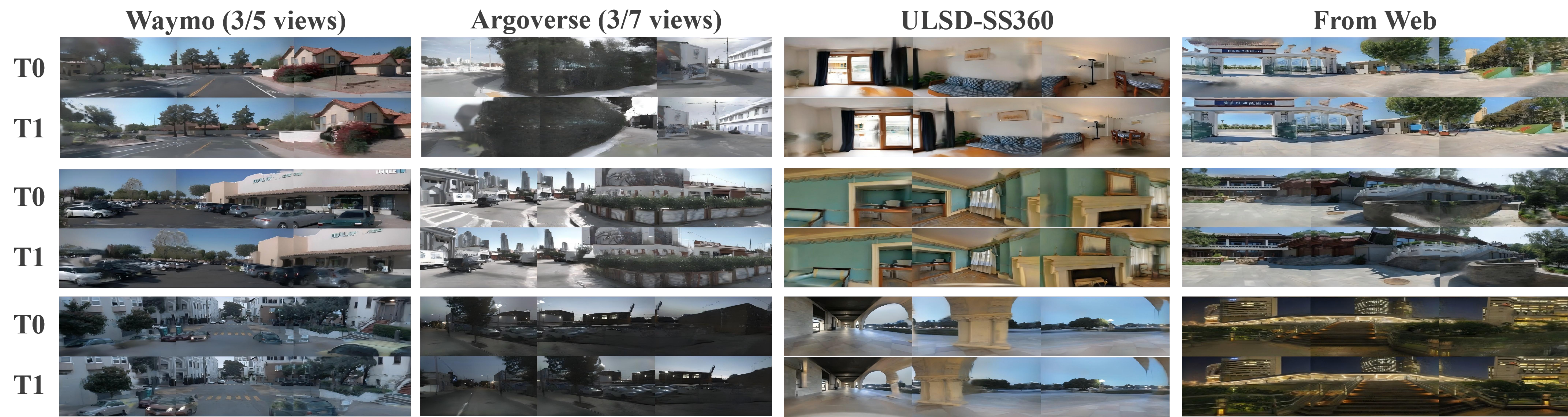}
	\caption{\textbf{Zero-shot results on other datasets.} The XYZCylinder, trained solely on nuScenes, is evaluated. Only the reconstructed 3 views are presented due to the limited space.}
	\label{zeroshot}
 \vspace{-12pt}
\end{figure*}

\noindent
\textbf{Ablation on the Design of X-shape and Y-shape Networks.}
The common designs in both networks are the dual-branch encoder and the flip operation of the counter-clockwise cylinder plane feature. To show the effectiveness of such designs, we remove the flip operation (w/o. X-flip and w/o. Y-flip), and replace the dual-branch encoder with a single-branch encoder which takes the clockwise cylinder plane feature as input (w/o. X-dual enc and w/o. Y-dual enc). In addition, the dual-branch decoder in the X-shape network (XNet$_{\rm vol}$) disentangles the geometry and appearance information of the scene, which is beneficial for the background reconstruction. The effectiveness of such a design is validated by replacing the dual-branch decoder with a single-branch decoder (w/o. X-dual dec), where the single output feature is used for the generation of all parameters for 3D Gaussians in $\mathcal{G}_{\mathcal{I}_v}$ and the generation of background. Finally, we remove the depth input of $\text{YNet}_{\rm pix}$ to evaluate the significance of depth input. Results are shown in \Cref{tab:xy}. As we can see, all the counterpart models achieve inferior performance to XYZCylinder, demonstrating the effectiveness of our designs. In addition, the models w/o. X-dual enc and w/o. Y-dual enc produce much worse results than others, indicating that the construction of the CPFG relies on better feature extraction.  

\noindent
\textbf{Zero-shot on Various Datasets.} 
Baselines lack compatibility across driving datasets because their feature representations are tailored to a specific number and distribution of views. The model architecture of DepthSplat and DrivingForward is independent of the number of views, as view-related operations are handled at the feature interaction stage. In contrast, Omni-Scene requires complete retraining for a different number of cameras. We construct its variant Omni-Scene-V, extending the embedding to a specified number of cameras, demonstrating degraded performance (\Cref{tab:jianrong}). In contrast, our XYZCylinder is robust to camera variations. UCCM aggregates all view features onto a Cylinder Plane. Consequently, our model, trained only on nuScenes, shows strong zero-shot generalization on the Waymo \citep{waymo} and Argoverse \citep{argoverse} datasets (\Cref{zeroshot}). We further confirmed this versatility on the ULSD-SS360 \cite{ULSD} panoramic indoor dataset and panoramic images from the web. 
See the supplementary materials for additional zero-shot results on Pandaset \citep{pandaset} and ONCE \citep{once}.

\begin{table}
        	\begin{center}
            \setlength{\tabcolsep}{1.6pt}
            \caption{\textbf{Zeroshot performance of models trained on nuScenes.}}
            \scriptsize
            \label{tab:jianrong}
        	\begin{tabular}{c|c|c|c|c|c|c|c|c|c}
                \Xhline{1.2pt}
        		         \textbf{nuScenes (6)}       & \multicolumn{3}{c|}{\makecell{ $\rightarrow$ \textbf{Argoverse (7)}}}             & \multicolumn{3}{c|}{\makecell{$\rightarrow$ \textbf{Waymo (5)}}}   & \multicolumn{3}{c}{\makecell{$\rightarrow$ \textbf{Pandaset (6)}}}       \\
                \hline
        		 \quad \textbf{Models}             & PSNR    & LPIPS    & SSIM    & PSNR      & LPIPS    & SSIM & PSNR      & LPIPS    & SSIM \\
                \hline 
                DrivingForward      & \third{17.21}   & \third{0.484}   & \third0.433   & \third15.47         & \second0.515        & \second0.456     & \second14.77 & \second0.528 & \second0.390   \\  
                DepthSplat      & \second{17.28}   & \second{0.445}   & 0.426   & \second16.69         & \third0.563        & \third0.404       & \third14.65& \third0.578&  \third0.361 \\  
                Omni-Scene      & \multicolumn{6}{c|}{\cellcolor[HTML]{F5F5F5} Need Retraining} & 13.54 &  0.784 & 0.341 \\  
                Omni-Scene-V      & 12.68 & 0.695 & \second0.454 & 13.63 & 0.723 & 0.382 & 13.54 &  0.784 & 0.341 \\  
        		XYZCylinder    & \best{18.40}     & \best{0.394}  & \best{0.520}     & \best{19.04}  & \best{0.422} & \best{0.501}  & \best18.09 & \best0.383 & \best0.567 \\
                \Xhline{1.2pt}
        	\end{tabular}
        	\end{center}
        	
\vspace{-6mm}
\end{table}

% \vspace{2mm}
\section{Conclusion}
% 我们推出了 XYZCylinder，这是一款专为驱动场景的稀疏三维重建而设计的前馈框架。XYZCylinder 采用了新颖的统一圆柱体提升方法，以提高兼容性和重建精度。我们的模型使用明确的、确定性的视图变换和混合表示来重建前景和背景，确保了高保真度的三维重建以及相机参数的兼容性。我们模型的成功进一步拓展了前馈模型在生成完整三维场景和为三维自动驾驶系统构建仿真环境方面的能力边界。在 nuScenes 和 Carla-Centric 上以及零样本结果的大量实验表明，XYZCylinder 具有出色的重建质量和相机兼容性，揭示了其将多个数据源连接起来以提供统一架构和接口来训练大规模模型的巨大潜力。
We present XYZCylinder, a feedforward framework designed for sparse 3D reconstruction of driving scenes. XYZCylinder employs a novel Unified Cylinder Lifting Method to enhance compatibility and reconstruction accuracy. Our model utilizes explicit, deterministic view transformations and hybrid representations to reconstruct scenes, ensuring high-fidelity 3D representation and camera compatibility. The success of our model further expands the capability boundary of the feed-forward models in generating 3D scenes and constructing simulation environments for autonomous driving systems. Extensive experiments on nuScenes and Carla-Centric as well as the zero-shot results demonstrate the outstanding reconstruction quality and camera compatibility of XYZCylinder, revealing its enormous potential for connecting multiple data sources to provide a unified architecture for training large-scale models.

\newpage
{
    \small
    \bibliographystyle{ieeenat_fullname}
    \bibliography{main}

\begin{thebibliography}{68}
\providecommand{\natexlab}[1]{#1}
\providecommand{\url}[1]{\texttt{#1}}
\expandafter\ifx\csname urlstyle\endcsname\relax
  \providecommand{\doi}[1]{doi: #1}\else
  \providecommand{\doi}{doi: \begingroup \urlstyle{rm}\Url}\fi

\bibitem[Cao and De~Charette(2022)]{monoscene}
Anh-Quan Cao and Raoul De~Charette.
\newblock Monoscene: Monocular 3d semantic scene completion.
\newblock In \emph{Proceedings of the IEEE/CVF Conference on Computer Vision
  and Pattern Recognition}, pages 3991--4001, 2022.

\bibitem[Caron et~al.(2021)Caron, Touvron, Misra, J{\'e}gou, Mairal,
  Bojanowski, and Joulin]{dino}
Mathilde Caron, Hugo Touvron, Ishan Misra, Herv{\'e} J{\'e}gou, Julien Mairal,
  Piotr Bojanowski, and Armand Joulin.
\newblock Emerging properties in self-supervised vision transformers.
\newblock In \emph{Proceedings of the IEEE/CVF international conference on
  computer vision}, pages 9650--9660, 2021.

\bibitem[Chan et~al.(2022)Chan, Lin, Chan, Nagano, Pan, De~Mello, Gallo,
  Guibas, Tremblay, Khamis, et~al.]{eg3d}
Eric~R Chan, Connor~Z Lin, Matthew~A Chan, Koki Nagano, Boxiao Pan, Shalini
  De~Mello, Orazio Gallo, Leonidas~J Guibas, Jonathan Tremblay, Sameh Khamis,
  et~al.
\newblock Efficient geometry-aware 3d generative adversarial networks.
\newblock In \emph{Proceedings of the IEEE/CVF conference on computer vision
  and pattern recognition}, pages 16123--16133, 2022.

\bibitem[Chang et~al.(2019)Chang, Lambert, Sangkloy, Singh, Bak, Hartnett,
  Wang, Carr, Lucey, Ramanan, et~al.]{argoverse}
Ming-Fang Chang, John Lambert, Patsorn Sangkloy, Jagjeet Singh, Slawomir Bak,
  Andrew Hartnett, De Wang, Peter Carr, Simon Lucey, Deva Ramanan, et~al.
\newblock Argoverse: 3d tracking and forecasting with rich maps.
\newblock In \emph{Proceedings of the IEEE/CVF conference on computer vision
  and pattern recognition}, pages 8748--8757, 2019.

\bibitem[Charatan et~al.(2024)Charatan, Li, Tagliasacchi, and
  Sitzmann]{pixelsplat}
David Charatan, Sizhe~Lester Li, Andrea Tagliasacchi, and Vincent Sitzmann.
\newblock pixelsplat: 3d gaussian splats from image pairs for scalable
  generalizable 3d reconstruction.
\newblock In \emph{Proceedings of the IEEE/CVF conference on computer vision
  and pattern recognition}, pages 19457--19467, 2024.

\bibitem[Chen et~al.(2024)Chen, Xu, Zheng, Zhuang, Pollefeys, Geiger, Cham, and
  Cai]{mvsplat}
Yuedong Chen, Haofei Xu, Chuanxia Zheng, Bohan Zhuang, Marc Pollefeys, Andreas
  Geiger, Tat-Jen Cham, and Jianfei Cai.
\newblock Mvsplat: Efficient 3d gaussian splatting from sparse multi-view
  images.
\newblock In \emph{European Conference on Computer Vision}, pages 370--386.
  Springer, 2024.

\bibitem[Chen et~al.(2025)Chen, Chen, Chen, Pons-Moll, and
  Xiu]{chen2025feat2gs}
Yue Chen, Xingyu Chen, Anpei Chen, Gerard Pons-Moll, and Yuliang Xiu.
\newblock Feat2gs: Probing visual foundation models with gaussian splatting.
\newblock In \emph{Proceedings of the Computer Vision and Pattern Recognition
  Conference}, pages 6348--6361, 2025.

\bibitem[Dosovitskiy et~al.(2017)Dosovitskiy, Ros, Codevilla, Lopez, and
  Koltun]{carla}
Alexey Dosovitskiy, German Ros, Felipe Codevilla, Antonio Lopez, and Vladlen
  Koltun.
\newblock Carla: An open urban driving simulator.
\newblock In \emph{Conference on robot learning}, pages 1--16. PMLR, 2017.

\bibitem[Fei et~al.(2024{\natexlab{a}})Fei, Zheng, Duan, Zhan, Tomizuka,
  Keutzer, and Lu]{driv3r}
Xin Fei, Wenzhao Zheng, Yueqi Duan, Wei Zhan, Masayoshi Tomizuka, Kurt Keutzer,
  and Jiwen Lu.
\newblock Driv3r: Learning dense 4d reconstruction for autonomous driving.
\newblock \emph{arXiv preprint arXiv:2412.06777}, 2024{\natexlab{a}}.

\bibitem[Fei et~al.(2024{\natexlab{b}})Fei, Zheng, Duan, Zhan, Tomizuka,
  Keutzer, and Lu]{pixelgaussian}
Xin Fei, Wenzhao Zheng, Yueqi Duan, Wei Zhan, Masayoshi Tomizuka, Kurt Keutzer,
  and Jiwen Lu.
\newblock Pixelgaussian: Generalizable 3d gaussian reconstruction from
  arbitrary views.
\newblock \emph{arXiv preprint arXiv:2410.18979}, 2024{\natexlab{b}}.

\bibitem[Fu et~al.(2024)Fu, Hamilton, Brandt, Feldman, Zhang, and
  Freeman]{featup}
Stephanie Fu, Mark Hamilton, Laura Brandt, Axel Feldman, Zhoutong Zhang, and
  William~T Freeman.
\newblock Featup: A model-agnostic framework for features at any resolution.
\newblock \emph{arXiv preprint arXiv:2403.10516}, 2024.

\bibitem[Gao et~al.(2024{\natexlab{a}})Gao, Chen, Li, Hong, Li, and
  Xu]{magicdrive3d}
Ruiyuan Gao, Kai Chen, Zhihao Li, Lanqing Hong, Zhenguo Li, and Qiang Xu.
\newblock Magicdrive3d: Controllable 3d generation for any-view rendering in
  street scenes.
\newblock \emph{arXiv preprint arXiv:2405.14475}, 2024{\natexlab{a}}.

\bibitem[Gao et~al.(2024{\natexlab{b}})Gao, Chen, Xiao, Hong, Li, and
  Xu]{magicdrivev2}
Ruiyuan Gao, Kai Chen, Bo Xiao, Lanqing Hong, Zhenguo Li, and Qiang Xu.
\newblock Magicdrive-v2: High-resolution long video generation for autonomous
  driving with adaptive control.
\newblock \emph{arXiv preprint arXiv:2411.13807}, 2024{\natexlab{b}}.

\bibitem[Gieruc et~al.(2024)Gieruc, K{\"a}stingsch{\"a}fer, Bernhard, and
  Salzmann]{6imgto3d}
Th{\'e}o Gieruc, Marius K{\"a}stingsch{\"a}fer, Sebastian Bernhard, and Mathieu
  Salzmann.
\newblock 6img-to-3d: Few-image large-scale outdoor driving scene
  reconstruction.
\newblock \emph{arXiv preprint arXiv:2404.12378}, 2024.

\bibitem[He et~al.(2016)He, Zhang, Ren, and Sun]{he2016deep}
Kaiming He, Xiangyu Zhang, Shaoqing Ren, and Jian Sun.
\newblock Deep residual learning for image recognition.
\newblock In \emph{Proceedings of the IEEE conference on computer vision and
  pattern recognition}, pages 770--778, 2016.

\bibitem[He et~al.(2022)He, Chen, Xie, Li, Doll\'ar, and Girshick]{mae}
Kaiming He, Xinlei Chen, Saining Xie, Yanghao Li, Piotr Doll\'ar, and Ross
  Girshick.
\newblock Masked autoencoders are scalable vision learners.
\newblock In \emph{Proceedings of the IEEE/CVF Conference on Computer Vision
  and Pattern Recognition (CVPR)}, pages 16000--16009, 2022.

\bibitem[Heinrich et~al.(2024)Heinrich, Ranzinger, Lu, Kautz, Tao, Catanzaro,
  Molchanov, et~al.]{Radio}
Greg Heinrich, Mike Ranzinger, Yao Lu, Jan Kautz, Andrew Tao, Bryan Catanzaro,
  Pavlo Molchanov, et~al.
\newblock Radio amplified: Improved baselines for agglomerative vision
  foundation models.
\newblock \emph{arXiv preprint arXiv:2412.07679}, 2024.

\bibitem[Hou et~al.(2024)Hou, Li, Guan, Zhang, Feng, Du, Xue, and Pu]{FastOcc}
Jiawei Hou, Xiaoyan Li, Wenhao Guan, Gang Zhang, Di Feng, Yuheng Du, Xiangyang
  Xue, and Jian Pu.
\newblock Fastocc: Accelerating 3d occupancy prediction by fusing the 2d
  bird’s-eye view and perspective view.
\newblock In \emph{2024 IEEE International Conference on Robotics and
  Automation (ICRA)}, pages 16425--16431, 2024.

\bibitem[Hu et~al.(2024)Hu, Yin, Zhang, Cai, Long, Chen, Wang, Yu, Shen, and
  Shen]{metric3dv2}
Mu Hu, Wei Yin, Chi Zhang, Zhipeng Cai, Xiaoxiao Long, Hao Chen, Kaixuan Wang,
  Gang Yu, Chunhua Shen, and Shaojie Shen.
\newblock Metric3d v2: A versatile monocular geometric foundation model for
  zero-shot metric depth and surface normal estimation.
\newblock \emph{IEEE Transactions on Pattern Analysis and Machine
  Intelligence}, 2024.

\bibitem[Huang et~al.(2023)Huang, Zheng, Zhang, Zhou, and Lu]{TPVFormer}
Yuanhui Huang, Wenzhao Zheng, Yunpeng Zhang, Jie Zhou, and Jiwen Lu.
\newblock Tri-perspective view for vision-based 3d semantic occupancy
  prediction.
\newblock In \emph{Proceedings of the IEEE/CVF conference on computer vision
  and pattern recognition}, pages 9223--9232, 2023.

\bibitem[Karras et~al.(2021)Karras, Laine, and Aila]{stylegan}
Tero Karras, Samuli Laine, and Timo Aila.
\newblock A style-based generator architecture for generative adversarial
  networks.
\newblock \emph{IEEE Transactions on Pattern Analysis and Machine
  Intelligence}, 43\penalty0 (12), 2021.

\bibitem[K{\"a}stingsch{\"a}fer et~al.(2025)K{\"a}stingsch{\"a}fer, Gieruc,
  Bernhard, Campbell, Insafutdinov, Najafli, and Brox]{seed4d}
Marius K{\"a}stingsch{\"a}fer, Th{\'e}o Gieruc, Sebastian Bernhard, Dylan
  Campbell, Eldar Insafutdinov, Eyvaz Najafli, and Thomas Brox.
\newblock Seed4d: A synthetic ego-exo dynamic 4d data generator, driving
  dataset and benchmark.
\newblock In \emph{2025 IEEE/CVF Winter Conference on Applications of Computer
  Vision (WACV)}, pages 7752--7764. IEEE, 2025.

\bibitem[Kerbl et~al.(2023)Kerbl, Kopanas, Leimkuehler, and Drettakis]{3DGS}
Bernhard Kerbl, Georgios Kopanas, Thomas Leimkuehler, and George Drettakis.
\newblock 3d gaussian splatting for real-time radiance field rendering.
\newblock \emph{ACM Trans. Graph.}, 42\penalty0 (4), 2023.

\bibitem[Lai et~al.(2024)Lai, Tian, Chen, Li, Yuan, Liu, and Jia]{LISA}
Xin Lai, Zhuotao Tian, Yukang Chen, Yanwei Li, Yuhui Yuan, Shu Liu, and Jiaya
  Jia.
\newblock Lisa: Reasoning segmentation via large language model.
\newblock In \emph{Proceedings of the IEEE/CVF Conference on Computer Vision
  and Pattern Recognition}, pages 9579--9589, 2024.

\bibitem[Li et~al.(2021)Li, Yu, Wang, Yang, Yu, and Scherer]{ULSD}
Hao Li, Huai Yu, Jinwang Wang, Wen Yang, Lei Yu, and Sebastian Scherer.
\newblock Ulsd: Unified line segment detection across pinhole, fisheye, and
  spherical cameras.
\newblock \emph{ISPRS Journal of Photogrammetry and Remote Sensing},
  178:\penalty0 187--202, 2021.

\bibitem[Li et~al.(2023)Li, Zhang, Zeng, Liu, Li, Ren, and Zhang]{dfa3d}
Hongyang Li, Hao Zhang, Zhaoyang Zeng, Shilong Liu, Feng Li, Tianhe Ren, and
  Lei Zhang.
\newblock Dfa3d: 3d deformable attention for 2d-to-3d feature lifting.
\newblock In \emph{Proceedings of the IEEE/CVF International Conference on
  Computer Vision}, pages 6684--6693, 2023.

\bibitem[Li et~al.(2024{\natexlab{a}})Li, Qiu, Cai, Yan, Lian, Liu, and
  Chen]{syntheocc}
Leheng Li, Weichao Qiu, Yingjie Cai, Xu Yan, Qing Lian, Bingbing Liu, and
  Ying-Cong Chen.
\newblock Syntheocc: Synthesize geometric-controlled street view images through
  3d semantic mpis.
\newblock \emph{arXiv preprint arXiv:2410.00337}, 2024{\natexlab{a}}.

\bibitem[Li et~al.(2024{\natexlab{b}})Li, Fang, Tombari, and Lee]{smilesplat}
Yanyan Li, Yixin Fang, Federico Tombari, and Gim~Hee Lee.
\newblock Smilesplat: Generalizable gaussian splats for unconstrained sparse
  images.
\newblock \emph{arXiv preprint arXiv:2411.18072}, 2024{\natexlab{b}}.

\bibitem[Li et~al.(2024{\natexlab{c}})Li, Wang, Li, Xie, Sima, Lu, Yu, and
  Dai]{bevformer}
Zhiqi Li, Wenhai Wang, Hongyang Li, Enze Xie, Chonghao Sima, Tong Lu, Qiao Yu,
  and Jifeng Dai.
\newblock Bevformer: learning bird's-eye-view representation from lidar-camera
  via spatiotemporal transformers.
\newblock \emph{IEEE Transactions on Pattern Analysis and Machine
  Intelligence}, 2024{\natexlab{c}}.

\bibitem[Liu et~al.(2024{\natexlab{a}})Liu, Sun, Wang, Wang, Sun, Ye, Zhang,
  and Duan]{reconx}
Fangfu Liu, Wenqiang Sun, Hanyang Wang, Yikai Wang, Haowen Sun, Junliang Ye,
  Jun Zhang, and Yueqi Duan.
\newblock Reconx: Reconstruct any scene from sparse views with video diffusion
  model.
\newblock \emph{arXiv preprint arXiv:2408.16767}, 2024{\natexlab{a}}.

\bibitem[Liu et~al.(2025{\natexlab{a}})Liu, Yu, Zou, Lyu, Mei, Chen, and
  Ma]{protocar}
Hongyuan Liu, Haochen Yu, Bochao Zou, Juntao Lyu, Qi Mei, Jiansheng Chen, and
  Huimin Ma.
\newblock Protocar: Learning 3d vehicle prototypes from single-view and
  unconstrained driving scene images.
\newblock \emph{Proceedings of the AAAI Conference on Artificial Intelligence},
  39\penalty0 (5):\penalty0 5460--5468, 2025{\natexlab{a}}.

\bibitem[Liu et~al.(2024{\natexlab{b}})Liu, Wang, Hu, Shen, Ye, Zang, Cao, Li,
  and Liu]{mvsgaussian}
Tianqi Liu, Guangcong Wang, Shoukang Hu, Liao Shen, Xinyi Ye, Yuhang Zang,
  Zhiguo Cao, Wei Li, and Ziwei Liu.
\newblock Mvsgaussian: Fast generalizable gaussian splatting reconstruction
  from multi-view stereo.
\newblock In \emph{European Conference on Computer Vision}, pages 37--53.
  Springer, 2024{\natexlab{b}}.

\bibitem[Liu et~al.(2025{\natexlab{b}})Liu, Fan, Yu, Li, Lu, and
  Yuan]{MonoSplat}
Yifan Liu, Keyu Fan, Weihao Yu, Chenxin Li, Hao Lu, and Yixuan Yuan.
\newblock Monosplat: Generalizable 3d gaussian splatting from monocular depth
  foundation models.
\newblock In \emph{Proceedings of the Computer Vision and Pattern Recognition
  Conference (CVPR)}, pages 21570--21579, 2025{\natexlab{b}}.

\bibitem[Loshchilov and Hutter(2019)]{AdamW}
Ilya Loshchilov and Frank Hutter.
\newblock Decoupled weight decay regularization.
\newblock In \emph{International Conference on Learning Representations}, 2019.

\bibitem[Mao et~al.(2021)Mao, Niu, Jiang, hanxue liang, Chen, Liang, Li, Ye,
  Zhang, Li, Yu, Xu, and Xu]{once}
Jiageng Mao, Minzhe Niu, Chenhan Jiang, hanxue liang, Jingheng Chen, Xiaodan
  Liang, Yamin Li, Chaoqiang Ye, Wei Zhang, Zhenguo Li, Jie Yu, Hang Xu, and
  Chunjing Xu.
\newblock One million scenes for autonomous driving: {ONCE} dataset.
\newblock In \emph{Thirty-fifth Conference on Neural Information Processing
  Systems Datasets and Benchmarks Track (Round 1)}, 2021.

\bibitem[Miao et~al.(2025)Miao, Huang, Bai, Yan, Zhou, Wang, Liu, Geiger, and
  Liao]{evolsplat}
Sheng Miao, Jiaxin Huang, Dongfeng Bai, Xu Yan, Hongyu Zhou, Yue Wang, Bingbing
  Liu, Andreas Geiger, and Yiyi Liao.
\newblock Evolsplat: Efficient volume-based gaussian splatting for urban view
  synthesis.
\newblock In \emph{Proceedings of the Computer Vision and Pattern Recognition
  Conference}, pages 11286--11296, 2025.

\bibitem[Mihajlovic et~al.(2024)Mihajlovic, Prokudin, Tang, Maier, Bogo, Tung,
  and Boyer]{splatfields}
Marko Mihajlovic, Sergey Prokudin, Siyu Tang, Robert Maier, Federica Bogo, Tony
  Tung, and Edmond Boyer.
\newblock Splatfields: Neural gaussian splats for sparse 3d and 4d
  reconstruction.
\newblock In \emph{European Conference on Computer Vision}, pages 313--332,
  2024.

\bibitem[Mildenhall et~al.(2021)Mildenhall, Srinivasan, Tancik, Barron,
  Ramamoorthi, and Ng]{nerf}
Ben Mildenhall, Pratul~P Srinivasan, Matthew Tancik, Jonathan~T Barron, Ravi
  Ramamoorthi, and Ren Ng.
\newblock Nerf: Representing scenes as neural radiance fields for view
  synthesis.
\newblock \emph{Communications of the ACM}, 65\penalty0 (1):\penalty0 99--106,
  2021.

\bibitem[Min et~al.(2024)Min, Luo, Sun, and Yang]{efreesplat}
Zhiyuan Min, Yawei Luo, Jianwen Sun, and Yi Yang.
\newblock Epipolar-free 3d gaussian splatting for generalizable novel view
  synthesis.
\newblock \emph{arXiv preprint arXiv:2410.22817}, 2024.

\bibitem[{New House Internet Services B.V.}(2023)]{PTGui}
{New House Internet Services B.V.}
\newblock Ptgui: Panorama stitching software.
\newblock \url{https://ptgui.com/}, 2023.
\newblock Version 12.29. Accessed on 2024-05-24.

\bibitem[Oquab et~al.(2024)Oquab, Darcet, Moutakanni, Vo, Szafraniec, Khalidov,
  Fernandez, HAZIZA, Massa, El-Nouby, Assran, Ballas, Galuba, Howes, Huang, Li,
  Misra, Rabbat, Sharma, Synnaeve, Xu, Jegou, Mairal, Labatut, Joulin, and
  Bojanowski]{dinov2}
Maxime Oquab, Timoth{\'e}e Darcet, Th{\'e}o Moutakanni, Huy~V. Vo, Marc
  Szafraniec, Vasil Khalidov, Pierre Fernandez, Daniel HAZIZA, Francisco Massa,
  Alaaeldin El-Nouby, Mido Assran, Nicolas Ballas, Wojciech Galuba, Russell
  Howes, Po-Yao Huang, Shang-Wen Li, Ishan Misra, Michael Rabbat, Vasu Sharma,
  Gabriel Synnaeve, Hu Xu, Herve Jegou, Julien Mairal, Patrick Labatut, Armand
  Joulin, and Piotr Bojanowski.
\newblock {DINO}v2: Learning robust visual features without supervision.
\newblock \emph{Transactions on Machine Learning Research}, 2024.
\newblock Featured Certification.

\bibitem[Ouyang et~al.(2023)Ouyang, He, Zhang, Luo, Guo, Zhan, and
  Huang]{EMAttention}
Daliang Ouyang, Su He, Guozhong Zhang, Mingzhu Luo, Huaiyong Guo, Jian Zhan,
  and Zhijie Huang.
\newblock Efficient multi-scale attention module with cross-spatial learning.
\newblock In \emph{ICASSP 2023-2023 IEEE International Conference on Acoustics,
  Speech and Signal Processing (ICASSP)}, pages 1--5. IEEE, 2023.

\bibitem[Philion and Fidler(2020)]{LSS}
Jonah Philion and Sanja Fidler.
\newblock Lift, splat, shoot: Encoding images from arbitrary camera rigs by
  implicitly unprojecting to 3d.
\newblock In \emph{Computer Vision--ECCV 2020: 16th European Conference,
  Glasgow, UK, August 23--28, 2020, Proceedings, Part XIV 16}, pages 194--210.
  Springer, 2020.

\bibitem[Reading et~al.(2021)Reading, Harakeh, Chae, and Waslander]{CaDDN}
Cody Reading, Ali Harakeh, Julia Chae, and Steven~L Waslander.
\newblock Categorical depth distribution network for monocular 3d object
  detection.
\newblock In \emph{Proceedings of the IEEE/CVF conference on computer vision
  and pattern recognition}, pages 8555--8564, 2021.

\bibitem[Ren et~al.(2024)Ren, Lu, hanxue liang, Wu, Ling, Chen, Fidler,
  Williams, and Huang]{scube}
Xuanchi Ren, Yifan Lu, hanxue liang, Jay~Zhangjie Wu, Huan Ling, Mike Chen,
  Sanja Fidler, Francis Williams, and Jiahui Huang.
\newblock {SC}ube: Instant large-scale scene reconstruction using voxsplats.
\newblock In \emph{The Thirty-eighth Annual Conference on Neural Information
  Processing Systems}, 2024.

\bibitem[Sim{\'e}oni et~al.(2025)Sim{\'e}oni, Vo, Seitzer, Baldassarre, Oquab,
  Jose, Khalidov, Szafraniec, Yi, Ramamonjisoa, et~al.]{dinov3}
Oriane Sim{\'e}oni, Huy~V Vo, Maximilian Seitzer, Federico Baldassarre, Maxime
  Oquab, Cijo Jose, Vasil Khalidov, Marc Szafraniec, Seungeun Yi, Micha{\"e}l
  Ramamonjisoa, et~al.
\newblock Dinov3.
\newblock \emph{arXiv preprint arXiv:2508.10104}, 2025.

\bibitem[Smart et~al.(2024)Smart, Zheng, Laina, and Prisacariu]{splatt3r}
Brandon Smart, Chuanxia Zheng, Iro Laina, and Victor~Adrian Prisacariu.
\newblock Splatt3r: Zero-shot gaussian splatting from uncalibrated image pairs.
\newblock \emph{arXiv preprint arXiv:2408.13912}, 2024.

\bibitem[Sun et~al.(2020)Sun, Kretzschmar, Dotiwalla, Chouard, Patnaik, Tsui,
  Guo, Zhou, Chai, Caine, et~al.]{waymo}
Pei Sun, Henrik Kretzschmar, Xerxes Dotiwalla, Aurelien Chouard, Vijaysai
  Patnaik, Paul Tsui, James Guo, Yin Zhou, Yuning Chai, Benjamin Caine, et~al.
\newblock Scalability in perception for autonomous driving: Waymo open dataset.
\newblock In \emph{Proceedings of the IEEE/CVF conference on computer vision
  and pattern recognition}, pages 2446--2454, 2020.

\bibitem[Szymanowicz et~al.(2024)Szymanowicz, Rupprecht, and
  Vedaldi]{szymanowicz2024splatter}
Stanislaw Szymanowicz, Chrisitian Rupprecht, and Andrea Vedaldi.
\newblock Splatter image: Ultra-fast single-view 3d reconstruction.
\newblock In \emph{Proceedings of the IEEE/CVF conference on computer vision
  and pattern recognition}, pages 10208--10217, 2024.

\bibitem[Tang et~al.(2025)Tang, Fan, Wang, Xu, Ranjan, Schwing, and
  Yan]{Mvdust3r}
Zhenggang Tang, Yuchen Fan, Dilin Wang, Hongyu Xu, Rakesh Ranjan, Alexander
  Schwing, and Zhicheng Yan.
\newblock Mv-dust3r+: Single-stage scene reconstruction from sparse views in 2
  seconds.
\newblock In \emph{Proceedings of the Computer Vision and Pattern Recognition
  Conference}, pages 5283--5293, 2025.

\bibitem[Tian et~al.(2025)Tian, Tan, Xie, and Ma]{drivingforward}
Qijian Tian, Xin Tan, Yuan Xie, and Lizhuang Ma.
\newblock Drivingforward: Feed-forward 3d gaussian splatting for driving scene
  reconstruction from flexible surround-view input.
\newblock In \emph{Proceedings of the AAAI Conference on Artificial
  Intelligence}, pages 7374--7382, 2025.

\bibitem[Wang et~al.(2025{\natexlab{a}})Wang, Chen, Karaev, Vedaldi, Rupprecht,
  and Novotny]{VGGT}
Jianyuan Wang, Minghao Chen, Nikita Karaev, Andrea Vedaldi, Christian
  Rupprecht, and David Novotny.
\newblock Vggt: Visual geometry grounded transformer.
\newblock In \emph{Proceedings of the Computer Vision and Pattern Recognition
  Conference}, pages 5294--5306, 2025{\natexlab{a}}.

\bibitem[Wang et~al.(2024{\natexlab{a}})Wang, Leroy, Cabon, Chidlovskii, and
  Revaud]{dust3r}
Shuzhe Wang, Vincent Leroy, Yohann Cabon, Boris Chidlovskii, and Jerome Revaud.
\newblock Dust3r: Geometric 3d vision made easy.
\newblock In \emph{Proceedings of the IEEE/CVF Conference on Computer Vision
  and Pattern Recognition}, pages 20697--20709, 2024{\natexlab{a}}.

\bibitem[Wang et~al.(2024{\natexlab{b}})Wang, Huang, Chen, and Lee]{freesplat}
Yunsong Wang, Tianxin Huang, Hanlin Chen, and Gim~Hee Lee.
\newblock Freesplat: Generalizable 3d gaussian splatting towards free view
  synthesis of indoor scenes.
\newblock In \emph{Advances in Neural Information Processing Systems}, pages
  107326--107349. Curran Associates, Inc., 2024{\natexlab{b}}.

\bibitem[Wang et~al.(2025{\natexlab{b}})Wang, Huang, Chen, and
  Lee]{freesplat++}
Yunsong Wang, Tianxin Huang, Hanlin Chen, and Gim~Hee Lee.
\newblock Freesplat++: Generalizable 3d gaussian splatting for efficient indoor
  scene reconstruction.
\newblock \emph{arXiv preprint arXiv:2503.22986}, 2025{\natexlab{b}}.

\bibitem[Wei et~al.(2025)Wei, Li, and Liu]{Omniscene}
Dongxu Wei, Zhiqi Li, and Peidong Liu.
\newblock Omni-scene: Omni-gaussian representation for ego-centric sparse-view
  scene reconstruction.
\newblock In \emph{Proceedings of the Computer Vision and Pattern Recognition
  Conference}, pages 22317--22327, 2025.

\bibitem[Wei et~al.(2023)Wei, Zhao, Zheng, Zhu, Zhou, and
  Lu]{wei2023surroundocc}
Yi Wei, Linqing Zhao, Wenzhao Zheng, Zheng Zhu, Jie Zhou, and Jiwen Lu.
\newblock Surroundocc: Multi-camera 3d occupancy prediction for autonomous
  driving.
\newblock In \emph{Proceedings of the IEEE/CVF International Conference on
  Computer Vision}, pages 21729--21740, 2023.

\bibitem[Xiao et~al.(2025)Xiao, Xu, Liang, and Kang]{gssplat}
Feng Xiao, Hongbin Xu, Wanlin Liang, and Wenxiong Kang.
\newblock Gssplat: Generalizable semantic gaussian splatting for novel-view
  synthesis in 3d scenes.
\newblock \emph{arXiv preprint arXiv:2505.04659}, 2025.

\bibitem[Xiao et~al.(2021)Xiao, Shao, Hao, Zhang, Chai, Jiao, Li, Wu, Sun,
  Jiang, et~al.]{pandaset}
Pengchuan Xiao, Zhenlei Shao, Steven Hao, Zishuo Zhang, Xiaolin Chai, Judy
  Jiao, Zesong Li, Jian Wu, Kai Sun, Kun Jiang, et~al.
\newblock Pandaset: Advanced sensor suite dataset for autonomous driving.
\newblock In \emph{2021 IEEE international intelligent transportation systems
  conference (ITSC)}, pages 3095--3101. IEEE, 2021.

\bibitem[Xie et~al.(2021)Xie, Wang, Yu, Anandkumar, Alvarez, and
  Luo]{segformer}
Enze Xie, Wenhai Wang, Zhiding Yu, Anima Anandkumar, Jose~M Alvarez, and Ping
  Luo.
\newblock Segformer: Simple and efficient design for semantic segmentation with
  transformers.
\newblock \emph{Advances in neural information processing systems},
  34:\penalty0 12077--12090, 2021.

\bibitem[Xiong et~al.(2023)Xiong, Muttukuru, Upadhyay, Chari, and
  Kadambi]{sparsegs}
Haolin Xiong, Sairisheek Muttukuru, Rishi Upadhyay, Pradyumna Chari, and Achuta
  Kadambi.
\newblock Sparsegs: Real-time 360 $\{$$\backslash$deg$\}$ sparse view synthesis
  using gaussian splatting.
\newblock \emph{arXiv preprint arXiv:2312.00206}, 2023.

\bibitem[Xu et~al.(2025)Xu, Peng, Wang, Blum, Barath, Geiger, and
  Pollefeys]{depthsplat}
Haofei Xu, Songyou Peng, Fangjinhua Wang, Hermann Blum, Daniel Barath, Andreas
  Geiger, and Marc Pollefeys.
\newblock Depthsplat: Connecting gaussian splatting and depth.
\newblock In \emph{Proceedings of the Computer Vision and Pattern Recognition
  Conference}, pages 16453--16463, 2025.

\bibitem[Yu et~al.(2024)Yu, Xing, Yuan, Hu, Li, Huang, Gao, Wong, Shan, and
  Tian]{viewcrafter}
Wangbo Yu, Jinbo Xing, Li Yuan, Wenbo Hu, Xiaoyu Li, Zhipeng Huang, Xiangjun
  Gao, Tien-Tsin Wong, Ying Shan, and Yonghong Tian.
\newblock Viewcrafter: Taming video diffusion models for high-fidelity novel
  view synthesis.
\newblock \emph{arXiv preprint arXiv:2409.02048}, 2024.

\bibitem[Yue et~al.(2024)Yue, Das, Engelmann, Tang, and Lenssen]{fit3d}
Yuanwen Yue, Anurag Das, Francis Engelmann, Siyu Tang, and Jan~Eric Lenssen.
\newblock Improving 2d feature representations by 3d-aware fine-tuning.
\newblock In \emph{European Conference on Computer Vision}, pages 57--74.
  Springer, 2024.

\bibitem[Zhang et~al.(2018)Zhang, Isola, Efros, Shechtman, and Wang]{lpips}
Richard Zhang, Phillip Isola, Alexei~A Efros, Eli Shechtman, and Oliver Wang.
\newblock The unreasonable effectiveness of deep features as a perceptual
  metric.
\newblock \emph{IEEE}, 2018.

\bibitem[Zhang et~al.(2025)Zhang, Wang, Xu, Xue, Rupprecht, Zhou, Shen, and
  Wetzstein]{Flare}
Shangzhan Zhang, Jianyuan Wang, Yinghao Xu, Nan Xue, Christian Rupprecht,
  Xiaowei Zhou, Yujun Shen, and Gordon Wetzstein.
\newblock Flare: Feed-forward geometry, appearance and camera estimation from
  uncalibrated sparse views.
\newblock In \emph{Proceedings of the Computer Vision and Pattern Recognition
  Conference}, pages 21936--21947, 2025.

\bibitem[Zheng et~al.(2025)Zheng, Jiang, He, Sun, Dong, Zhang, and Du]{nexusgs}
Yulong Zheng, Zicheng Jiang, Shengfeng He, Yandu Sun, Junyu Dong, Huaidong
  Zhang, and Yong Du.
\newblock Nexusgs: Sparse view synthesis with epipolar depth priors in 3d
  gaussian splatting.
\newblock In \emph{Proceedings of the Computer Vision and Pattern Recognition
  Conference}, pages 26800--26809, 2025.

\bibitem[Zou et~al.(2024)Zou, Yu, Guo, Li, Liang, Cao, and Zhang]{TGS}
Zi-Xin Zou, Zhipeng Yu, Yuan-Chen Guo, Yangguang Li, Ding Liang, Yan-Pei Cao,
  and Song-Hai Zhang.
\newblock Triplane meets gaussian splatting: Fast and generalizable single-view
  3d reconstruction with transformers.
\newblock In \emph{Proceedings of the IEEE/CVF conference on computer vision
  and pattern recognition}, pages 10324--10335, 2024.

\end{thebibliography}
}

\clearpage
\setcounter{page}{1}
\maketitlesupplementary

{\setstretch{1}
\begin{strip}
\startcontents[chapters]
\printcontents[chapters]
    {l}{1}
    {\setcounter{tocdepth}{5}}
\end{strip}
}

\clearpage
\newpage
\section{Preliminaries} \label{Preliminaries}
\subsection{3D Gaussian Splatting}
Representing 3D Gaussians as ellipsoids establishes an isomorphism between them. Consequently, a collection of 3D Gaussians can model arbitrary 3D geometry. Each Gaussian is parameterized by its spatial position (mean) $\mathbf{X}$, an anisotropic covariance matrix $\mathbf{\Sigma}$, an opacity $\mathbf{\alpha}$, and its spherical harmonics coefficients $\mathbf{c}$. The covariance matrix $\mathbf{\Sigma}$ determines the ellipsoid's geometry and is decomposed into a scaling matrix $\mathbf{S}$ and a rotation matrix $\mathbf{\Psi}$. As illustrated in \Cref{gs-1}, the final shape is formed by applying an axis-aligned scaling followed by a rotation.
\begin{equation} \label{gs-1}
    G(\mathbf{X}) = e^{-\frac{1}{2}(\mathbf{X})^T \mathbf{\Sigma}^{-1} (\mathbf{X})}, where \mathbf{\Sigma} = \mathbf{\Psi} \mathbf{S}\mathbf{S}^T \mathbf{\Psi}^T
\end{equation}
The ellipsoids are rendered onto 2D images using a fast rasterization pipeline that projects each ellipsoid onto the image plane. The 2D covariance for each projection is computed using the viewing transformation $\mathbf{W}$ and the Jacobian $\mathbf{J}$ of the projective transformation, as detailed in \Cref{gs-3}.
\begin{equation}
    \bar{\mathbf{\Sigma}} = \mathbf{J}\mathbf{W} \mathbf{\Sigma} \mathbf{W}^T \mathbf{J}^T \label{gs-3}
\end{equation}
The final pixel color is synthesized via the alpha compositing technique from \Cref{gs-4}. This involves blending the contributions of $N$ Gaussians that overlap the pixel. These Gaussians are first sorted by depth, and for the $\mathbf{i}$-th Gaussian, its color $\mathbf{c_i}$ is evaluated from its Spherical Harmonics (SH) coefficients according to the viewing direction. The colors are then blended in front-to-back order to yield the final pixel value.
\begin{equation}
    C=\sum_{\mathbf{i}\in N}\mathbf{c_{i}}\mathbf{\alpha_{i}}\prod_{\mathbf{j}=1}^{\mathbf{i}-1}(1-\mathbf{\alpha_{j}}), \label{gs-4}
\end{equation}

\section{Detailed Related Work} \label{relate_detail}

\noindent
\subsection{Feed-forward Reconstruction Models}Feed-forward reconstruction methods can be broadly categorized based on their underlying 3D scene representation. The first category comprises pixel-based models~\citep{dust3r,Mvdust3r,VGGT,pixelsplat,mvsplat,depthsplat,Flare,driv3r,evolsplat,mvsgaussian,nexusgs,pixelgaussian,freesplat,freesplat++,MonoSplat,gssplat,smilesplat,efreesplat,splatt3r}, which typically reconstruct scenes as explicit point clouds or, more recently, Gaussian splats. These approaches operate by predicting per-pixel attributes like depth and local offsets, enabling them to generate dense and detailed geometry, particularly in forward-facing scenarios with substantial view overlap. However, their reliance on direct 2D-to-3D projection makes them vulnerable in challenging outward-facing settings, such as autonomous driving. In these cases, they often produce reconstructions characterized by voids in unobserved regions and struggle to reason about complex occlusions.
In contrast, volumetric methods~\citep{6imgto3d,scube,TGS,protocar} represent the scene implicitly within a continuous or discrete volume (e.g., a neural radiance field or a voxel grid). This inherent volumetric nature ensures the generation of topologically complete and hole-free geometry, offering a natural mechanism for handling occlusions. The trade-off, however, is that these representations often yield overly smooth surfaces, struggling to capture the high-frequency textural and geometric details at which pixel-based methods excel.
To harness the complementary strengths of both paradigms, hybrid methods like Omni-Scene~\citep{Omniscene} have emerged. These approaches aim to integrate the detail-rich output of pixel-based techniques with the completeness of volumetric representations. Our work aligns with this hybrid philosophy but introduces a key architectural divergence. Whereas Omni-Scene employs a pixel-guided strategy in which one representation heavily influences the other, our model utilizes independent, parallel prediction branches for each representation. This decoupled design allows each branch to specialize, culminating in a final reconstruction that is more geometrically complete and more finely detailed than previous methods.

\subsection{2D Feature Lifting}The task of lifting 2D image features into a 3D spatial representation is a cornerstone of modern 3D perception. Pioneering works in this area often rely on explicit depth estimation. For instance, CaDDN~\citep{CaDDN} generates a categorical depth distribution for each pixel to cast 2D features into a 3D volume. Similarly, LSS~\citep{LSS} generates a BEV map by splatting per-pixel features into a grid, guided by a predicted continuous depth distribution.
More recent paradigms have shifted towards attention mechanisms to establish this 2D-to-3D correspondence. BEVFormer~\citep{bevformer} pioneered the use of spatial cross-attention to query and aggregate multi-camera 2D features into a unified BEV representation. TPVFormer~\citep{TPVFormer} extends this concept by constructing a more comprehensive Tri-Perspective View (TPV) representation, populating three orthogonal planes from the 2D inputs. Further advancing this line of work, DFA3D~\citep{dfa3d} integrates depth-aware sampling into a 3D deformable attention mechanism for more precise feature lifting.
Diverging from these approaches, which learn to lift features via explicit depth prediction or complex attention queries, our method introduces a novel, principled restructuring technique. We leverage the inherent spatial coherence within 2D feature maps, reordering the channel dimensions to directly form a semantically rich pillar-based representation. This pillar field serves as an effective and efficient intermediate representation for downstream 3D tasks.

\begin{figure*}[htb]
	\centering
	\includegraphics[width=\linewidth]{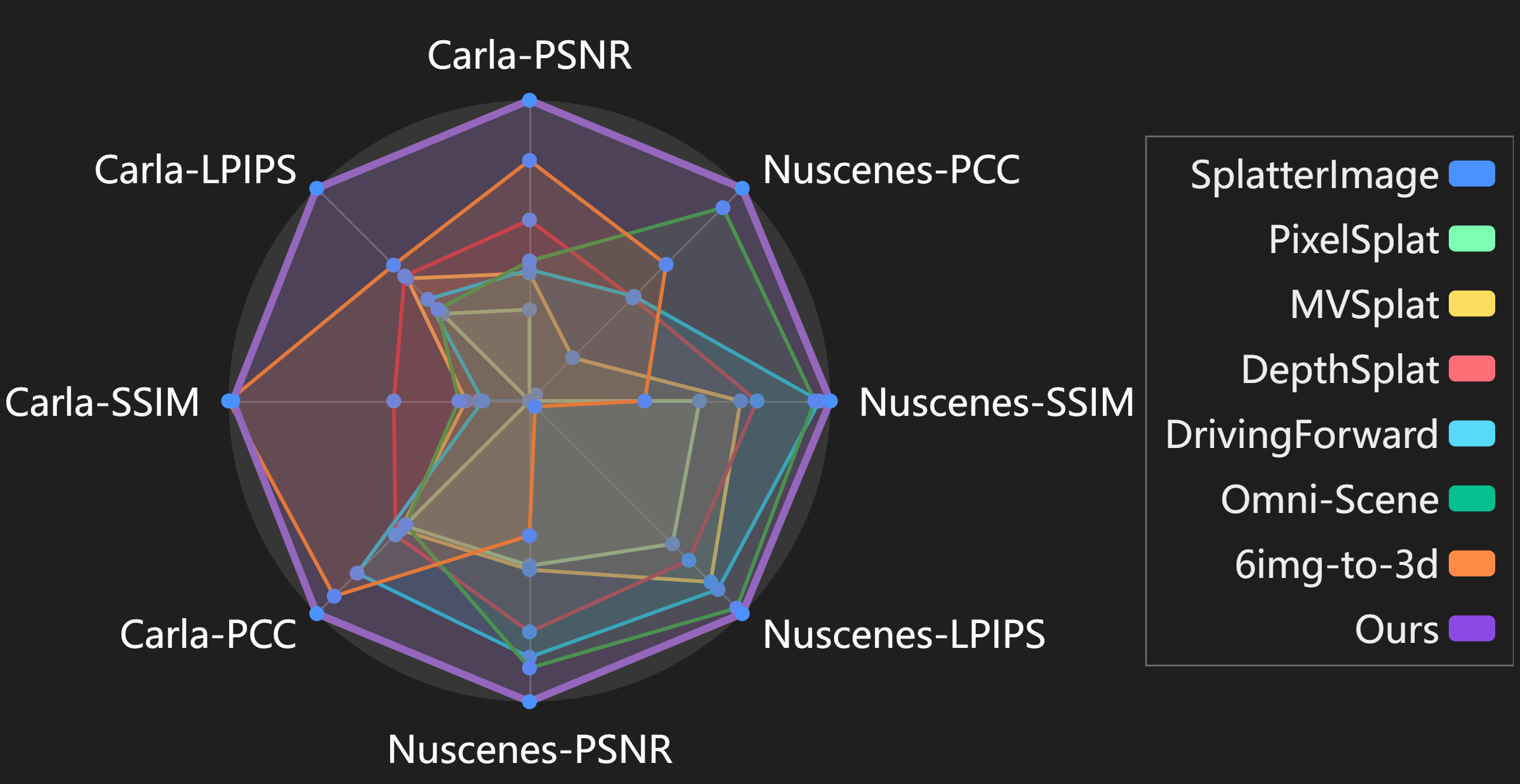}
	\caption{\textbf{Comparison of model performance via a radar chart.} The area of the radar chart represents the comprehensive performance of each model. A larger enclosed area indicates a stronger capability to unify the two experimental settings.} 
	\label{radar}
\end{figure*}

\section{Task Settings} \label{tasksettings_sec}
\begin{figure}[htb]
	\centering
\includegraphics[width=\linewidth]{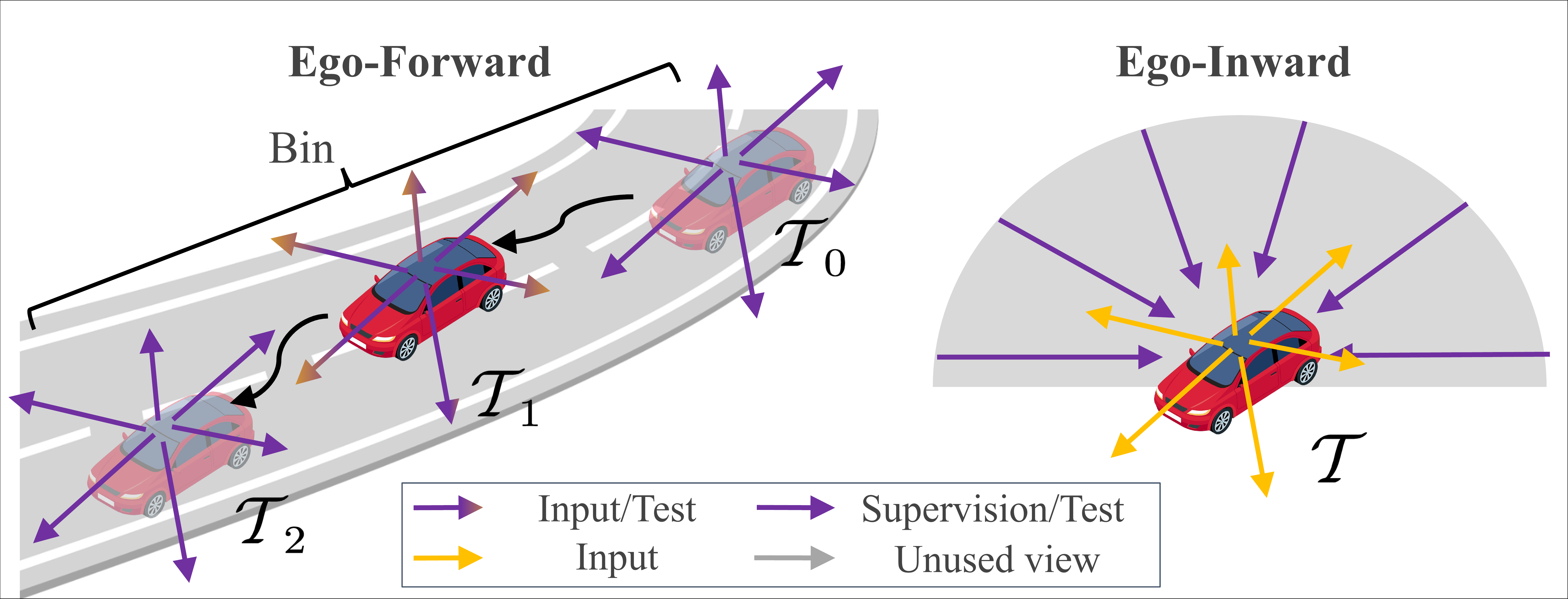}
	\caption{\textbf{Two reconstruction settings for comprehensive assessment.} We introduce two tasks for autonomous driving scene reconstruction: the Ego-Forward setting, evaluated on the nuScenes dataset, and the Ego-Inward setting, evaluated on a custom Carla-Centric dataset. These settings are designed to probe different model capabilities and serve distinct downstream applications. The Ego-Forward task targets common driving scenarios, producing assets suitable for testing and simulating perception algorithms. In contrast, the Ego-Inward task generates a comprehensive, omnidirectional scene representation. This holistic view is particularly valuable for applications involving heterogeneous data sources, such as Vehicle-to-Everything (V2X) collaboration and air-to-ground joint perception.
} 
	\label{tasksettings}
\end{figure}

The ego-forward setting utilizes forward-facing sequences from autonomous driving scenarios to assess generalizable reconstruction from sparse views, thereby measuring a model's potential of perception in standard driving situations. In contrast, the ego-inward setting focuses on the scene itself, evaluating performance under large viewpoint variations. This latter test probes a model's robustness, its ability to handle heterogeneous data sources, and its capacity for enhancing perception features across wide-span views.

As illustrated in \Cref{tasksettings}, we define two distinct task settings. The ego-forward setting evaluates forward-facing sequence reconstruction: the model processes six input views from time $\mathcal{T}_1$ and is supervised using ground-truth views from a temporal window spanning $\mathcal{T}_0$, $\mathcal{T}_1$, and $\mathcal{T}_2$. In contrast, the ego-inward setting assesses spatial generalization. Given the same six views from time $\mathcal{T}$, the model is evaluated on its ability to synthesize novel views from a surrounding hemisphere at that identical timestep.

The radar chart in \Cref{radar} illustrates the strong unification capability of our model, where a larger area signifies superior comprehensive performance. It demonstrates that our model achieves excellent overall results on two reconstruction tasks under different settings.

\begin{figure*}[t]
	\centering
	\includegraphics[width=\linewidth]{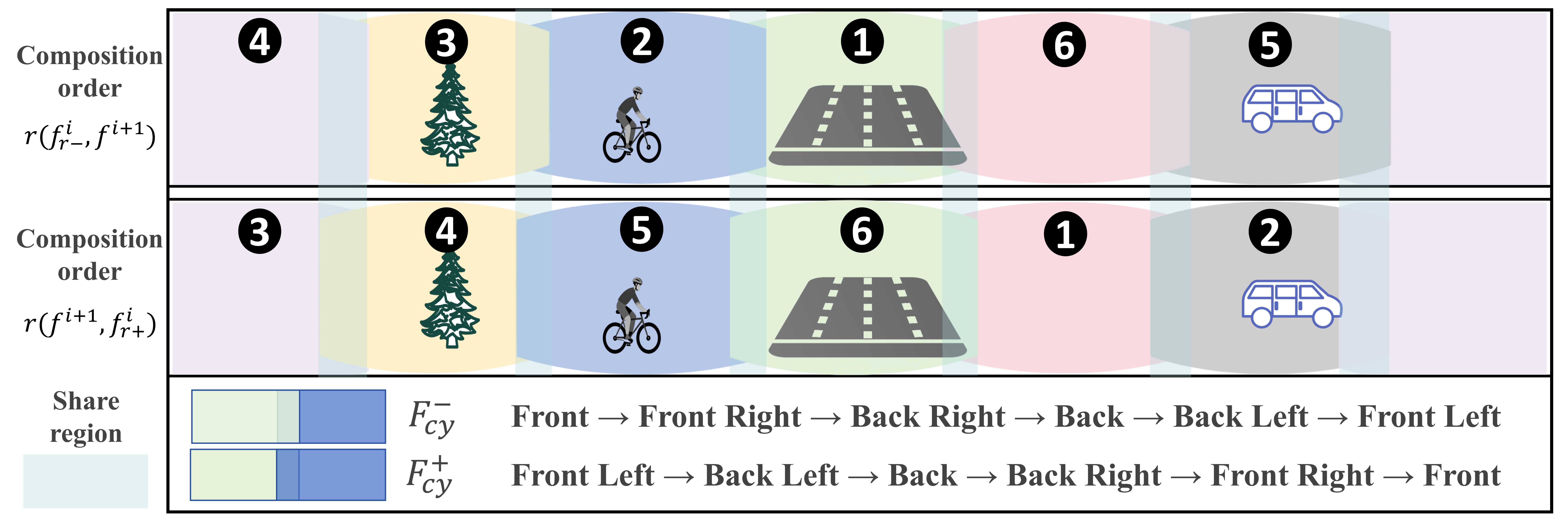}
	\caption{\textbf{Effectiveness of directional composition.} By employing clockwise manner and counterclockwise manner, the covered area constitutes only a small fraction of the six images.}
	\label{com_order}
\end{figure*}

\section{Discussion} \label{dis}
\subsection{Inherent Advantages of Our Approach}Unlike previous approaches such as Omni-Scene \citep{Omniscene} and 6Img-to-3D \citep{6imgto3d}, we introduce a more interpretable and parameter-free Unified Cylinder Camera Model (UCCM) for handling camera transformations. This design eliminates the need for complex attention mechanisms to learn large angular shifts between views. The network architecture operates on the Cylinder Plane Feature, making it inherently compatible with established 2D feature enhancement techniques, such as the patching strategy from Masked Autoencoders \citep{mae}.
Furthermore, visualizations of the Cylinder Plane Feature reveal that while projection artifacts do create black borders, these regions are proportionally small. This characteristic allows us to frame the task as a feature inpainting problem, drawing a direct parallel to the MAE framework. In essence, our model functions as an autoencoder. During supervised training, it learns to progressively fill these border regions with meaningful semantic content, a process that simultaneously strengthens the representation of the learned features.

Owing to the effective design of the UCCM, our model exhibits two primary advantages. First, it demonstrates strong zero-shot generalization, enabling compatibility with surround-view autonomous driving data from various camera types without requiring specific pre-training. Furthermore, it can be adapted into a customized model by fine-tuning on just a small number of samples. Second, the unified architecture supports joint training on diverse data sources, including various autonomous driving and panoramic datasets. This positions our model with the potential to evolve into a foundational reconstruction model for driving scenes.

\subsection{Limitations}
While our model currently supports automotive-grade surround-view cameras and panoramic images, our experiments reveal a performance degradation on indoor scenes. This is attributable to a bias in the training data. Furthermore, the model's output quality deteriorates when handling extreme perspectives, such as aerial views characterized by distant and densely packed buildings.

The parameters $\rho$ and $\delta h$ for the UCCM are currently set heuristically, as our model does not yet incorporate a mechanism for automatic inference or adjustment. While setting these values to 1 and 0 respectively yields favorable results, fine-tuning them can enhance the zero-shot performance to some extent.

\subsection{Application}Fundamentally, our model is designed as a direct 2D-to-3D lifting framework. This architectural choice provides a straightforward yet powerful pathway to elevate existing 2D autonomous driving scene generation methods into the 3D domain. For instance, state-of-the-art 2D generation models like MagicDrive-v2 \citep{magicdrivev2}, which excel at producing diverse and realistic driving scenarios, could be seamlessly integrated with our approach. By leveraging their powerful 2D backbones and applying our lifting module, we can directly generate high-fidelity, geometrically consistent 3D scenes without the need to retrain a large-scale 3D generator from scratch. This not only democratizes 3D scene generation but also significantly accelerates the development cycle. Beyond its application in generation, our work pioneers a new paradigm for 2D-to-3D lifting in the context of autonomous driving. By re-framing the geometric transformation in a more interpretable and efficient manner, we unlock latent potential for various downstream perception tasks. For example, our method could enhance 3D object detection by providing richer geometric cues or improve BEV (Bird's-Eye-View) segmentation by warping 2D features into a more robust and spatially aware representation. We believe this novel approach opens up promising new avenues for research, offering a flexible and effective bridge between the mature 2D vision ecosystem and the burgeoning field of 3D autonomous driving perception.

\subsection{Future Work}Our proposed framework opens up several promising avenues for future research. We outline four key directions below:

a) \textit{Extension to 4D Reconstruction for Dynamic Objects}: Our current model primarily focuses on static scenes. A natural and critical next step is to extend our 2D-to-3D lifting paradigm to the temporal domain, enabling 4D reconstruction. This would involve incorporating temporal cues to model the motion and deformation of dynamic agents, such as vehicles and pedestrians.

b) \textit{End-to-End Semantic and Instance Segmentation in 3D}: While our method generates geometrically accurate 3D structures, integrating rich semantic understanding is crucial. We plan to explore end-to-end architectures that jointly perform 3D reconstruction and semantic/instance segmentation. Our interpretable feature-lifting mechanism provides a strong foundation for this, as the lifted 2D semantic features can be directly supervised in the 3D space, potentially leading to more accurate and consistent segmentation in complex urban environments.

c) \textit{Generalizable Material and Texture Reconstruction}: To enhance the realism of generated scenes for simulation and data augmentation, we aim to reconstruct not only geometry but also surface materials and textures. This involves predicting spatially varying Bidirectional Reflectance Distribution Functions (BRDFs) or other appearance models. Our framework's ability to handle complex projective geometry could be adapted to disentangle material properties from illumination effects, paving the way for high-fidelity, relightable 3D asset creation for autonomous systems.

d) \textit{Large-Scale Scene Generation}: This direction involves investigating method for stitching the outputs of multiple, spatially-distinct predictive modules to generate large-scale, cohesive 3D environments. This requires ensuring geometric, semantic, and instance-level consistency across the boundaries of the individual predictions.

\section{Definition and Derivation} \label{proofs}

\subsection{The Overlaying Procedure} \label{overlaying}
In an ideal scenario where the $N$ cameras share the same location and intrinsic parameters, and only differ in the poses, the FoV would perfectly match the vertical angle required for cylindrical projection. Unfortunately, there exist inevitable camera-to-camera and camera-to-central axis offsets, leaving the top and bottom regions in the cylinder plane unprojected and introducing boundary artifacts. To mitigate this issue, A small height offset is applied to the central point $\mathbf{c}_{u}^x$ using $\mathbf{c}_{u}^x=\sum_{n=0}^{N-1}\mathbf{c}_n^e / N + (0, 0, \Delta h)^T$ and the effective vertical angle is set to the minimum FoV $f_{min} \in \mathbb{R}$ of these $N$ views, which is further multiplied by a factor of $\rho$, thereby mitigating the unprojected regions. Then, the radius is calculated by $R_u=\frac{Z_u / 2}{\tan{(\rho \cdot f_{min}/2)}}$. 

Note that the parameters $\Delta h$ and $\rho$ are adjustable and are set to different values for occupancy-aware, volume-aware and pixel-aware modules, as well as different camera configurations, which means that the cylinder plane features are slightly different for different modules and different camera configureations. 

\begin{definition}
 The ordered composition operators $\gamma(\mathbf{f}_a, \mathbf{f}_b)$ are defined by the following equations:
 \begin{equation} \label{eq-22}
  \gamma(\mathbf{f}_a, \mathbf{f}_b)=\begin{cases}
      \mathbf{f}_b, & \mathbf{f}_b\neq0\\
      \mathbf{f}_a, & \mathbf{f}_b=0
  \end{cases}
\end{equation}
\end{definition}
\Cref{com_order} provides a schematic illustration of the ordered composition operators applied to autonomous driving data.

A homogeneous point $\mathbf{X}_{cy}=(R_u, \theta, z, 1)^T \in \mathbb{R}^4$ on the cylinder plane is represented in the Cartesian coordinate system as $(x, y, z, 1)^T$.
Acquiring image features for the $\mathbf{i}$-th discrete points on the cylinder requires projecting them back onto the camera pixel coordinates $\mathbf{X}_{uv}^i \in \mathbb{R}^4$. 
 Using the normalized projected 2D coordinates $\hat{\mathbf{X}}_{uv}^{\mathbf{i}}=(2x_{uv}^\mathbf{i}/z_{uv}^\mathbf{i}-1,2y_{uv}^\mathbf{i}/z_{uv}^\mathbf{i}-1) \in \mathbb{R}^2$, we sample the corresponding $\mathbf{i}$-th image feature $\mathbf{F}_{\mathbf{i}} \in \mathbb{R}^{D_{feat}}$ obtained by Radio-v2.5 \citep{Radio} via bilinear interpolation to obtain the cylindrical point feature $\mathbf{f}_{uv}^\mathbf{i} \in \mathbb{R}^{D_{feat}}$ using $\mathbf{f}_{uv}^\mathbf{i} = bilinear(\mathbf{F}_\mathbf{i}, \hat{\mathbf{X}}_{uv}^\mathbf{i})$.
 
 The features of $\mathbf{X}_{cy}$ are constructed via an ordered composition process which is governed by \Cref{eq-8} where $r(\cdot,\cdot)$ is defined above. For $i=0,\dots,N-2$, we have:

\begin{equation} \label{eq-8}
\begin{aligned}
  &\mathbf{f}_{r-}^{\mathbf{i}+1} = \gamma(\mathbf{f}_{r-}^\mathbf{i}, \mathbf{f}^{\mathbf{i}+1}), \quad \mathbf{f}_{r-}^0 = \mathbf{f}_{uv}^0, \quad \\
  &\mathbf{f}_{r+}^{\mathbf{i}+1} = \gamma(\mathbf{f}^{\mathbf{i}+1}, \mathbf{f}_{r+}^\mathbf{i}), \quad \mathbf{f}_{r+}^0 = \mathbf{f}_{uv}^0
\end{aligned}
\end{equation}

The ordered composition process is executed using two distinct traversal orders: clockwise ($+$) and counter-clockwise ($-$), as depicted in \Cref{com_order}. For any given point on the Cylinder Plane, these two traversals independently produce final feature vectors, denoted $\mathbf{f}_{r+} = \mathbf{f}_{r+}^{N-1} \in \mathbb{R}^{D_{feat}}$ and $\mathbf{f}_{r-} = \mathbf{f}_{r-}^{N-1} \in \mathbb{R}^{D_{feat}}$, respectively. Applying this process across the entire plane generates two complete feature maps: $\mathbf{F}_{cy}^{+} \in \mathbb{R}^{D_{feat}\times H_u \times W_u}$ and $\mathbf{F}_{cy}^{-} \in \mathbb{R}^{D_{feat}\times H_u \times W_u}$.

\label{sample_cpfg}
\subsection{Sampling in Cylinder Plane Feature Group}
\begin{figure}[htb]
	\centering
	\begin{minipage}{0.45\linewidth}
		\centering
		\includegraphics[width=\linewidth]{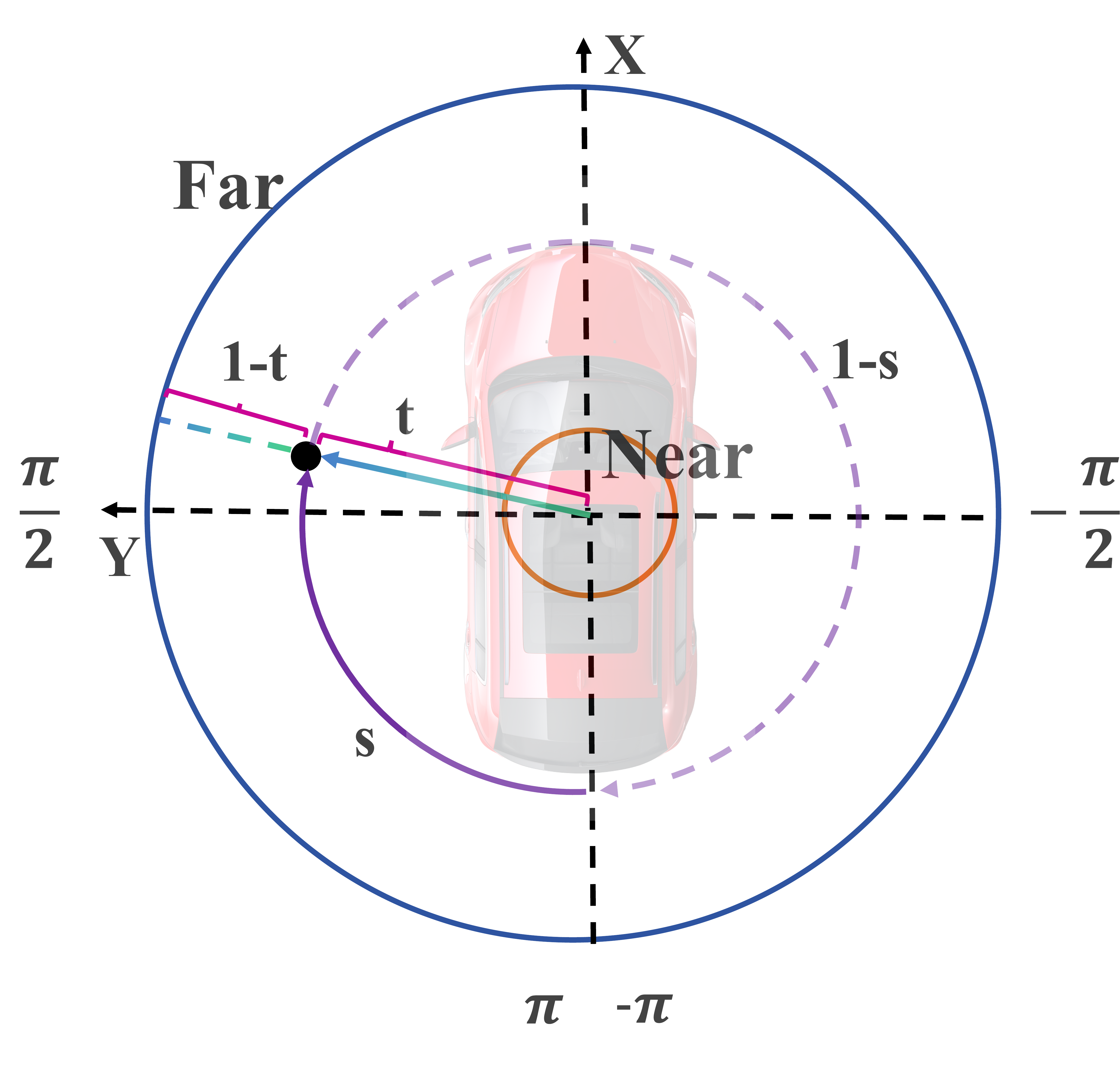}
	    \label{sample1}
	\end{minipage} \hfill
	\begin{minipage}{0.53\linewidth}
		\centering
		\includegraphics[width=\linewidth]{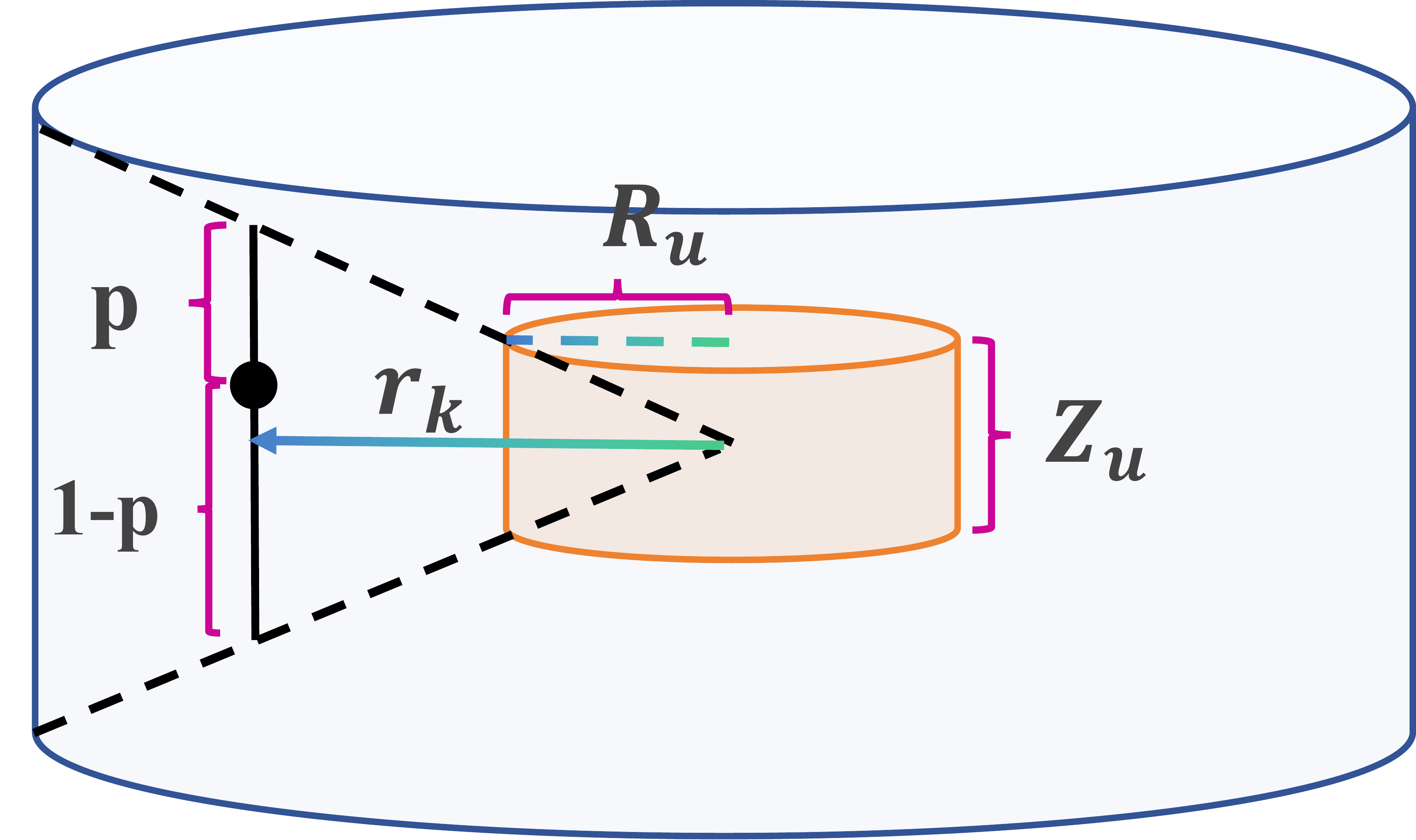}
		\label{sample2}
	\end{minipage}
 \caption{Proportional sampling in the radial, angular and vertical directions.} 
\end{figure}

For any point $\mathbf{X}_{o}[;k]=(r_k,\theta_k,z_k)$ within CPFG space with a maximum radius of $R_{max}$ and a minimum radius of $R_{min}$, the geometric relationship is illustrated in \Cref{sample1} and \Cref{sample2}. Within the annular cylindrical region bounded by a near radius $R_{min}$ and a far radius $R_{max}$, the normalized ratios for the radial component $t_k$ and the normalized angular component $s_k$ are computed as follows:
\begin{equation} \label{eq-25}
    t_k = \frac{r_k-R_{min}}{R_{max}-R_{min}}, \quad s_k=\frac{\pi-\theta_k}{2\pi}
\end{equation}
The normalization ratio for the height component $p_k$ is linearly dependent on the radial distance $r_k$. This is because the volume is a frustum, where the floor and ceiling heights change linearly with the radius. Based on the principle of similar triangles, we can derive $p_k$ as follows:
\begin{equation} \label{eq-26}
    \frac{r_k}{R_u}=\frac{Z_{cur}}{Z_u}, \quad \frac{(\frac{Z_{cur}}{2}+z_k)}{Z_{cur}} = 1 - p_k
\end{equation}
Substituting and rearranging the terms, we obtain:
\begin{equation} \label{eq-27}
\begin{aligned}
    p_k = &1 - \frac{(\frac{Z_{cur}}{2}+z_k)}{Z_{cur}} = 1 - (\frac{1}{2}+\frac{z_k}{Z_{cur}})\\
    &=\frac{1}{2}-\frac{z_k}{Z_{cur}}=\frac{1}{2}-\frac{z_kR_u}{Z_ur_k}
\end{aligned}
\end{equation}
To obtain the final normalized coordinates for CPFG sampling, the component ratios are rescaled from the [0, 1] interval to the canonical range of [-1, 1]:
\begin{equation} \label{eq-28}
\begin{aligned}
    \mathbf{X}_{n}^k = (2t_k-1,2s_k-1,2p_k-1)^T \\
    = (\frac{2(r_k-r_{min})}{r_{max}-r_{min}}-1, \frac{-\theta_k}{\pi}, \frac{-2 z_k R_u}{Z_u r})^T
\end{aligned}
\end{equation}
Then, we can obtain the occ feature of $\mathbf{X}_{o}[;k]$ using the following equation:
\begin{equation} \label{eq-10}
    \quad \mathbf{f}_{occ}^k = \text{Trilinear}(\mathbf{F}^{occ}, \mathbf{X}_{o}[;k])
\end{equation}
Performing the above operation on each point, we can obtain $\mathbf{F}_{cy}^{occ}$ for the occupancy grid $\mathbf{O}$.

\label{sample_bg}
\subsection{Sampling Background Feature on the Cylinder Plane Feature}In our cylindrical projection-based background rendering method, we need to compute ray direction vectors from the camera viewpoint for each pixel in the image. This process involves geometric transformations from pixel coordinates to the world coordinate system.

First, we transform pixel coordinates $(\mathbf{i}, \mathbf{j})$ in the image to normalized camera coordinates. Given the camera intrinsic matrix:
\begin{equation}
K = \begin{bmatrix}
f_x & 0 & c_x \\
0 & f_y & c_y \\
0 & 0 & 1
\end{bmatrix}
\end{equation}
where $f_x, f_y$ represent the focal lengths in the $x$ and $y$ directions respectively, and $(c_x, c_y)$ denotes the principal point coordinates. For any pixel position $(\mathbf{i}, \mathbf{j})$ in the image, its normalized coordinates $(x_c, y_c, 1)$ in the camera coordinate system can be computed as:
\begin{align}
x_c &= \frac{\mathbf{i} - c_x + 0.5}{f_x}, \quad y_c = \frac{\mathbf{j} - c_y + 0.5}{f_y}, \quad z_c = 1
\end{align}
The $+0.5$ offset is used to convert pixel coordinates from the top-left origin to the pixel center, ensuring that rays emanate from the center of each pixel.

Next, we transform the direction vector $\mathbf{d}c = (x_c, y_c, 1)$ from the camera coordinate system to the world coordinate system. This transformation is achieved through the camera's rotation matrix $\mathbf{R}$.

Given the camera extrinsic matrix (camera-to-world transformation matrix):
\begin{equation}
\mathbf{T}_{c2w} = \begin{bmatrix}
\mathbf{R} & \mathbf{t} \\
\mathbf{0}^T & 1
\end{bmatrix}
\end{equation}
where $\mathbf{R}$ is a $3 \times 3$ orthogonal rotation matrix and $\mathbf{t}$ is the translation vector. The direction vector in the camera coordinate system is transformed to the world coordinate system as:
\begin{equation}
\mathbf{d}_w = \mathbf{R} \cdot \mathbf{d}_c = \mathbf{R} \cdot \begin{bmatrix} x_c \\ y_c \\ 1 \end{bmatrix}
\end{equation}
Since the direction vector only represents direction, we typically normalize it to a unit vector:
\begin{equation}
\mathbf{d} = \frac{\mathbf{d}_w}{\|\mathbf{d}_w\|} = (d_x, d_y, d_z)^T
\end{equation}
Finally, for camera position $\mathbf{p}=(x_0, y_0, z_0)$ and normalized direction vector $\mathbf{d}$, the ray can be parameterized as:
\begin{equation}
\mathbf{P}(t) = \mathbf{p} + t\mathbf{d}, \quad t \geq 0
\end{equation}
where parameter $t$ represents the distance along the ray direction. This parameterized ray equation will be used for subsequent intersection calculations with the background cylinder to determine the background texture sampling location for each pixel.

For a Cylinder Plane with radius $R_u$, the intersection point $\mathbf{P}(t) = \mathbf{p} + t\mathbf{d}$ must satisfy the following constraint:
\begin{equation}
(\mathbf{p} + t\mathbf{d})^2_x+(\mathbf{p} + t\mathbf{d})^2_y = R_u^2
\end{equation}
This can be rearranged into a linear equation with two variables:
\begin{equation}
t^2(d_x^2+d_y^2) + 2t(x_0d_x+y_0d_y)+(x_0^2+y_0^2-R_u^2)=0
\end{equation}
Solving for $t$, we get:
\begin{multline} \label{eq:multline_example}
\hat{t} = \frac{-(x_0d_x+y_0d_y)}{d_x^2+d_y^2} \\
+ \frac{\sqrt{(x_0d_x+y_0d_y)^2-(d_x^2+d_y^2)(x_0^2+y_0^2-R_u^2)}}{d_x^2+d_y^2}
\end{multline}
The intersection point is therefore given by the coordinates:
\begin{equation}
P_{intersect} = \mathbf{p}+\hat{t}\cdot \mathbf{d} = (p_x, p_y, p_z)^T
\end{equation}
To convert Cartesian coordinates to cylindrical coordinates, use the following formulas:
\begin{equation}
\alpha = \sqrt{p_x^2+p_y^2}, \quad \theta = \arctan({p_y/p_x}), \quad z=p_z
\end{equation}
To generate sampling points suitable for interpolation, the cylindrical coordinates must be normalized. Since $\theta \in [-\pi, \pi), \ z \in [-\frac{Z_u}{2}, \frac{Z_u}{2}]$, the mapping to a normalized space is performed as follows:
\begin{equation}
u = -\frac{\theta}{\pi}, \quad v = \frac{-z}{Z_u}
\end{equation}
The normalized sampling coordinates for bilinear interpolation are therefore given by:
\begin{equation}\label{eq-37}
\mathbf{P}_{intersect} = (u, v)^T
\end{equation}

\label{PPF}
\subsection{Pixel Projection Features} 
Drawing inspiration from the feature projection method in Omni-Scene \citep{Omniscene}, we propose a novel projection paradigm for the CPFG representation. In contrast to Omni-Scene, which projects pixel features directly onto TPV \citep{TPVFormer} planes, our approach projects each pixel feature to its two nearest neighboring CPFG planes based on its spatial location. These projected features are then interpolated and fused using inverse distance weighting.

We transform the $k$-th pixel Gaussian centers with pixel features $\mathbf{f}^k_{pix}$, denoted as $p^k_{\text{pixel}}$, from the pixel branch into the CPFG space. For any point $\mathbf{X}_{k}=(r_k,\theta_k,z_k)$ within CPFG space, we compute the normalized sampling coordinates $(t_k, s_k, p_k)$ according to \Cref{eq-25} and \Cref{eq-27}. Subsequently, using the radial component $t_k$ and the total number of CPFG planes, $K$, we identify the adjacent planes enclosing the sampling point. Specifically, the point is located between the CPFG planes with indices $\lfloor t_k \cdot K \rfloor$ and $\lfloor t_k \cdot K \rfloor + 1$. Finally, we calculate the precise relative position of the point between these two planes, which serves as the interpolation weight for subsequent operations.
For the projected feature tensor $\mathbf{F}^{proj}$, which is initialized with zeros, we assign values to the corresponding positions according to \Cref{eq-41}.

\begin{equation}\label{eq-41}
\begin{aligned}
&\mathbf{F}^{proj}[\lfloor t_k \cdot K \rfloor, \lfloor s_k \rfloor, \lfloor p_k \rfloor] = (1 - q_k) \cdot \mathbf{f}^{k}_{pix} \\
&\mathbf{F}^{proj}[\lfloor t_k \cdot K \rfloor+1, \lfloor s_k \rfloor, \lfloor p_k \rfloor] = q_k \cdot \mathbf{f}^{k}_{pix} \\ 
&where \quad q_k = t_k \cdot K - \lfloor t_k \cdot K \rfloor
\end{aligned}
\end{equation}

Subsequently, this projected feature $\mathbf{F}^{proj}$ is flipped and then added to both the clockwise and counter-clockwise volume features ($\mathbf{F}_{cy}^{+}$ and $\mathbf{F}_{cy}^{-}$), where $CV_{proj}$ is a convolution module:

\begin{equation}\label{eq-42}
\begin{aligned}
\mathbf{F}_{cy}^{+} = \mathbf{F}_{cy}^{+} + CV_{proj}(\mathbf{F}^{proj}) \\ 
\mathbf{F}_{cy}^{-} = \mathbf{F}_{cy}^{-} + CV_{proj}(\kappa(\mathbf{F}^{proj})) \\ 
\end{aligned}
\end{equation}

\section{Data Preprocessing} \label{datapre}
\subsection{Generated Sky Masks for the nuScenes Dataset}For precise sky segmentation, we employed the LISA model by loading its pre-trained weights from the LISA-13B-llama2-v1-explanatory version. LISA is an advanced segmentation model fine-tuned from a Large Language Model (LLM), with the key advantage of comprehending natural language instructions. Accordingly, we utilized the text prompt ``sky" to guide the model in automatically identifying and segmenting the sky regions within the images.
\begin{figure}[htb]
	\centering
	\includegraphics[width=\linewidth]{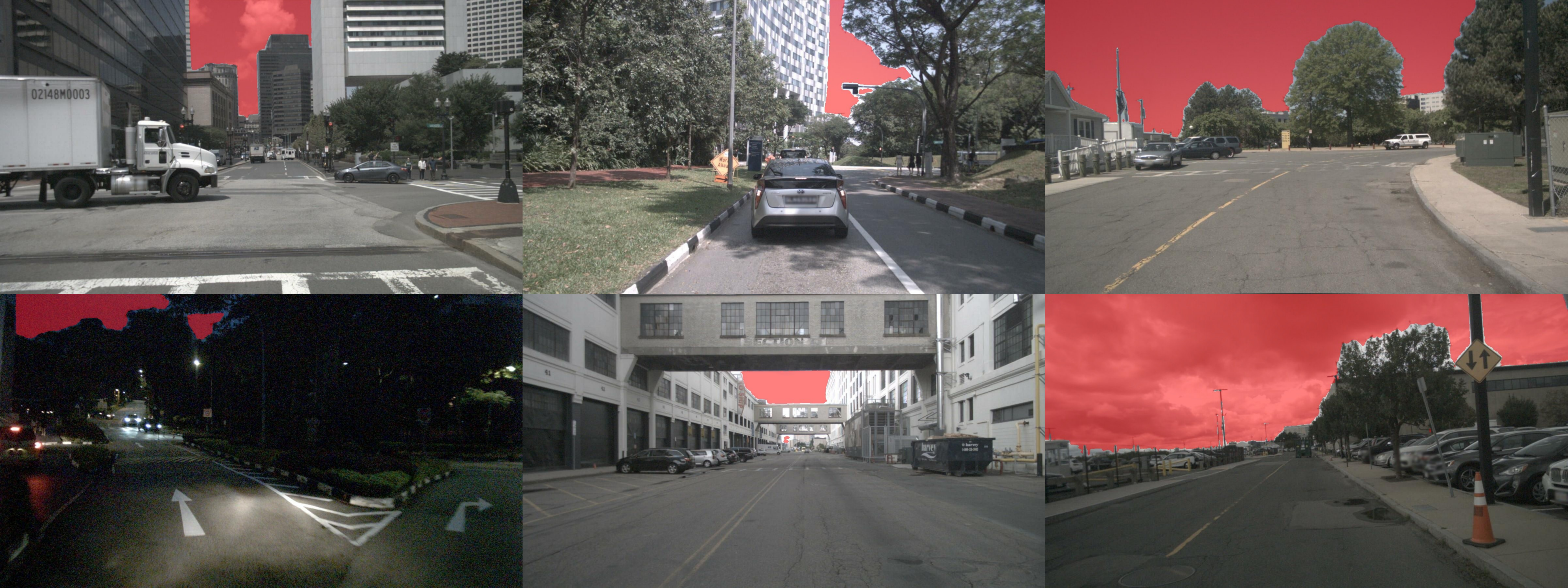}
	\caption{\textbf{Visualization of sky segmentation results from the LISA model on the nuScenes dataset.} Leveraging a language-guided segmentation model, we achieved highly accurate sky segmentation, even in challenging conditions such as complex and low-light scenes.} 
	\label{sky_mask}
\end{figure}

\Cref{sky_mask} illustrates a selection of visual results from our sky segmentation process. As can be clearly observed, the LISA model achieves excellent segmentation performance even under challenging low-light conditions, such as at night. Compared to conventional semantic segmentation models like SegFormer \citep{segformer}, LISA demonstrates significant advantages in both segmentation accuracy and robustness. We applied this segmentation procedure to all RGB images within the nuScenes dataset, encompassing both keyframes (samples) and intermediate frames (sweeps).

\subsection{Construction of a Custom nuScenes 3D Occupancy Dataset}Referring to \citet{wei2023surroundocc}, we utilize a comprehensive pipeline for generating dense 3D semantic occupancy grids from nuScenes LiDAR sequences. The method begins by initializing scene parameters including voxel size $\delta_v=0.4$, occupancy grid dimensions $\Omega=[L_o, H_o, W_o]=[40, 200, 200]$, and point cloud range $\mathcal{R}=[-40, -40, -3, 40, 40, 13]$. For each scene, the system processes sequential LiDAR frames, extracting both static background points and dynamic objects. 

Keyframe processing involves several stages. First, static points are aggregated across frames and transformed to a consistent coordinate system. Dynamic objects are handled through instance-centric point accumulation, where object points from multiple frames are transformed to canonical coordinates. The complete scene is reconstructed using Poisson surface reconstruction with configurable depth parameters and density-based vertex filtering, generating a watertight mesh $\mathcal{M}$. 

The mesh is then voxelized within the predefined spatial bounds $\mathcal{R}$, converting continuous surfaces into discrete 3D grids. Semantic labels are propagated to voxels through nearest-neighbor matching using the Chamfer distance between voxel centers and annotated sparse points. Finally, a denoising stage applies connected-component filtering to remove small noise regions, 3D morphological operations to smooth boundaries, and flood-fill algorithms to complete internal cavities while preserving semantic consistency.

The pipeline efficiently handles large-scale driving scenes through temporal aggregation of static elements and object-centric processing of dynamic entities. Output occupancy grids preserve fine-grained scene structures with accurate semantic labeling, suitable for downstream autonomous driving perception tasks. Memory management includes explicit garbage collection after scene processing to maintain computational efficiency.

\begin{figure}[htb]
	\centering
	\includegraphics[width=\linewidth]{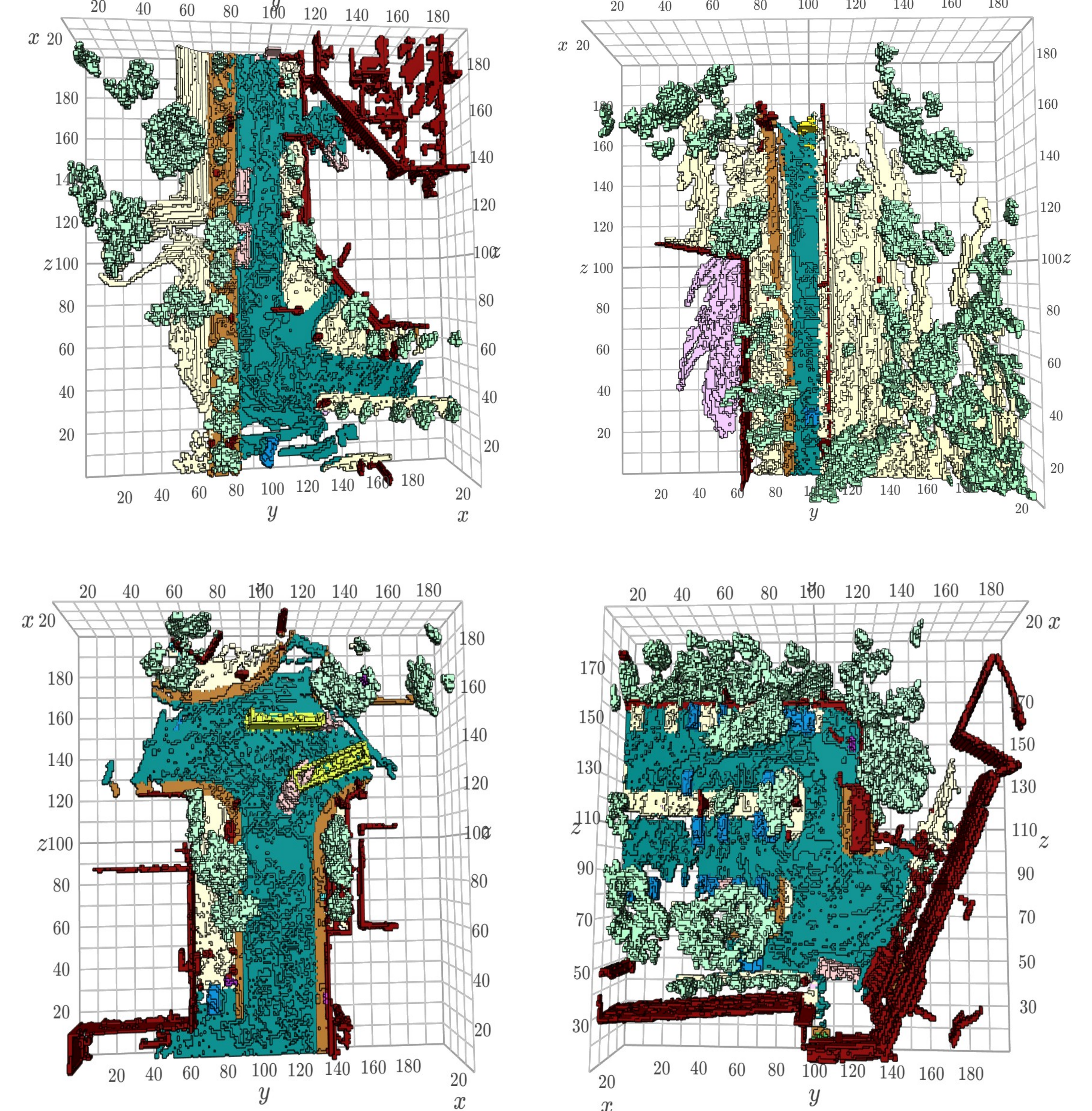}
	\caption{\textbf{Visualization of nuScenes occupancy labels generated from semantic point cloud processing.} Due to the limited precision of real-world data acquisition and errors from semantic point cloud processing, the occupancy representation in the nuScenes dataset exhibits a considerable amount of noise artifacts and sharp geometric features.} 
	\label{occvis}
\end{figure}

\Cref{occvis} illustrates visualization examples from the Occ dataset, which contains 17 semantic classes distinguished by different colors. In our methodology, we merge all non-air classes into a single foreground class, thereby simplifying the original multi-class task into a binary classification problem for supervision.

\subsection{Construction of the Carla-Centric Dataset}We adopted and extended SEED4D \citep{seed4d}, a synthetic data generation system based on the CARLA simulator \citep{carla}, which is designed to create dynamic driving scenarios with ego-inward views. To ensure our generated data fully aligns with the format and specifications of the nuScenes benchmark dataset, we made three key modifications to the original SEED4D system. First, we adjusted the camera configuration to match that of nuScenes, deploying five surround-view cameras with a 70-degree Field of View (FoV) and one rear-view camera with a 110-degree FoV. Second, we enabled the rendering of the ego-vehicle, ensuring it is visible in both ego-inward perspectives. Third, we integrated an additional occupancy grid generation module to synthesize precise, scene-aligned occupancy labels, with its configuration also adhering to the nuScenes standard.

\begin{figure}[htb]
	\centering
	\includegraphics[width=\linewidth]{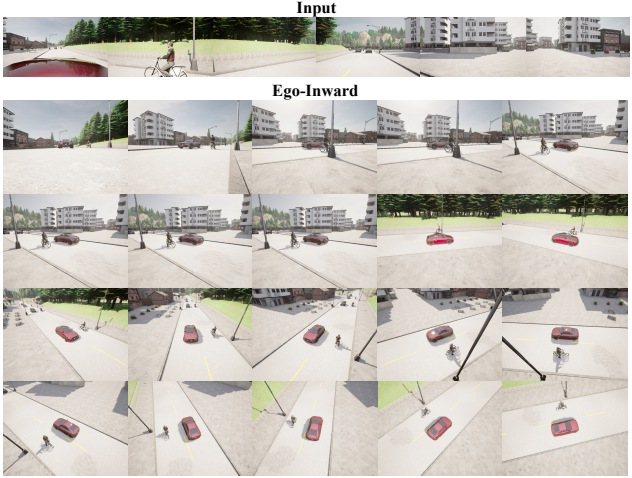}
	\caption{\textbf{A visualization example of a scene from the Carla-Centric dataset.} We present a set of six outward-looking images from the ego vehicle, along with 20 images sampled from a larger collection of 100 views that are uniformly distributed on a hemisphere and oriented towards the scene center.} 
	\label{seed4dsample}
\end{figure}

\Cref{seed4dsample} illustrates a typical scene from our constructed Carla-Centric dataset. Each scene consists of 6 input images and 100 ego-inward images. For clarity, only a subset of 20 ego-inward images from this scene is presented in the figure.

\begin{figure}[htb]
	\centering
	\includegraphics[width=\linewidth]{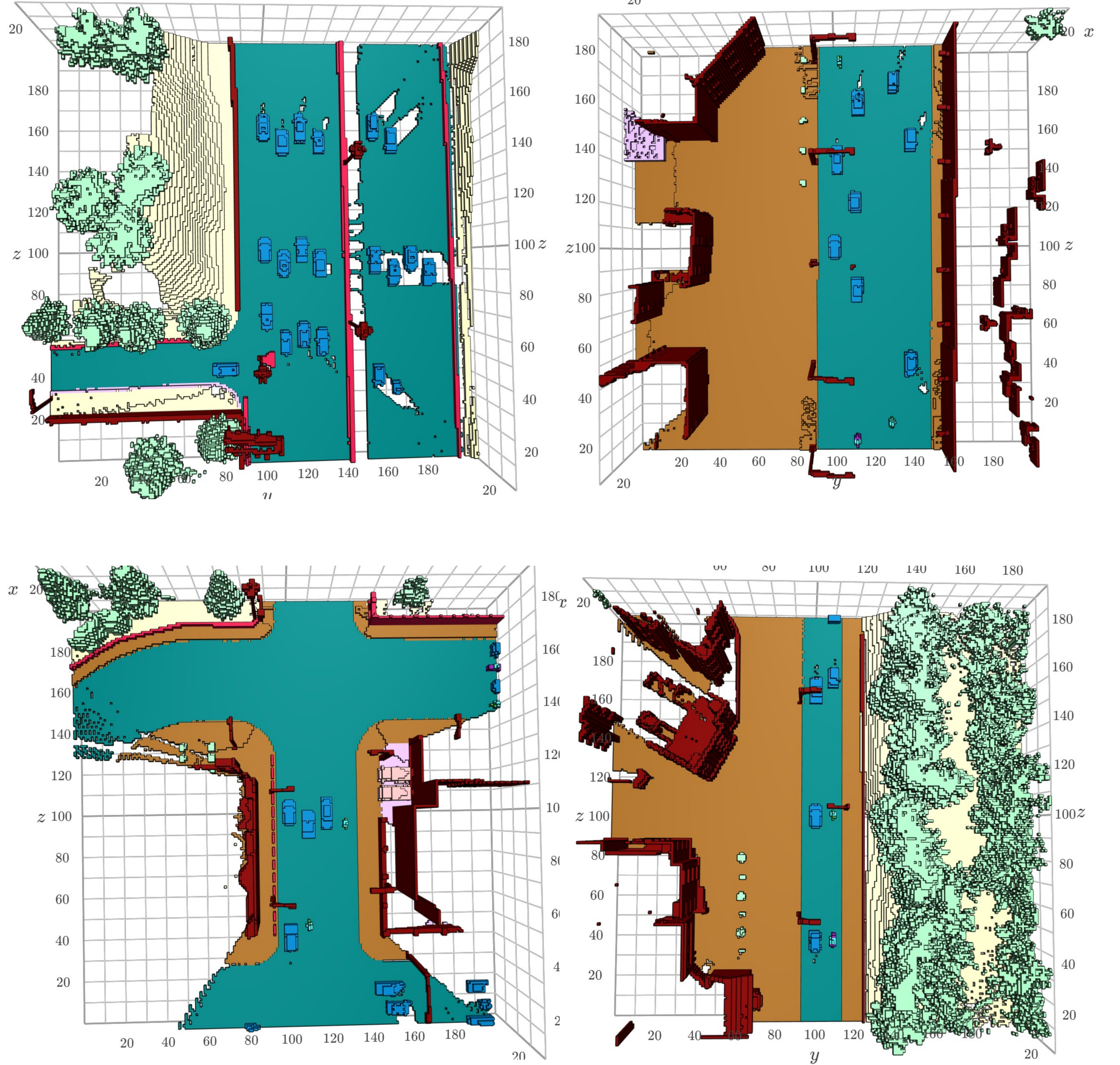}
	\caption{\textbf{The visualization of our proposed Carla-centric 3D occupancy.} In stark contrast to the results from nuScenes, our method generates a significantly smoother geometry that is free from noise artifacts and topological holes.}
	\label{carlaocc3d}
\end{figure}

\Cref{carlaocc3d} showcases the visualization of occupancy labels from our self-developed Carla-Centric dataset. In contrast to datasets based on real-world sensor data like nuScenes, our approach leverages the near-perfect ground truth available from the CARLA simulation environment. Benefiting from this high-fidelity ground truth, the resulting aggregated occupancy grids exhibit significant smoothness and spatial consistency, effectively mitigating the noise and sparsity issues commonly found in real-world data.

\subsection{Selection of the Feature Extractor}
\begin{figure*}[htb]
	\centering
	\includegraphics[width=\linewidth]{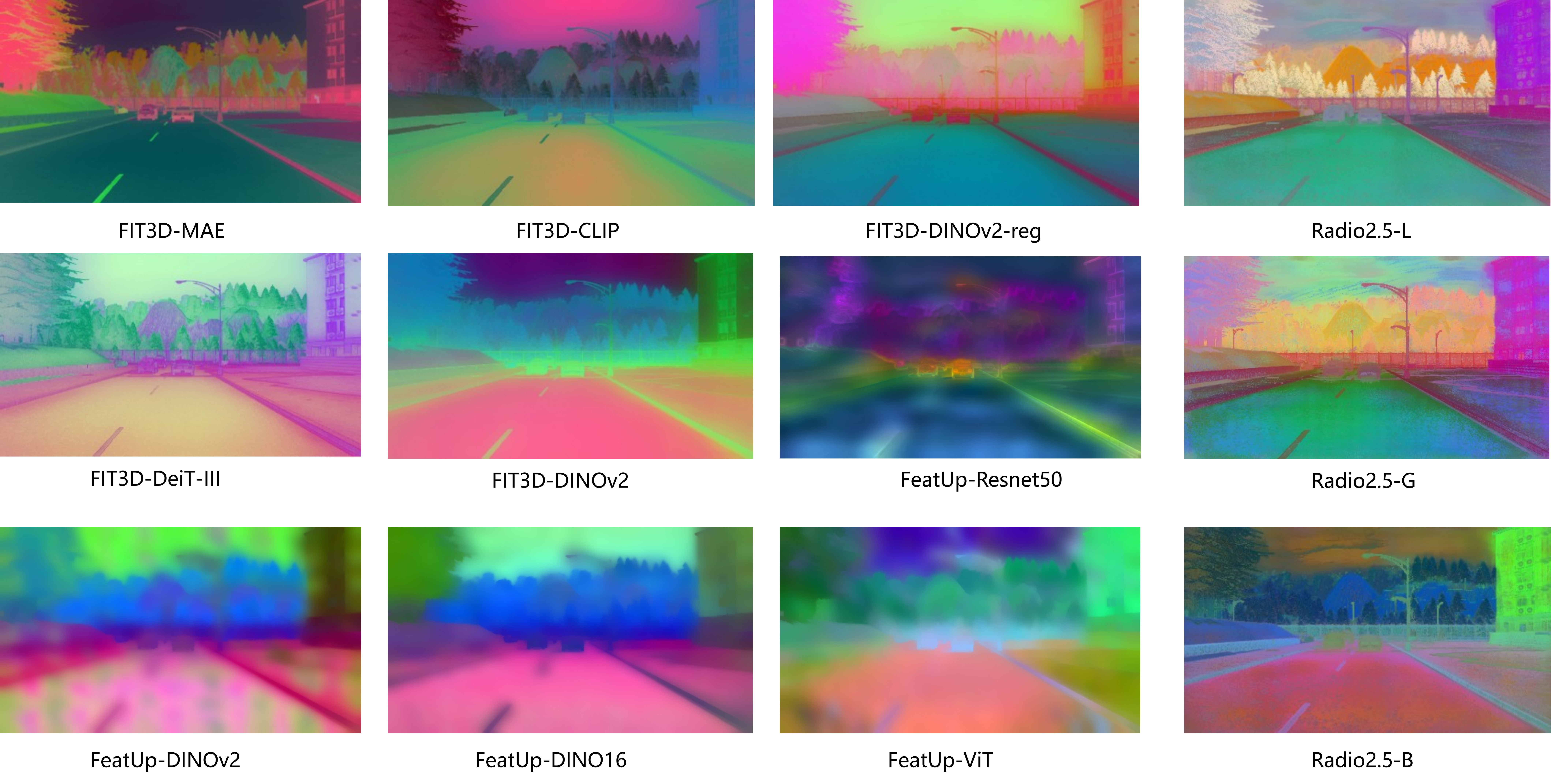}
	\caption{\textbf{Comparative visualization of feature maps extracted by different backbones.} Different families of feature extractors demonstrate distinct characteristics, which in turn lead to a significant disparity in performance.}
	\label{feat}
\end{figure*}

To identify the optimal feature extractor, we conducted a comparative analysis of several mainstream models as \Cref{feat} shows. Although the DINO series models (e.g., DINOv2 \citep{dinov2}) demonstrate powerful feature extraction capabilities, they are known to have certain issues, such as the presence of artifacts in the feature maps generated by DINOv2. Consequently, we focused our investigation on fine-tuned variants of DINO \citep{dino}, primarily examining FeatUp \citep{featup} and FiT3D \citep{fit3d}. Specifically, FeatUp addresses the low-resolution issue of the original DINO features, while FiT3D is optimized for 3D Gaussian Splatting (3DGS) scenes, making it more suitable for 3D tasks. Furthermore, for additional comparative visualization, we also utilized features extracted by the open-source Radio-v2.5 \citep{Radio} model from NVIDIA as a supplementary baseline.

To assess the representation quality of different models, we employed Principal Component Analysis (PCA) to visualize their feature maps. The analysis revealed that models fine-tuned with FiT3D, regardless of the backbone, produced overly smooth feature maps, potentially leading to a loss of fine-grained details. Meanwhile, models fine-tuned with FeatUp consistently exhibited severe artifacts and blurring, which compromised feature reliability. In stark contrast, the Radio-v2.5 series demonstrated a superior capability in preserving feature details and maintaining strong spatial consistency, an observation that aligns with the high praise it received in Feat2GS \citep{chen2025feat2gs}. Therefore, based on this qualitative analysis and its clear advantages in feature quality, we ultimately selected Radio-v2.5-B as the feature extractor for our method.

\section{Implement details} \label{impl}
\subsection{Overview of Evaluation Metrics} \label{metrics_more}
\textbf{Peak Signal-to-Noise Ratio (PSNR).} A widely used metric for quantifying the reconstruction quality of lossy compression and generation tasks. It measures the ratio between the maximum possible power of a signal and the power of corrupting noise that affects its fidelity. For an 8-bit image with a maximum possible pixel value of $L=255$ and a size of $H \times W$, the PSNR between a ground-truth image $I$ and a reconstructed image $\hat{I}$ is defined in decibels (dB) as:
\begin{equation}
\text{PSNR}(I, \hat{I}) = 10 \cdot \log_{10}\left(\frac{L^2}{\text{MSE}(I, \hat{I})}\right)
\end{equation}
Where the Mean Squared Error (MSE) is calculated as $\text{MSE}(I, \hat{I}) = \frac{1}{HW} \sum_{i=1}^{H}\sum_{j=1}^{W} (I_{ij} - \hat{I}_{ij})^2$. A higher PSNR value indicates a lower level of error, signifying that the reconstructed image is closer to the original. While simple and computationally efficient, PSNR's reliance on pixel-wise differences means it may not always align perfectly with human perceptual judgment of image quality.

\noindent
\textbf{Structural Similarity Index Measure (SSIM).} A perceptual metric designed to better approximate the human visual system's assessment of image similarity. Unlike PSNR, SSIM evaluates the degradation of quality as a change in structural information. It compares two images, $I$ and $\hat{I}$, based on three components: luminance ($l$), contrast ($c$), and structure ($s$). For two image patches $x$ and $y$ from $I$ and $\hat{I}$ respectively, the SSIM is computed as:
\begin{equation}
\text{SSIM}(x, y) = \frac{(2\mu_x\mu_y + c_1)(2\sigma_{xy} + c_2)}{(\mu_x^2 + \mu_y^2 + c_1)(\sigma_x^2 + \sigma_y^2 + c_2)}
\end{equation}
Where $\mu_x, \mu_y$ are the local means, $\sigma_x, \sigma_y$ are the local standard deviations, and $\sigma_{xy}$ is the cross-covariance. The constants $c_1 = (k_1L)^2$ and $c_2 = (k_2L)^2$ are included to stabilize the division. The final SSIM score is the mean of the SSIM values computed over all local windows in the image. The score ranges from -1 to 1, where 1 indicates perfect structural similarity.

\noindent
\textbf{Learned Perceptual Image Patch Similarity (LPIPS).} Also known as ``perceptual loss", it measures the distance between two images in a perceptually relevant feature space. It more closely mirrors human perception of image similarity than traditional metrics like PSNR and SSIM. To compute LPIPS, two images, $I$ and $\hat{I}$, are passed through a pre-trained deep neural network (e.g., VGG or AlexNet). The feature activations are extracted from multiple layers, $l$. For each layer, the activations are unit-normalized in the channel dimension ($F^l, \hat{F}^l \in \mathbb{R}^{H_l \times W_l \times C_l}$). The L2 distance is then computed, scaled by a learned weight vector $w_l$, and averaged over spatial dimensions ($H_l, W_l$):
\begin{equation}
\text{LPIPS}(I, \hat{I}) = \sum_{l} \frac{1}{H_l W_l} \sum_{h,w} \| w_l \odot (F^l_{hw} - \hat{F}^l_{hw}) \|_2^2
\end{equation}
The final LPIPS score is the sum of distances across all considered layers. A lower LPIPS score signifies that the two images are more similar from a perceptual standpoint, indicating a higher-quality reconstruction.

\noindent
\textbf{Pearson Correlation Coefficient (PCC).} A statistical measure that evaluates the linear relationship between two sets of data. In the context of computer vision, it is often applied to depth map evaluation, where it assesses the correlation between the predicted depth values and the ground-truth depth values, irrespective of absolute scale and shift. For a predicted depth map $\hat{D}$ and a ground-truth depth map $D$, the PCC is defined as the covariance of the two variables divided by the product of their standard deviations:

\begin{equation}
\begin{aligned}
&\text{PCC}(D, \hat{D}) = \frac{\text{cov}(D, \hat{D})}{\sigma_D \sigma_{\hat{D}}}\\
 &= \frac{\sum_{i=1}^{N}(D_i - \mu_D)(\hat{D}_i - \mu_{\hat{D}})}{\sqrt{\sum_{i=1}^{N}(D_i - \mu_D)^2} \sqrt{\sum_{i=1}^{N}(\hat{D}_i - \mu_{\hat{D}})^2}}
\end{aligned}
\end{equation}

Where $N$ is the total number of valid pixels, $\mu_D$ and $\mu_{\hat{D}}$ are the mean depth values, and $\sigma_D$ and $\sigma_{\hat{D}}$ are their standard deviations. The PCC ranges from -1 to +1, where +1 indicates a perfect positive linear correlation, 0 indicates no linear correlation, and -1 indicates a perfect negative linear correlation. For depth evaluation, a higher PCC value is desirable.

\subsection{A Brief Overview of The Baselines.}

\textbf{6Img-to-3D.}\label{baselines_more} 
We utilize the official implementation of 6Img-to-3D\footnote{\url{https://github.com/continental/6Img-to-3D}}. Diverging from the original seed4d dataset employed in their work, we construct a dataset that is fully aligned with nuScenes. Please refer to Appendix A.6 for detailed distinctions between these datasets. Additionally, we integrate PCC evaluation metrics into the codebase to facilitate comprehensive experimental comparisons.

\noindent
\textbf{Omni-Scene.} 
We utilize the official repository\footnote{\url{https://github.com/WU-CVGL/Omni-Scene}} of Omni-Scene. For the Carla-Centric dataset, we adjust the computation of ray origins and direction vectors to accommodate its distinct coordinate system.
To align with the dataset resolution of nuScenes as used in Omni-Scene, we downsampled the Carla-Centric input and target images by a factor of four. For the experiments on nuScenes, we kept the settings identical to theirs.

We assign the maximum depth value to sky regions, with an associated confidence of 1.0. To maintain a rigorous and fair evaluation, the rendered depth maps undergo a filtering step at inference time: any depth value surpassing a predefined threshold is subsequently set to zero, thereby ensuring the consistency of the depth map in our model.

\noindent
\textbf{DrivingForward.} 
We utilize the official repository\footnote{\url{https://github.com/fangzhou2000/DrivingForward}} of DrivingForward. The original DrivingForward architecture is designed to leverage contextual information from three consecutive frames (t-1, t, t+1) to achieve high-quality reconstruction. For a fair comparison with our single-frame approach, we adapted its single-frame (SF) variant for our experiments. Furthermore, we addressed several underlying issues within its codebase to ensure compatibility with our dataset. A key modification was replacing its rendering kernel to enable the accurate generation of depth maps, a prerequisite for computing the Pearson Correlation Coefficient (PCC) metric.

\noindent
\textbf{Depthsplat \& MVSplat \& PixelSplat.} 
We utilize the official repositories of Depthsplat\footnote{\url{https://github.com/cvg/depthsplat}}, MVSplat\footnote{\url{https://github.com/donydchen/mvsplat}} and PixelSplat\footnote{\url{https://github.com/dcharatan/pixelsplat}}. For the Carla-Centric dataset, which provides ground-truth (GT) depth maps, we directly utilize these for supervision, bypassing the need for pseudo-labels from models like DepthAnything. To ensure the optimal performance of baseline models that require specific input aspect ratios, we maintain their original image resolutions during their respective training and inference processes. For a fair and consistent evaluation, the final outputs from all methods are rendered to a common resolution matching that of our model before comparison.

\textbf{SplatterImage.} 
We utilize the official repository\footnote{\url{https://github.com/szymanowiczs/splatter-image}} of SplatterImage. We observe that the baseline model suffers from significant training instability when applied to our dataset. The training process is prone to divergence, often collapsing after approximately 3,000 iterations. This instability is highly sensitive to the random initialization seed, frequently resulting in catastrophic performance with PSNR values below 10. To establish a meaningful benchmark for comparison, we conduct an extensive search across dozens of random seeds and report the results from the most successful run.

\subsection{Detailed Architecture of Our Model} \label{Detailedarch}
\Cref{tab:arch_hyperparameters_nuScenes} and \Cref{tab:arch_hyperparameters_carla} provide a detailed breakdown of our model's architecture, including the parameters for each perception branch, the configurations of the Cylinder Plane and CPFG modules, and key training hyperparameters.
We trained our model on 8 NVIDIA L40S GPUs for approximately 2 days for occupancy prediction and 5 days for reconstruction. The training process required about 32 GB of GPU memory per device.

We employ Radio-V2.5 \citep{Radio} as the 2D image encoder, which performs 16x downsampling to transform an input image of size 3x896x1600 (4x upsample to the original image) into a 768x56x100 feature map. These features are then used to construct a Cylinder Plane Feature by interpolating information from six source views into a 56x512 feature map, yielding a 768x56x512 feature volume with its Field of View (FOV) coefficients set to $\rho_o, \rho_v, \rho_p$ and height offsets set to $\Delta h_o, \Delta h_v, \Delta h_p$. For the volume-based module, the feature is processed by XNet, a U-Net-based network. Each encoder block contains two ResNet blocks and two ema attention modules, and each decoder block includes three corresponding blocks, with channel dimensions of 432, 624, and 624. XNet outputs two feature vectors of lengths 432 and 144, which are used for the construction of the Geometry CPFG and Appearance CPFG. Specifically, the 432-dimensional appearance feature is structured into an appearance CPFG of 48 equidistant cylindrical bins, each with a feature length of 36, while the 144-dimensional geometry feature is processed into a 12-dimensional Geometry CPFG. The YNet, which handles occlusion and pixel-level information, shares a similar architecture but with different feature dimensions as detailed in \Cref{tab:arch_hyperparameters_nuScenes} and \Cref{tab:arch_hyperparameters_carla}. The ZNet comprises three ResNet blocks with an upsample module followed by synthesis blocks with progressively decreasing feature lengths and finally synthesizing an RGB image. Each network has its corresponding decoders to form occupancy, BEV, geometry, and texture.
\label{totalloss}
We formulate occupancy prediction as a binary classification task to distinguish between occupied and free space, rather than predicting semantic labels. 
The module is trained with a composite loss $L_{occ}$ from \Cref{eq-loss}, combining semantic ($L^{sem}_{scal}$), geometric ($L^{geo}_{scal}$), and cross-entropy ($L_{ce}$) losses from MonoScene \citep{monoscene} with a BEV loss ($L_{bev}$) from FastOcc \citep{FastOcc}. The ground-truth labels $\mathbf{P}_{3D}$ and $\mathbf{P}_{2D}$ are generated as described in \Cref{datapre}. The terms $\lambda_{sem}, \lambda_{geo}, \lambda_{ce}, \lambda_{bev}$ are weights for each loss component.
\begin{equation} \label{eq-loss}
\begin{aligned}
    L_{occ} = \lambda_{sem} L^{sem}_{scal}(\hat{\mathbf{P}}_{3D}, \mathbf{P}_{3D}) \\
    + \lambda_{geo} L^{geo}_{scal}(\hat{\mathbf{P}}_{3D}, \mathbf{P}_{3D}) \\
    + \lambda_{ce} L_{ce}(\hat{\mathbf{P}}_{3D}, \mathbf{P}_{3D}) + \lambda_{bev} L_{bev}(\hat{\mathbf{P}}_{2D}, \mathbf{P}_{2D})
\end{aligned}
\end{equation}

The overall optimization objective of our model is composed of several loss components: an image reconstruction loss $L_1^{I}(\hat{\mathbf{I}},\mathbf{I})$, a perceptual similarity loss $L_{lpips}(\hat{\mathbf{I}},\mathbf{I})$, an L1 loss for the sky mask $L_{sky}(\hat{\mathbf{A}}_{fg},\mathbf{A}_{fg})$, and both L1 and Pearson depth losses refer to \citep{sparsegs} for the depth map $L_1^{depth}(\hat{\mathbf{D}}_{fg},\mathbf{D}_{fg}), L_{pear}(\hat{\mathbf{D}}_{fg},\mathbf{D}_{fg})$, as formulated in \Cref{eq-loss2}. The ground truth for the depth map is generated by the Metric3D-v2 \citep{metric3dv2}, while that for the sky mask is obtained from the LISA \citep{LISA} as described in \Cref{datapre}. The terms $\lambda_{I}, \lambda_{lpips}, \lambda_{sky}, \lambda_{depth}, \lambda_{pear}$ are weights for each loss component.
\begin{equation} \label{eq-loss2}
\begin{aligned}
    L_{total} = \lambda_I L_{1}^{I} + \lambda_{lpips} L_{lpips} \\
    + \lambda_{sky} L_{sky} + \lambda_{depth} L_1^{depth} + \lambda_{pear} L_{pear}
\end{aligned}
\end{equation}

\clearpage
\newpage

\begin{table*}[t]
\centering
\caption{Model architecture and training specifications of our model on nuScenes.}
\scriptsize
\label{tab:arch_hyperparameters_nuScenes}
\renewcommand{\arraystretch}{0.5}
\setlength{\tabcolsep}{3pt}
\resizebox{0.65\linewidth}{!}{ 
    \begin{tabular}{lll}
    \toprule
    \multicolumn{3}{l}{\textbf{(a) Network Architecture}} \\
    \midrule
    \multirow{2}{*}{2D Image Encoder} & backbone              & Radio-v2.5 \citep{Radio} \\
                                      & Out resolution        & $768\times56\times100$ \\
    \midrule
    \addlinespace 
    \multirow{2}{*}{Cylinder Plane} & resolution $D_{feat} \times H_u \times W_u$              & $768\times56\times512$ \\
                                      & \#$\rho_o,\rho_v,\rho_p,\Delta h_o,\Delta h_v,\Delta h_p$        & 0.9, 0.98, 0.98,0.0,0.4,0.4 \\
    \midrule
    \addlinespace
    \multirow{4}{*}{Occupancy-Aware}    & \# blocks per resolution & 2 \\
                                      & \# downsample dims      & 384, 576, 576 \\
                                      & \# upsample dims      & 576, 576, 384 \\
                                      & \# out dims      & 384 \\
    \multirow{2}{*}{Occupancy-CPFG} & \# CPFG planes num $K$            & 48 \\
                                & \# CPFG planes dims $D_{occ}$        & 8 \\                          
    \midrule
    \addlinespace                           
    \multirow{5}{*}{Volume-Aware} & \# blocks per resolution             & 2 \\
                                          & \# downsample dims         & 432, 624, 624 \\
                                          & \# upsample dims      & 624, 624, 432 \\
                                          & \# texture out dims      & 432 \\
                                          & \# geometry out dims         & 144 \\
    \multirow{2}{*}{Volume-CPFG} & \# CPFG planes num $K$            & 48 \\
                                & \# CPFG planes dims $D_{geo},D_{app}$       & 12,36 \\
            
    \midrule
    \addlinespace
    \multirow{6}{*}{Pixel-Aware}     & \# Upsampling factor $k_i,k_o$ & 4,1 \\
    & \# Upsample output dim $D_{pix}$& 128\\
                                      & \# blocks per resolution &1 \\
                                      & \# downsample dims      & 128, 256, 512, 512 \\
                                      & \# upsample dims      &512, 512, 256. 128\\
                                      & \# out dims      & 128 \\
    \midrule
    \addlinespace
    \multirow{3}{*}{Background} & \# block dims   & 24, 24, 24 \\
                                      & \# block dims after ray cast     & 24, 12, 6, 3 \\
                                      & \# target resolution $H_t,W_t$     & 224,400 \\
    \midrule
    \addlinespace
    \multirow{4}{*}{Occ Decoder}     
                                        & \# MLP layers     & 4\\
                                        & \# MLP input dims     & 8\\
                                        & \# MLP width     & 4\\
                                        & \# MLP output dims     &4\\
    \multirow{4}{*}{BEV Decoder}     
                                        & \# MLP layers     & 3\\
                                        & \# MLP input dims     & 128\\
                                        & \# MLP width     & 64\\
                                        & \# MLP output dims     &2\\
    \midrule
    \addlinespace
    \multirow{7}{*}{Pixel Decoder}      
                                        & \# groupnorm channel     & 128 \\
                                        & \# groupnorm groups     & 32\\
                                        & \# MLP layers     & 1\\
                                        & \# MLP input dims     & 128\\
                                        & \# MLP width     & None\\
                                        & \# MLP output dims     & 14\\
                                        & \# Gaussians per pixel $G_p$     & 1\\
    \midrule
    \addlinespace
    \multirow{5}{*}{Volume Decoder}     
                                        & \# MLP layers     & 3\\
                                        & \# MLP input dims     & 48\\
                                        & \# MLP width     & 96\\
                                        & \# MLP output dims     & 42\\
                                        & \# Gaussians per voxel $G_v$     & 3\\
    \midrule

    \multicolumn{3}{l}{\textbf{(b) Hyperparameters}} \\
    \midrule
    \multirow{2}{*}{Loss Weights}                       & \# $\lambda_{sem}, \lambda_{geo}, \lambda_{ce}, \lambda_{bev}$ & 0.02, 0.02, 0.1, 0.1 \\
    & \# $\lambda_{1}, \lambda_{lpips}, \lambda_{sky}, \lambda_{depth}, \lambda_{pear}$ & 1.0, 0.05, 0.5, 0.01, 0.01 \\
    \addlinespace
    \multirow{8}{*}{Training Details} & learning rate scheduler & Cosine \\
                                      & \# iterations           & 100,000 \\
                                      & \# learning rate        & 1e-4 \\
                                      & optimizer               & AdamW \citep{AdamW} \\
                                      & \# beta1, beta2         & 0.9, 0.999 \\
                                      & \# weight decay         & 0.01 \\
                                      & \# warm-up              & 1000 \\
                                      & \# gradient clip        & 1.0 \\
    \bottomrule
    \end{tabular}
    }
\end{table*}
\begin{table*}[t]
\centering
\caption{Model architecture and training specifications of our model on Carla-Centric.}
\scriptsize
\label{tab:arch_hyperparameters_carla}
\renewcommand{\arraystretch}{0.5}
\setlength{\tabcolsep}{3pt}
\resizebox{0.65\linewidth}{!}{ 
    \begin{tabular}{lll}
    \toprule
    \multicolumn{3}{l}{\textbf{(a) Network Architecture}} \\
    \midrule
    \multirow{2}{*}{2D Image Encoder} & backbone              & Radio-v2.5 \citep{Radio} \\
                                      & Out resolution        & $768\times56\times100$ \\
    \midrule
    \addlinespace 
    \multirow{2}{*}{Cylinder Plane} & resolution $D_{feat} \times H_u \times W_u$             & $768\times56\times512$ \\
                                      & \#$\rho_o,\rho_v,\rho_p,\Delta h_o,\Delta h_v,\Delta h_p$        & 0.9, 1.6, 1.6,0.0,0.0,0.0 \\
    \midrule
    \addlinespace                          
    \multirow{5}{*}{Volume-Aware} & \# blocks per resolution             & 2 \\
                                          & \# downsample dims         & 432, 624, 624 \\
                                          & \# upsample dims      & 624, 624, 432 \\
                                          & \# texture out dims      & 432 \\
                                          & \# geometry out dims         & 144 \\
    \multirow{2}{*}{Volume-CPFG} & \# CPFG planes num             & 48 \\
                                & \# CPFG planes dims $D_{geo},D_{app}$        & 12,36 \\
    \midrule
    \addlinespace
    \multirow{4}{*}{Occupancy-Aware}    & \# blocks per resolution & 2 \\
                                      & \# downsample dims      & 384, 576, 576 \\
                                      & \# upsample dims      & 576, 576, 384 \\
                                      & \# out dims      & 384 \\
    \multirow{2}{*}{Occupancy-CPFG} & \# CPFG planes num  $K$            & 48 \\
                                & \# CPFG planes dims $D_{occ}$        & 8 \\                                  
    \midrule
    \addlinespace
    \multirow{6}{*}{Pixel-Aware}     & \# Upsampling factor $k_i,k_o$ & 2,2 \\
    & \# Upsample output dim $D_{pix}$&  24\\
                                      & \# blocks per resolution & 2 \\
                                      & \# downsample dims      & 48, 48, 48 \\
                                      & \# upsample dims      & 48, 48, 48\\
                                      & \# out dims      & 48 \\
    \midrule
    \addlinespace
    \multirow{3}{*}{Background} & \# block dims   & 24, 24, 24 \\
                                      & \# block dims after ray cast     & 24, 12, 6, 3 \\
                                      & \# target resolution $H_t,W_t$     & 150,200 \\
    \midrule
    \addlinespace
    \multirow{4}{*}{Occ Decoder}     
                                        & \# MLP layers     & 4\\
                                        & \# MLP input dims     & 8\\
                                        & \# MLP width     & 4\\
                                        & \# MLP output dims     &4\\
    \multirow{4}{*}{BEV Decoder}     
                                        & \# MLP layers     & 3\\
                                        & \# MLP input dims     & 128\\
                                        & \# MLP width     & 64\\
                                        & \# MLP output dims     &2\\
    \midrule
    \addlinespace
    \multirow{7}{*}{Pixel Decoder}      
                                        & \# groupnorm channel     & 24 \\
                                        & \# groupnorm groups     & 6 \\
                                        & \# MLP layers     & 4\\
                                        & \# MLP input dims     & 24\\
                                        & \# MLP width     & 24\\
                                        & \# MLP output dims     & 14\\
                                        & \# Gaussians per pixel $G_p$     & 1\\
    \midrule
    \addlinespace
    \multirow{7}{*}{Volume Decoder}     
                                        & \# MLP layers     & 8\\
                                        & \# tex MLP input dims     & 48\\
                                        & \# geo MLP input dims     & 12\\
                                        & \# MLP width     & 64\\
                                        & \# tex MLP output dims     & 33\\
                                        & \# geo MLP output dims     & 9\\
                                        & \# Gaussians per voxel $G_v$     & 3\\
    \midrule

    \multicolumn{3}{l}{\textbf{(b) Hyperparameters}} \\
    \midrule
    \multirow{2}{*}{Loss Weights}                       & \# $\lambda_{sem}, \lambda_{geo}, \lambda_{ce}, \lambda_{bev}$ & 0.02, 0.02, 0.1, 0.1 \\
    & \# $\lambda_{1}, \lambda_{lpips}, \lambda_{sky}, \lambda_{depth}, \lambda_{pear}$ & 1.0, 0.05, 0.5, 0.01, 0.01 \\
    \addlinespace
    \multirow{8}{*}{Training Details} & learning rate scheduler & Cosine \\
                                      & \# iterations           & 100,000 \\
                                      & \# learning rate        & 1e-4 \\
                                      & optimizer               & AdamW \citep{AdamW} \\
                                      & \# beta1, beta2         & 0.9, 0.999 \\
                                      & \# weight decay         & 0.01 \\
                                      & \# warm-up              & 1000 \\
                                      & \# gradient clip        & 1.0 \\
    \bottomrule
    \end{tabular}
    }
\end{table*}

\clearpage
\newpage
\section{More Experimental Results} 
\label{moreexp}

\subsection{Efficiency Analysis}
We compare the performance of different feed-forward models for autonomous driving in terms of inference speed and GPU memory consumption. As shown in \Cref{tab:efficiency_nuscenes}, while the inference speed of our proposed model is marginally slower than that of existing methods, it remains within an acceptable range. Furthermore, the model has a low VRAM footprint, allowing it to be readily deployed on consumer-grade graphics cards. Importantly, our model achieves state-of-the-art performance in both visual quality and depth estimation, and it exhibits excellent compatibility. Therefore, the slower inference speed is a reasonable trade-off for these superior performance metrics.
\begin{table}[htbp]
\centering
\setlength{\tabcolsep}{1pt}
\caption{Efficiency comparison on the nuScenes dataset.}
\label{tab:efficiency_nuscenes}
\small
\begin{tabular}{l|c|c|c|c|c|c}
\hline
Method & FPS$\uparrow$ & Mem.$\downarrow$ &  PSNR$\uparrow$ & SSIM$\uparrow$ &  PCC $\uparrow$ & Compatible\\
\hline
6Img-to-3D & \best{2.63}          & 15.65    & 20.74          & 0.560          & 0.570 & $\times$ \\
Omin-Scene & \second{2.5}          & \best{8.34}    & \second{24.11}          & \second{0.734}          & \second{0.816} & $\times$ \\
XYZCylinder     & 2.1 & \second{9.81}             & \best{24.97} & \best{0.750} & \best{0.887}      &  \checkmark \\
\hline
\end{tabular}
\end{table}

\begin{table}[htbp]
\centering
\setlength{\tabcolsep}{1pt}
\caption{Ablation of Feature Extractor.}
\label{tab:feature_ablation}
\small
\begin{tabular}{l|c|c|c|c}
\hline
Extractor &  PSNR$\uparrow$ & SSIM$\uparrow$ &  LPIPS$\downarrow$ & PCC $\uparrow$\\
\hline
DINOv2-vitb14 & 24.34          & 0.732    & 0.242          & 0.878          \\
DINOv3-vitb16 & \third{24.53}          & \second{0.747}  & \best{0.230}          & \second{0.884}          \\
FIT3D-MAE & \second{24.81}          & \third{0.741}     & \third{0.233}          & \best{0.887}          \\
Radio-v2.5b (ours)    & \best{24.97} & \best{0.750}             & \second{0.231} & \best{0.887}     \\
\hline
\end{tabular}
\end{table}

\begin{table*}[t]
    \centering
    \caption{\textbf{Quantitative comparison of our model against the baselines.} The \colorbox[HTML]{FF9898}{best}, \colorbox[HTML]{FFCB98}{second-best}, and \colorbox[HTML]{FFFF98}{third-best} results are marked with corresponding cell colors, reflecting nuanced performance across different metrics.}
    \label{chaocan-table-updated}
    % \scriptsize
    \begin{tabular}{c|c|c|c|c|c|c|c|c|c}
        \Xhline{1.2pt}
                &          & \multicolumn{4}{c|}{Carla-Centric}             & \multicolumn{4}{c}{nuScenes}          \\
        \hline
        $K$ &   $G_v$      & PSNR$\uparrow$      & LPIPS$\downarrow$    & SSIM$\uparrow$     & PCC$\uparrow$      & PSNR$\uparrow$    & LPIPS$\downarrow$   & SSIM$\uparrow$    & PCC$\uparrow$     \\
        \hline 
        36 & 1     & 17.61     & 0.448     & 0.582     & 0.796     & 24.01   & 0.275   & 0.705   & 0.853   \\
        36 & 2     & 17.58     & 0.451     & 0.579     & 0.794     & 24.16   & 0.268   & 0.712   & 0.859   \\
        36 & 3     & 17.63     & 0.449     & 0.581     & 0.795     & 24.28   & 0.262   & 0.719   & 0.866   \\
        48 & 1     & \third18.38     & \second0.362     & \third0.619     & \second0.818     & \third24.71   & 0.250   & 0.733   & 0.871   \\
        48 & 2     & 18.37     & \third0.364     & \second0.620     & \best0.819     & \second24.92   & \third0.244  & \best0.755   & \best0.889   \\
        48 & 3     & \second18.40     & \best0.359     & \best0.622     & \third0.817     & \best24.97   & \best0.231   & \second0.750   & \second0.887   \\
        60 & 1     & 18.28     & 0.375     & 0.611     & 0.809     & 24.46   & 0.255   & 0.729   & 0.868   \\
        60 & 2     & \best18.42     & 0.368     & 0.616     & 0.814     & 24.68   & 0.246   & \third0.746   & \third0.880   \\
        60 & 3     & 18.35     & 0.371     & 0.615     & 0.812     & 24.65   & \second0.237   & 0.742   & 0.875   \\
        \Xhline{1.2pt}
    \end{tabular}
    \label{tab:my-table-updated}
\end{table*}

\subsection{Ablation of Feature Extractor}

We compare our model's performance on the nuScenes dataset using different feature extractors. We selected the base versions of the most common feature extractors, such as DINOv2 \cite{dinov2}, DINOv3 \cite{dinov3}, FIT3D \cite{fit3d} and Radio-v2.5 \cite{Radio}, all of which have comparable parameter counts. The results in \Cref{tab:feature_ablation} demonstrate the robust performance of Radio-v2.5. The DINO series of models were slightly outperformed by Radio and FIT3D. We attribute this to the fact that FIT3D fine-tunes its feature extractor on 3D data, whereas Radio distills knowledge from multiple extractors, yielding a knowledge space that is either more expansive or better aligned with 3D representations. DINOv3 outperforms DINOv2 as it was trained on a more extensive dataset.

\begin{figure}[t]
	\centering
		\centering
		\includegraphics[width=\linewidth]{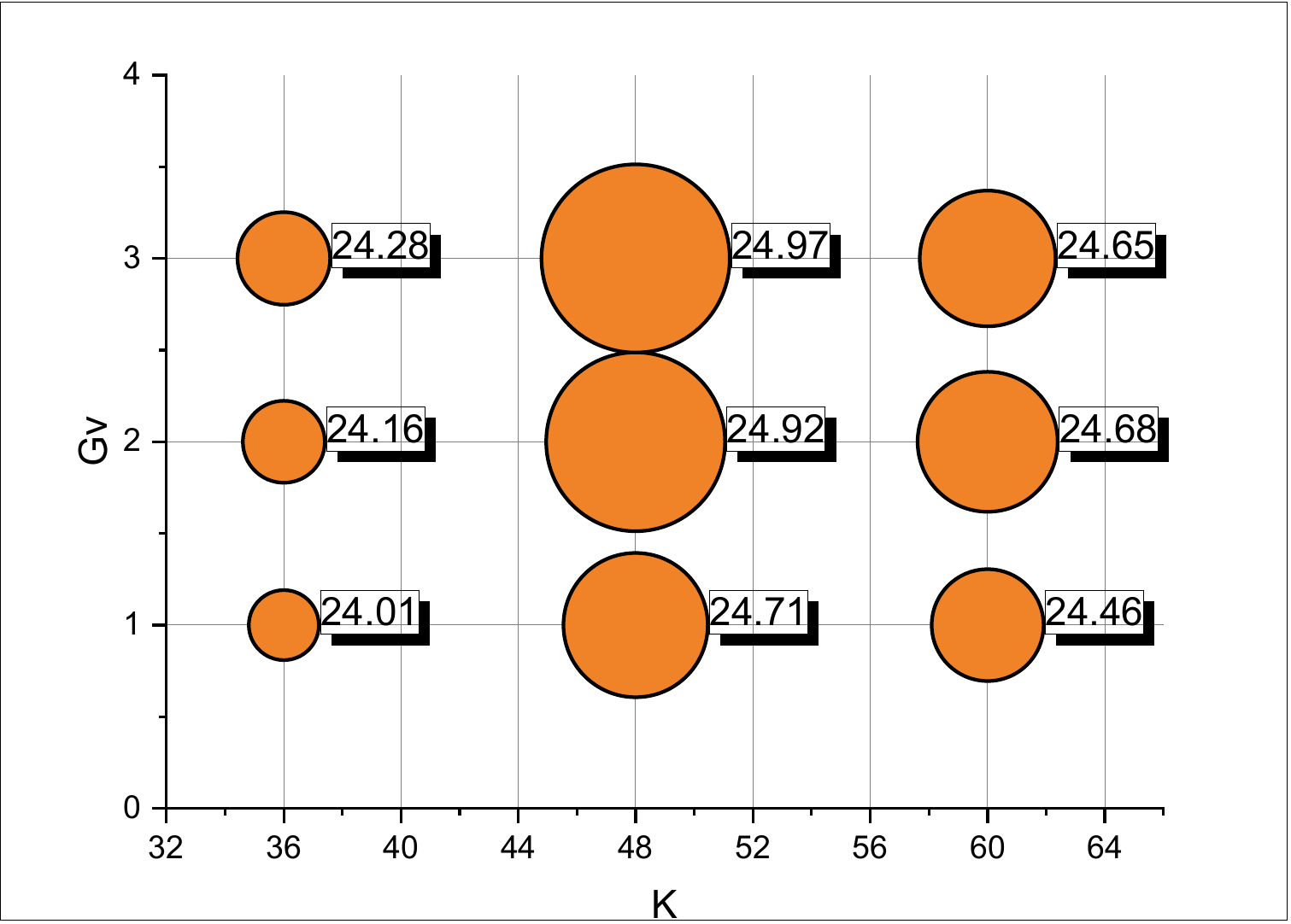}
		\caption{PSNR on the nuScenes dataset for models trained from scratch with varying hyperparameters.} 
		\label{chaocan}
\end{figure}
\begin{figure}[t]
		\centering
		\includegraphics[width=\linewidth]{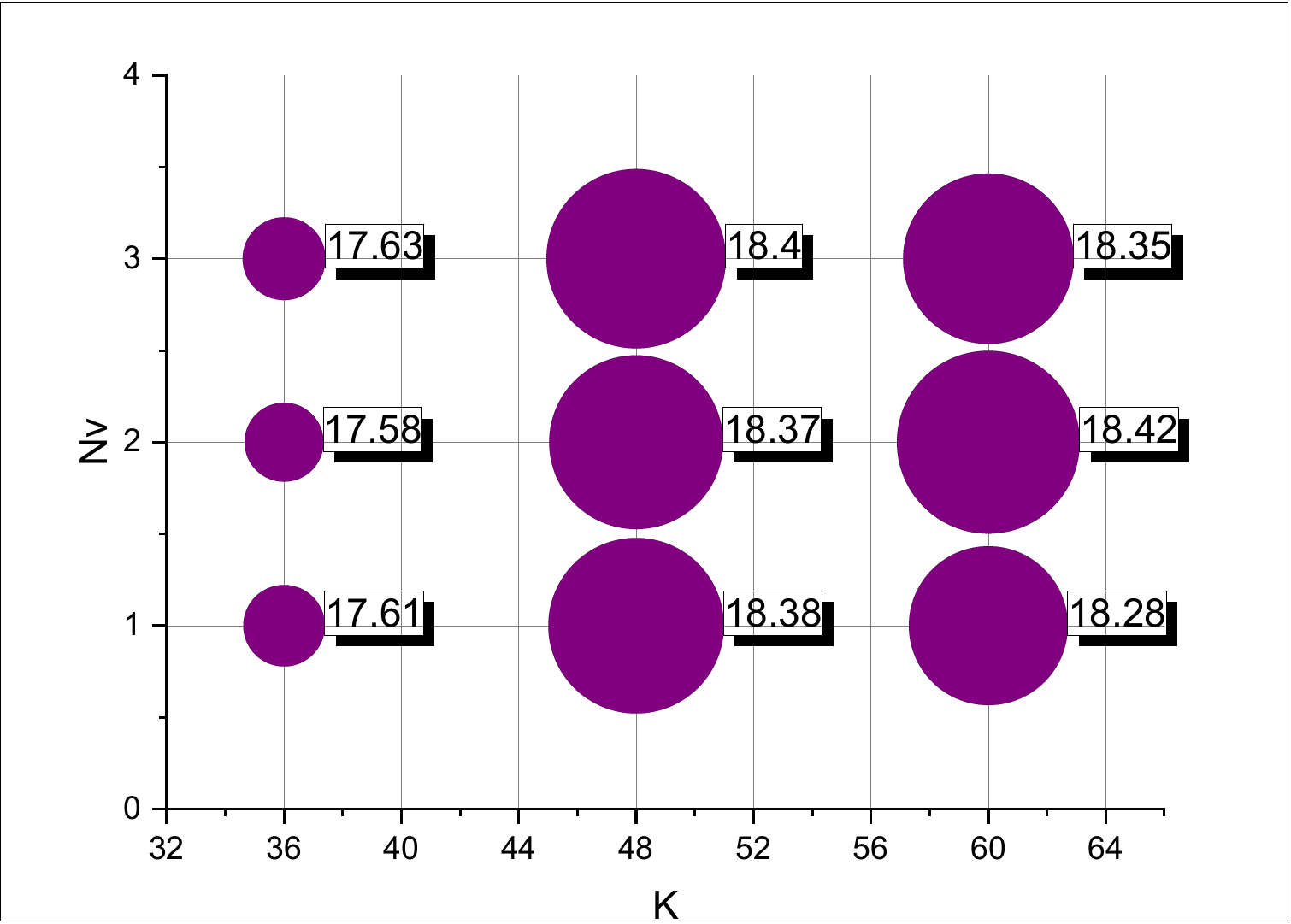}
		\caption{PSNR on the Carla-Centric dataset for models trained from scratch with varying hyperparameters.}
		\label{chaocan-carla}
\end{figure}

\begin{figure*}[t]
	\centering
	\includegraphics[width=0.85\linewidth]{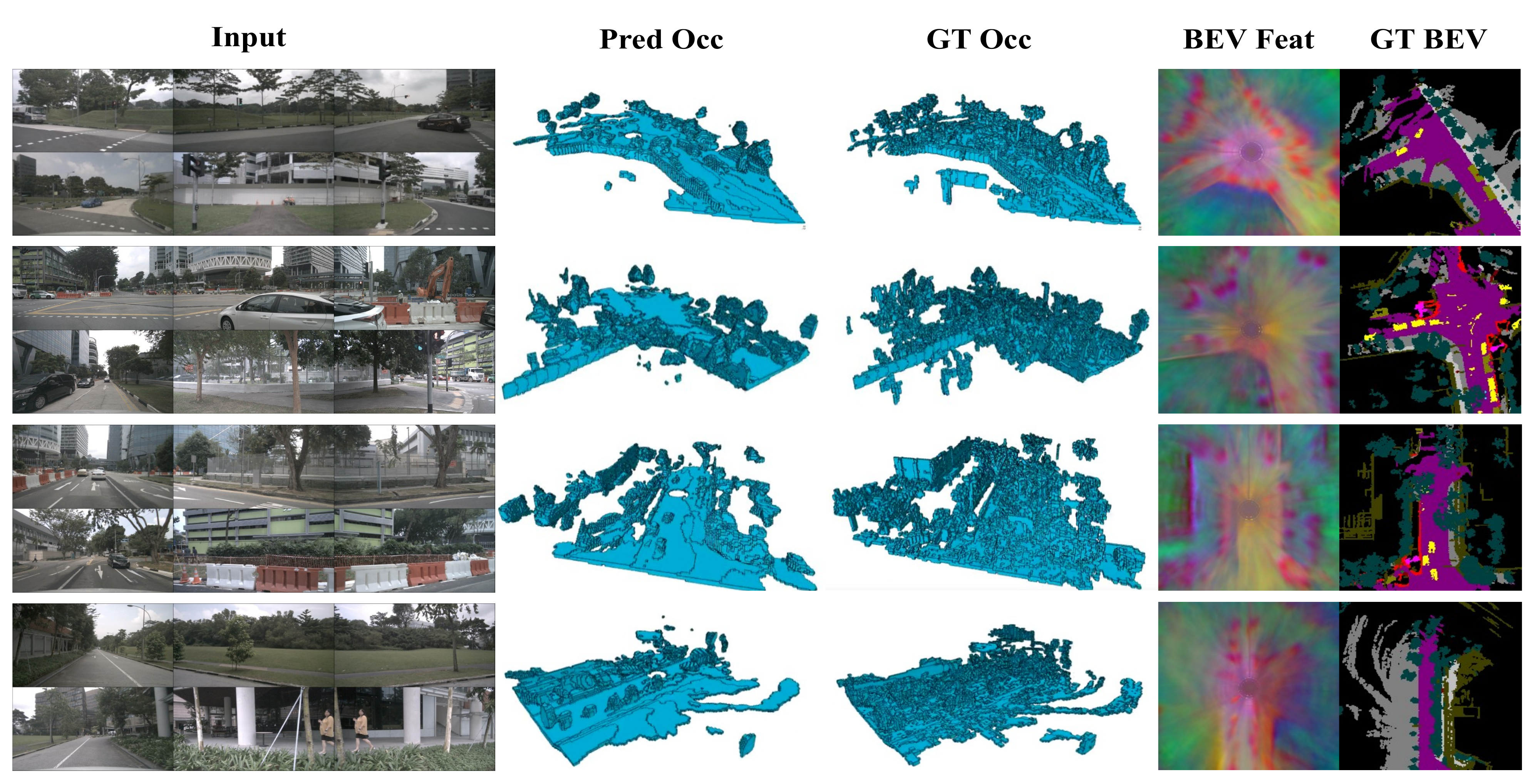}
	\caption{\textbf{Visualization of occupancy prediction results on the nuScenes dataset.} We present visualizations of the features from our BEV branch. In contrast to the Ground Truth, the predicted occupancy exhibits superior smoothness. We attribute this improvement to the strong generalization capability of our feedforward architecture, which enhances the model's robustness against inherent noise within the dataset. The BEV feature maps also demonstrate strong semantic discriminability, particularly for the ``tree" category.} 
	\label{occpred}
\end{figure*}

\begin{figure*}[t]
	\centering
	\includegraphics[width=0.85\linewidth]{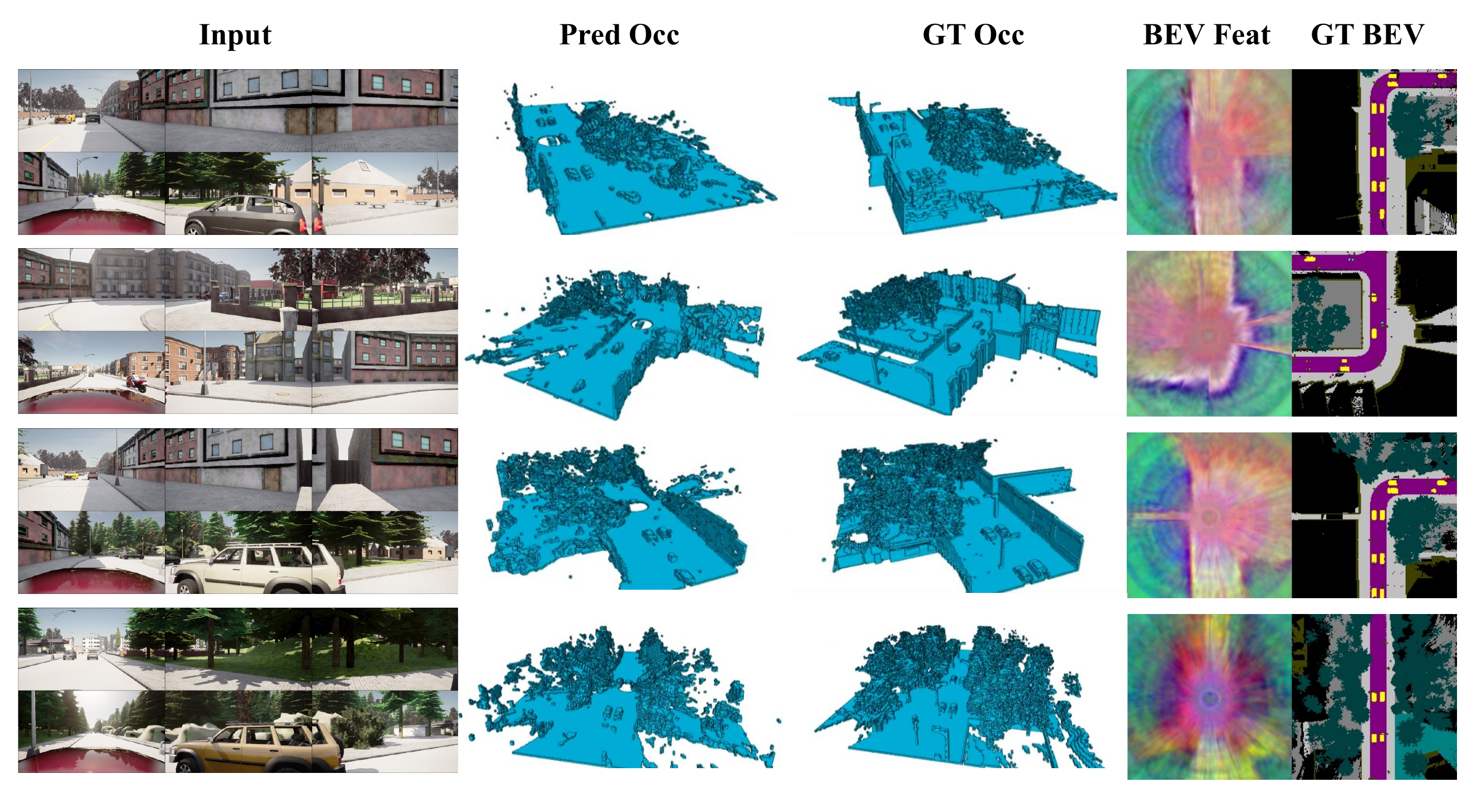}
	\caption{\textbf{Qualitative results on sample frames from the nuScenes dataset.} As demonstrated, the model produces high-fidelity predictions, where the BEV features exhibit clear semantic discriminability. This is evident in the distinct separation between classes such as ``wall" and ``drivable surface".} 
	\label{occpredcarla}
\end{figure*}

\subsection{Hyperparameters Searching}
We conduct an ablation study to analyze the impact of two critical hyperparameters on reconstruction quality: the number of CPFG feature groups, K, and the number of predicted Gaussians per voxel in the volume-aware branch, $G_v$. We evaluated combinations of K from {36, 48, 60} and $G_v$ from {1, 2, 3}, resulting in nine experimental configurations per dataset. 

As shown in \Cref{chaocan} and \Cref{chaocan-carla}, on the nuScenes dataset, the model achieves its best PSNR with the configuration ($K=48, G_v=3$). The PSNR is notably sensitive to the value of $G_v$; performance improves as $G_v$ increases. This is because models trained on nuScenes are challenged by high-frequency texture details, which cause the pixel branch to produce holes or voids. This, in turn, requires the volumetric representation branch to fill these voids as much as possible. A smaller $G_v$ value weakens this gap-filling capability, leading to lower performance.

In contrast, for the Carla-Centric dataset, the model demonstrates greater sensitivity to the value of $K$. This stems from its ego-inward setting, which demands high fidelity across all parts of the scene, unlike the limited forward-facing perspective. This makes the volumetric representation dominant. Consequently, a small $K$ leads to low field resolution and thus an inaccurate volumetric representation. Conversely, while a larger K might improve performance, it comes at the cost of significantly higher GPU memory consumption.

Taking into account the performance on both datasets, the frequency of peak performance occurrences in the ablation table, and the inherent trade-offs, we select $K=48, G_v=3$ as our optimal configuration.

\subsection{Visualization of the Prediction Results from the Occupancy-Aware Module}\Cref{occpred} visualizes the results from the occupancy-aware (occ-aware) branch of our model. This branch takes six-view images as input and outputs a classification score for each voxel in the 3D space. A voxel is classified as occupied if its occupancy score surpasses its ``air" score.

\begin{table*}[htbp]
    \centering
    \caption{Summary of camera configurations for common autonomous driving datasets.}
    \label{tab:camera_configs}
    \begin{tabularx}{0.85\textwidth}{l c c c}
        \toprule
        \textbf{Dataset} & \textbf{Number of cameras} & \textbf{Coverage} & \textbf{Horizontal Field of View} \\
        \midrule
        nuScenes    & 6 ring cameras           & $=$360°                    & 70°,70°,70°,110°,70°,70° \\
        Waymo       & 5 ring cameras            & $>$180°                & 50°,50°,50°,50°,50° \\
        Pandaset    & 6 ring cameras   & $=$360°                         & 50°,107°,107°,107°,107°,107°\\
        ONCE        & 6 ring cameras + 1 wide-angle camera           & $=$360°                         & 90°,90°,90°,90°,90°,90°\\
        Argoverse   & 7 ring cameras + 2 stereo cameras           & $=$360°                         & 69°,69°,69°,69°,69°,69°,69° \\
        \bottomrule
    \end{tabularx}
\end{table*}

\begin{figure}[t]
	\centering
	\includegraphics[width=\linewidth]{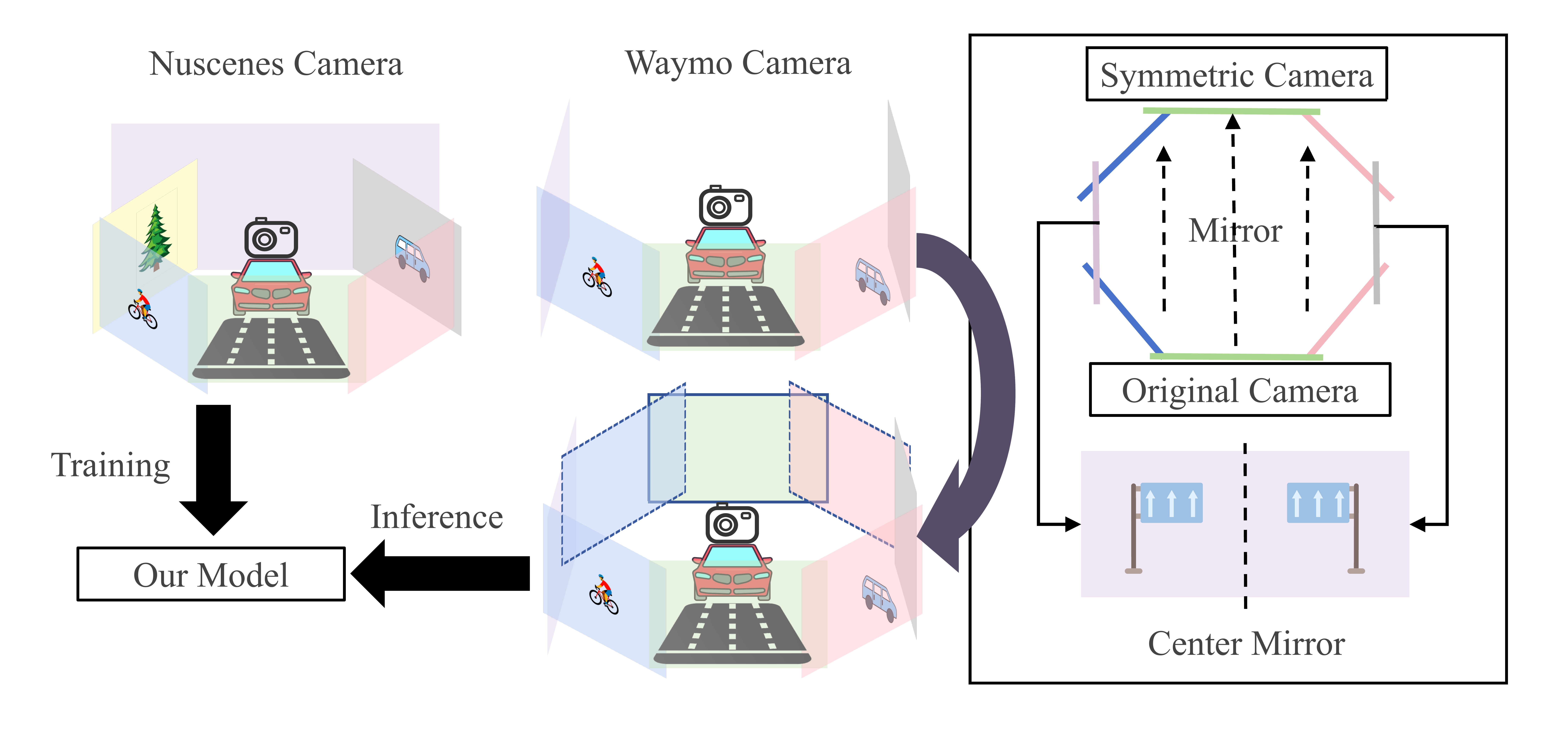}
	\caption{\textbf{Processing method for Waymo cameras.} Symmetric virtual cameras are constructed to complete unseen views and ensure pixel-level continuity. This process allows models trained on nuScenes to generalize directly to the Waymo dataset.} 
	\label{waymocam}
\end{figure}

Compared with the ground truth (GT) occupancy, the geometry predicted by our model is considerably smoother and more regular. This is attributed to the model's strong generalization capability, which effectively filters out the significant noise present in the GT labels. This effect is particularly evident in the geometry of ``trees". Due to the high uncertainty inherent in the complex structure of dense branches and leaves, the model tends to learn and predict a smooth, ``averaged" overall geometry, rather than a fine-grained structure with numerous internal voids. Furthermore, we also present the Bird's-Eye-View (BEV) feature map from this branch. As can be clearly observed, the feature response in the regions corresponding to trees is highly prominent and exhibits a consistent color. This indicates that our model has successfully learned a common and robust feature representation for the tree category.

 \label{morezeroshot}
\begin{table}[htbp]
    \centering
    \caption{UCCM Parameter Settings for Zero-shot Generalization on Other Autonomous Driving Datasets.}
    \label{tab:zeroshot_uccm}
    \begin{tabularx}{\linewidth}{c|c|c|c|c|c|c}
        \toprule
        & \multicolumn{6}{c}{\textbf{UCCM parameters}} \\
        \hline
        \textbf{Dataset} & $\rho_o$ & $\Delta h_o$ & $\rho_v$ & $\Delta h_v$ & $\rho_p$ & $\Delta h_p$ \\
        \hline
        nuScenes    & 0.90 & 0.00 & 0.98 & 0.40 & 0.98 & 0.40 \\
        Waymo       & 1.20 & 0.00 & 0.98 & 0.40 & 0.98 & 0.40 \\
        Pandaset    & 2.00 & 0.00 & 1.60 & 0.00 & 1.60 & 0.00 \\
        ONCE        & 1.80 & 0.00 & 1.00 & 0.50 & 1.00 & 0.50 \\
        Argoverse   & 0.09 & 0.00 & 0.98 & 0.00 & 0.98 & 0.00 \\
        \bottomrule
    \end{tabularx}
\end{table}

\Cref{occpredcarla} visualizes the occupancy-branch predictions trained on Carla-Centric synthetic data. Thanks to the absence of sensor noise in simulation, the occupancy labels are highly accurate, and the scenes are less complex than those in real-world nuScenes. Consequently, the model achieves superior predictive performance.

\subsection{More Visualization of Models Trained on Carla-Centric} 
\begin{figure*}[t]
	\centering
	\includegraphics[width=0.95\linewidth]{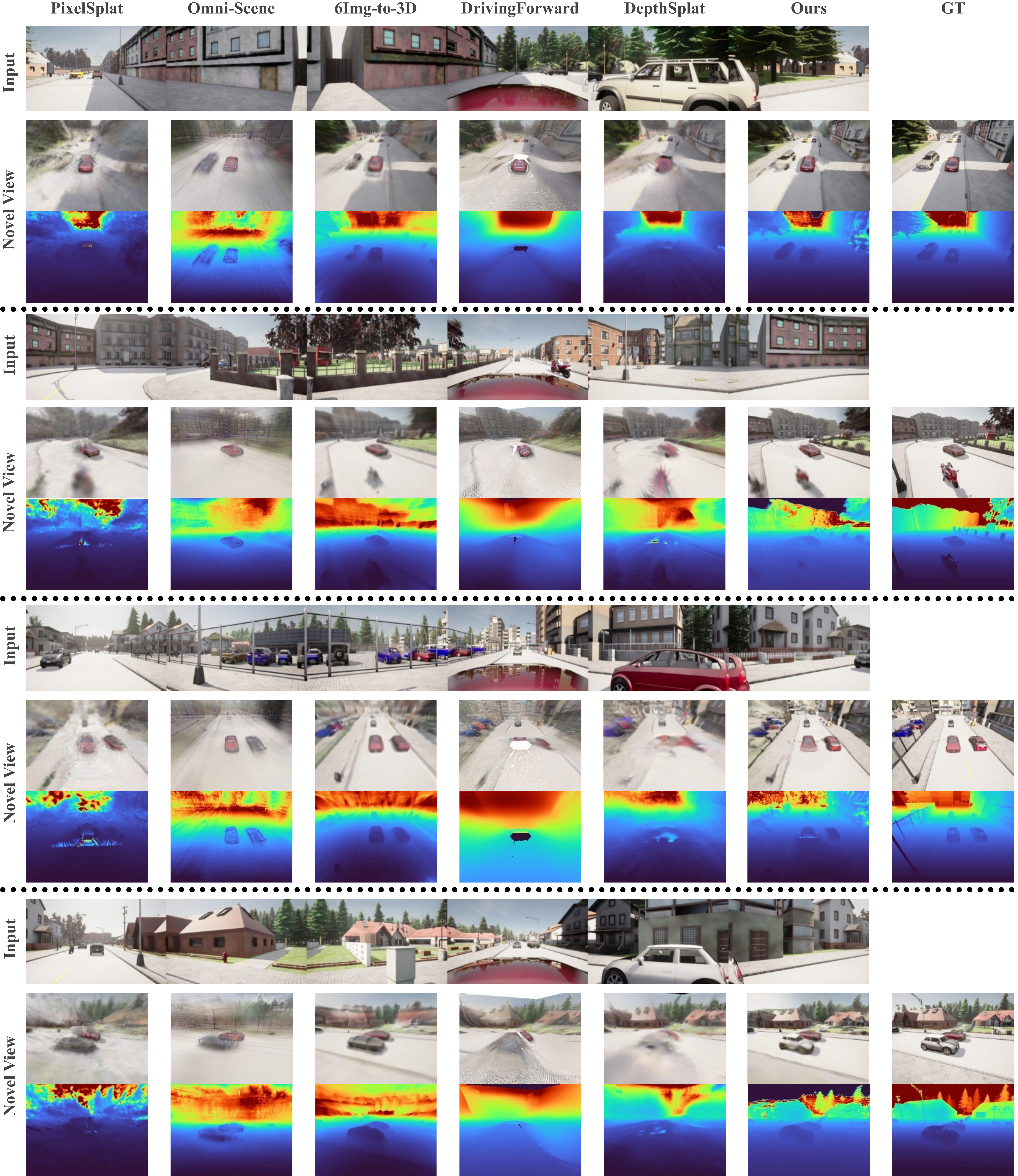}
	\caption{\textbf{Qualitative results on the Carla-Centric dataset.} Our model demonstrates superior performance in generating both photorealistic imagery and accurate geometry. It produces images with significantly sharper texture details and concurrently estimates highly accurate depth maps. Notably, our method effectively eliminates the blurring artifacts and voids (holes) that commonly plague other approaches, resulting in visually clean and structurally complete scene representations.} 
	\label{carlaappendix}
\end{figure*}

\Cref{carlaappendix} presents additional qualitative results on the Carla-centric dataset to further demonstrate the superiority of our method. Our model not only synthesizes novel-view images with richer high-frequency details but also produces depth maps that more accurately align with the ground truth. In stark contrast, the baseline methods suffer from noticeable blurring artifacts and significant voids (holes) in their rendered images and depth estimations.

\subsection{More Visualization of Models Trained on nuScenes} 
\begin{figure*}[t]
	\centering
	\includegraphics[width=\linewidth]{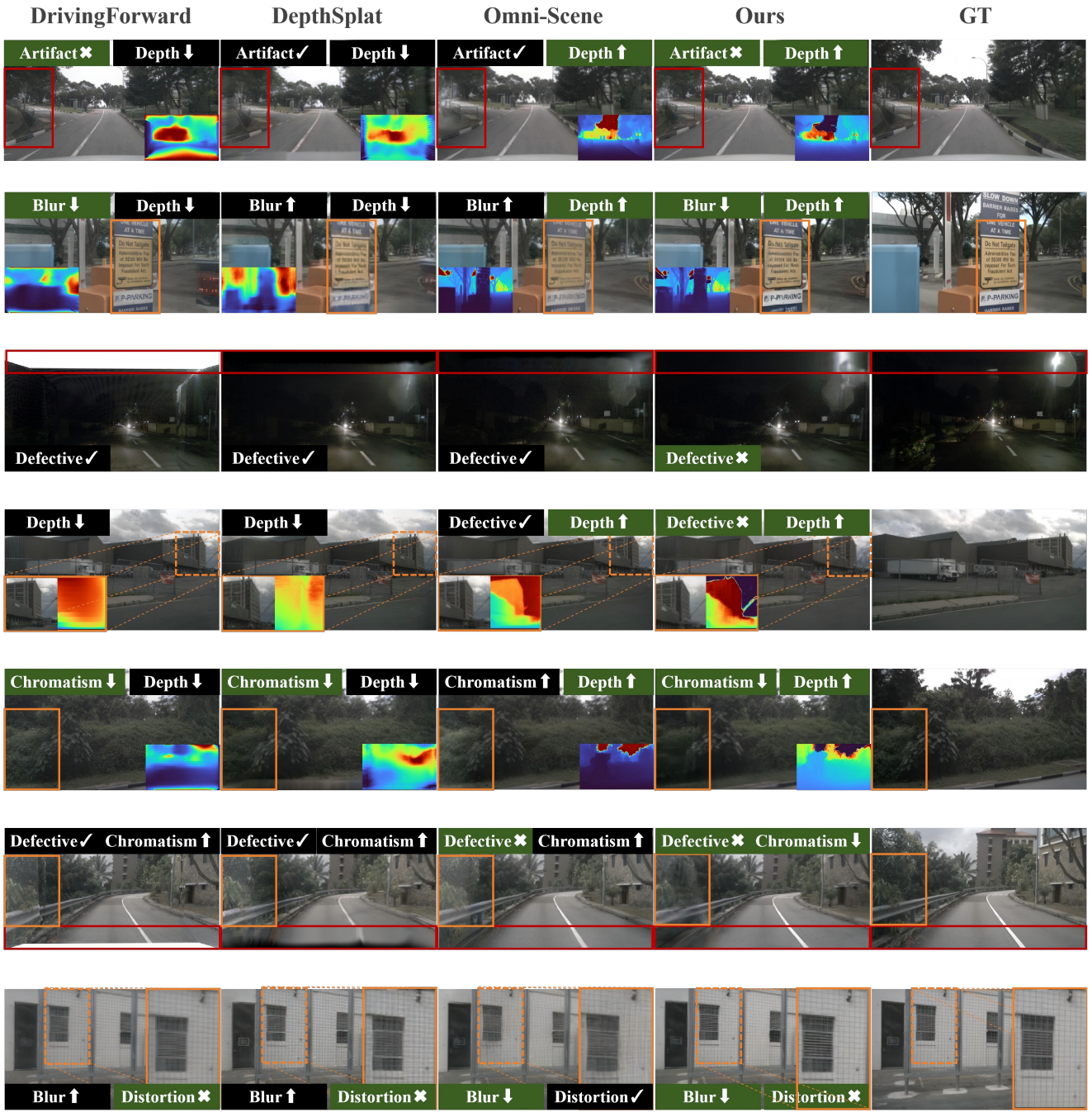}
	\caption{\textbf{The visualization results on the nuScenes dataset.} Our model achieves smaller photometric error and superior geometric representation. Particularly for high-frequency details like fences, our model renders fine textures without significant distortion. Furthermore, our model is less prone to artifacts and demonstrates better clarity on certain textures, such as text. Additionally, thanks to our sky-decoupled representation, we obtain a more complete geometry. In contrast, baseline methods tend to incorrectly learn distant, detailed objects into the sky's depth.} 
	\label{nuScenesmore}
\end{figure*}

As visualized in \Cref{nuScenesmore}, our model demonstrates a remarkable ability to reconstruct scenes with both high photometric realism and strong geometric integrity on the nuScenes dataset. In comparison, baseline methods fail on both fronts, suffering from a range of issues such as blurred textures, color shifts, distorted geometry, and unreliable depth. Our approach, conversely, excels in all these aspects, producing sharp textures, accurate colors, and a structurally sound 3D representation, as confirmed by its highly accurate depth maps.

\subsection{More Zero-shot Visualization} 
We use the model trained on nuScenes to test its zero-shot capability on the Waymo dataset. It is worth noting that nuScenes employs a 360-degree surround-view setup with a total of six cameras: five with a 70-degree Field of View (FOV) and one rear camera with a 110-degree FOV. All six cameras share the same resolution.
In contrast, the Waymo dataset has only five cameras, covering a scene of slightly more than 180 degrees, and there are resolution differences between the cameras. The Pandaset and ONCE datasets are similar to nuScenes, both featuring six cameras and a 360-degree view, but their camera intrinsics differ. The Argoverse dataset contains seven cameras, also covering a 360-degree scene. The camera configurations for these datasets are summarized in \Cref{tab:camera_configs}.

We conduct our experiments using only the images captured by the surround-view cameras and modify the relevant parameters of UCCM for each dataset, as shown in \Cref{tab:zeroshot_uccm}. Special handling is required for the Waymo dataset due to the severe lack of viewing angles.

\begin{figure*}[t]
	\centering
	\includegraphics[width=0.94\linewidth]{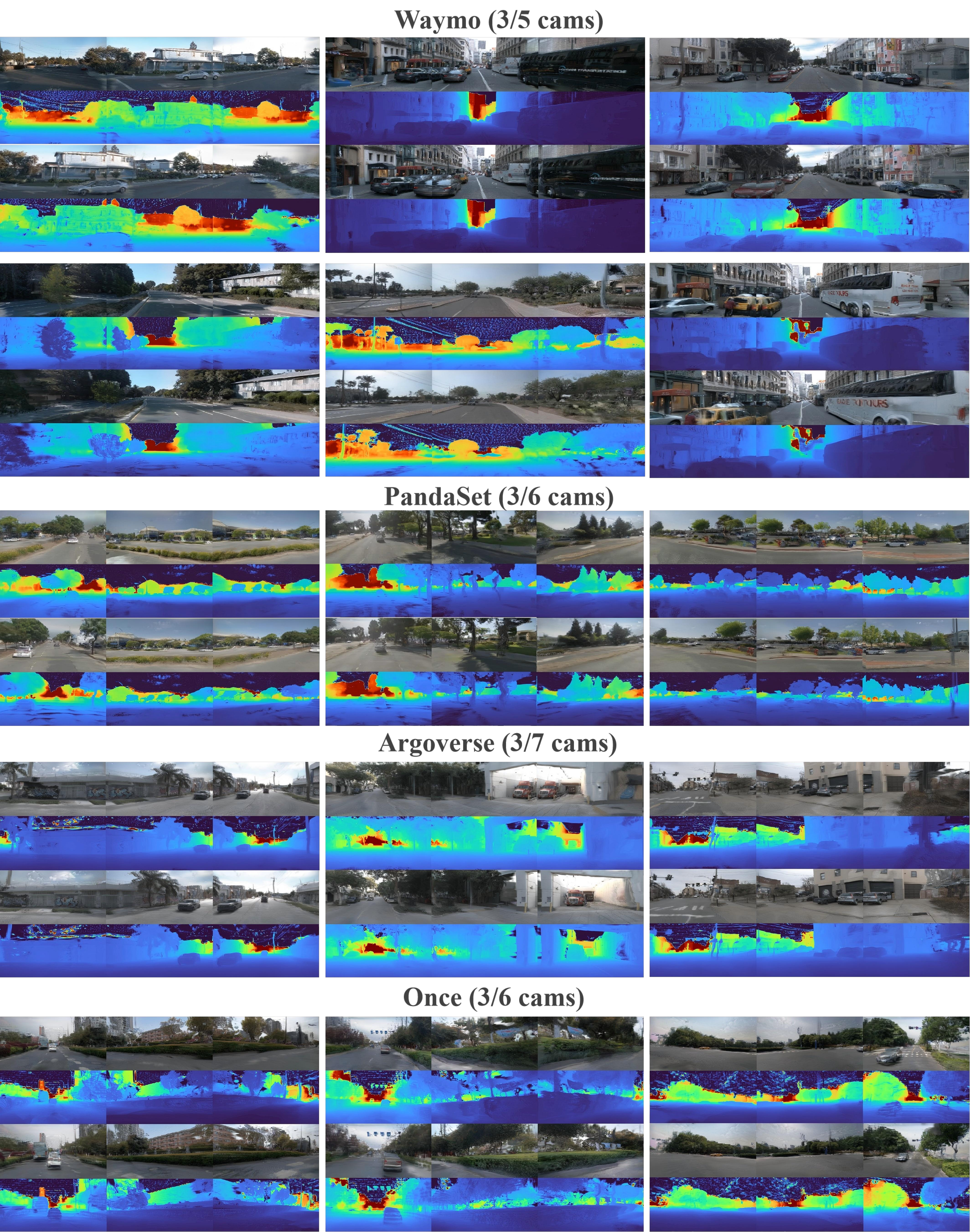}
	\caption{\textbf{More zero-shot visualization results.} The model trained on nuScenes generalizes well to other datasets.} 
	\label{zeroshotmore}
\end{figure*}

\begin{figure*}[t]
	\centering
	\includegraphics[width=\linewidth]{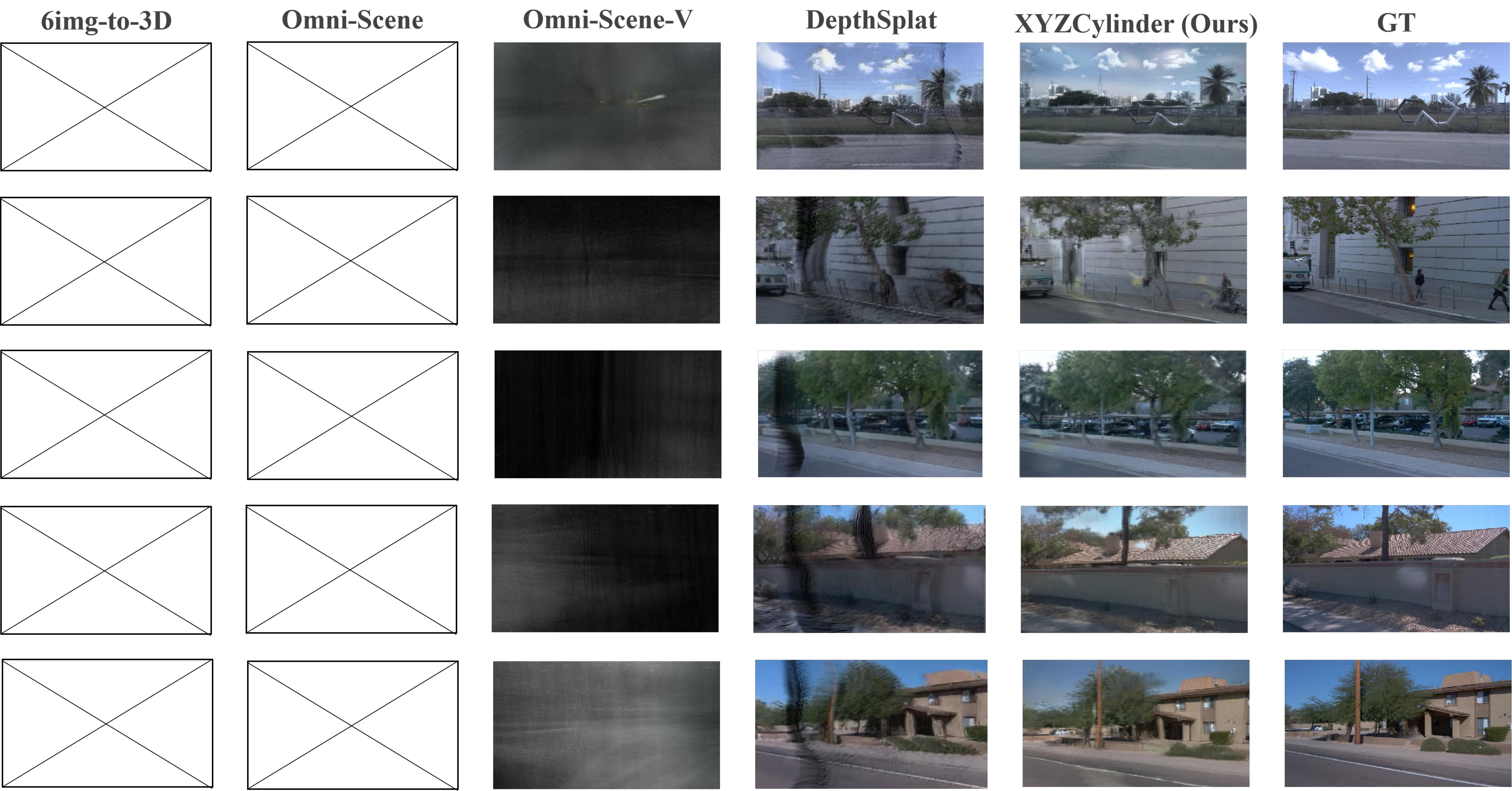}
	\caption{\textbf{A comparison of our model's performance against the baselines under a zero-shot setting.} Our model demonstrates a superior advantage in maintaining geometric accuracy. Due to their inherent model designs, 6img-to-3D and Omni-Scene cannot be directly transferred to driving datasets with a different number of cameras. Although the architecture of Omni-Scene-V, a variant of Omni-Scene, is aligned with the new dataset, it still fails to achieve effective compatibility.}
	\label{zeroshotcompare}
\end{figure*}

For the Waymo dataset, we adopt the processing method illustrated in \Cref{waymocam}. Specifically, we mirror the front, front-left, and front-right cameras across a plane to the rear of the vehicle, thereby constructing three symmetric virtual cameras. The images for these virtual cameras are mirrored versions of the forward-facing ones. For the side cameras, we horizontally flip their images along the vertical centerline to create virtual side cameras. This process results in an 8-camera configuration including the virtual cameras, whose collective views are seamlessly connected from left to right.
\Cref{zeroshotmore} and \Cref{zeroshotcompare} showcase the strong camera adaptability of our model with additional zero-shot results on the Waymo, Pandaset, and Argoverse datasets.

By design, the UCCM is inherently compatible with panoramic image inputs. We validated this capability using a selection of images sourced from Polyhaven and other panoramic imagery websites, with the results illustrated in the figure. For supplementary validation, we also captured images using game engines (Genshin Impact and Star Rail) and a smartphone camera, and subsequently stitched them into panoramic images using the PTGUI tool \cite{PTGui}.

\section{More Application Results} 
\subsection{Integrate with Generative Models}As a model designed for lifting 2D driving scenes to 3D, our approach exhibits a natural compatibility with 2D generative models. To demonstrate this, we have selected MagicDrive-V2 \citep{magicdrivev2}, a classic model in driving scene video generation, to enable the generation of 3D driving scenes from text prompts.

\begin{figure*}[htb]
	\centering
	\includegraphics[width=\linewidth]{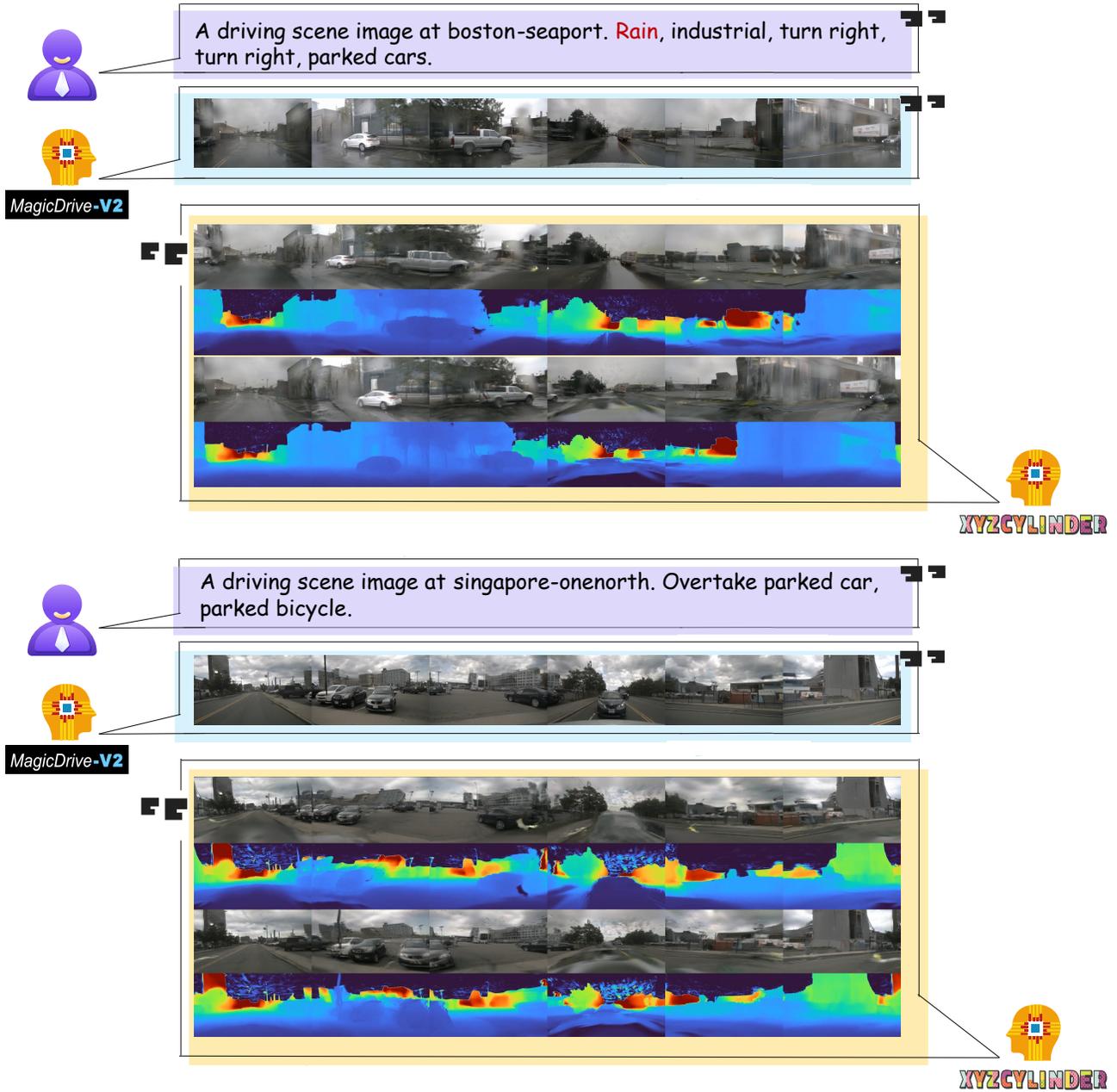}
	\caption{\textbf{Application cases 1:} Combining our model with advanced driving scene generators with text prompts.}
	\label{chatmagicdrivedit}
\end{figure*}

As illustrated in \Cref{chatmagicdrivedit}, our model can serve as a powerful interface for state-of-the-art 2D generative models. We demonstrate this by integrating it with MagicDrive-V2 for text-to-3D driving scene generation. In this workflow, our model is tasked with establishing a robust and geometrically accurate 3D scene structure, which MagicDrive-V2 then ``paints" with photorealistic textures guided by a text prompt. This synergy highlights that our method effectively bridges the gap from 2D generation to 3D-consistent scene creation, empowering existing models with strong 3D control.

\begin{figure*}[htb]
	\centering
	\includegraphics[width=0.9\linewidth]{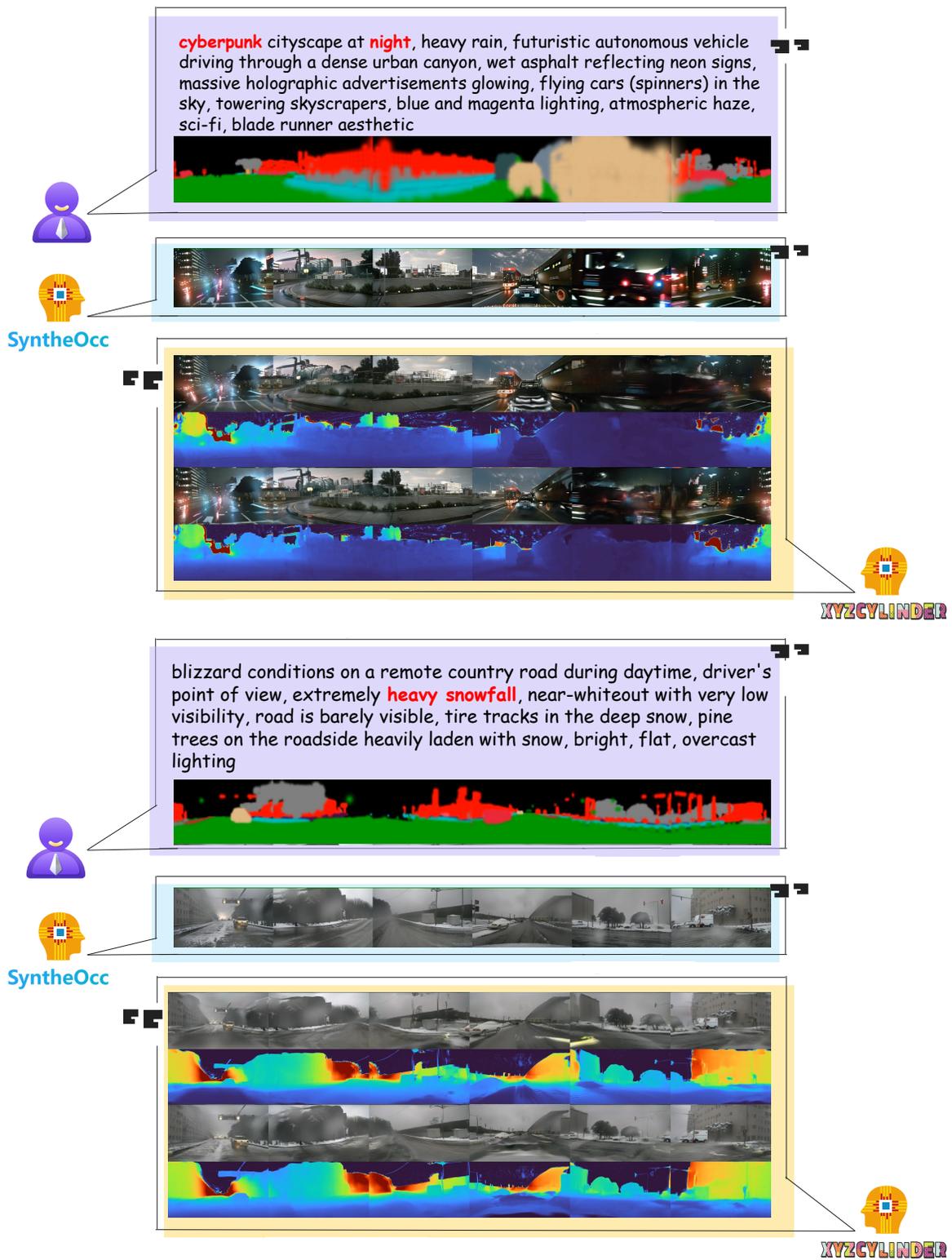}
	\caption{\textbf{Application cases 2:} Combining our model with advanced driving scene generators with text prompts and occupancy conditions.}
	\label{chatsyntheocc}
\end{figure*}

\begin{figure*}[htb]
	\centering
	\includegraphics[width=0.84\linewidth]{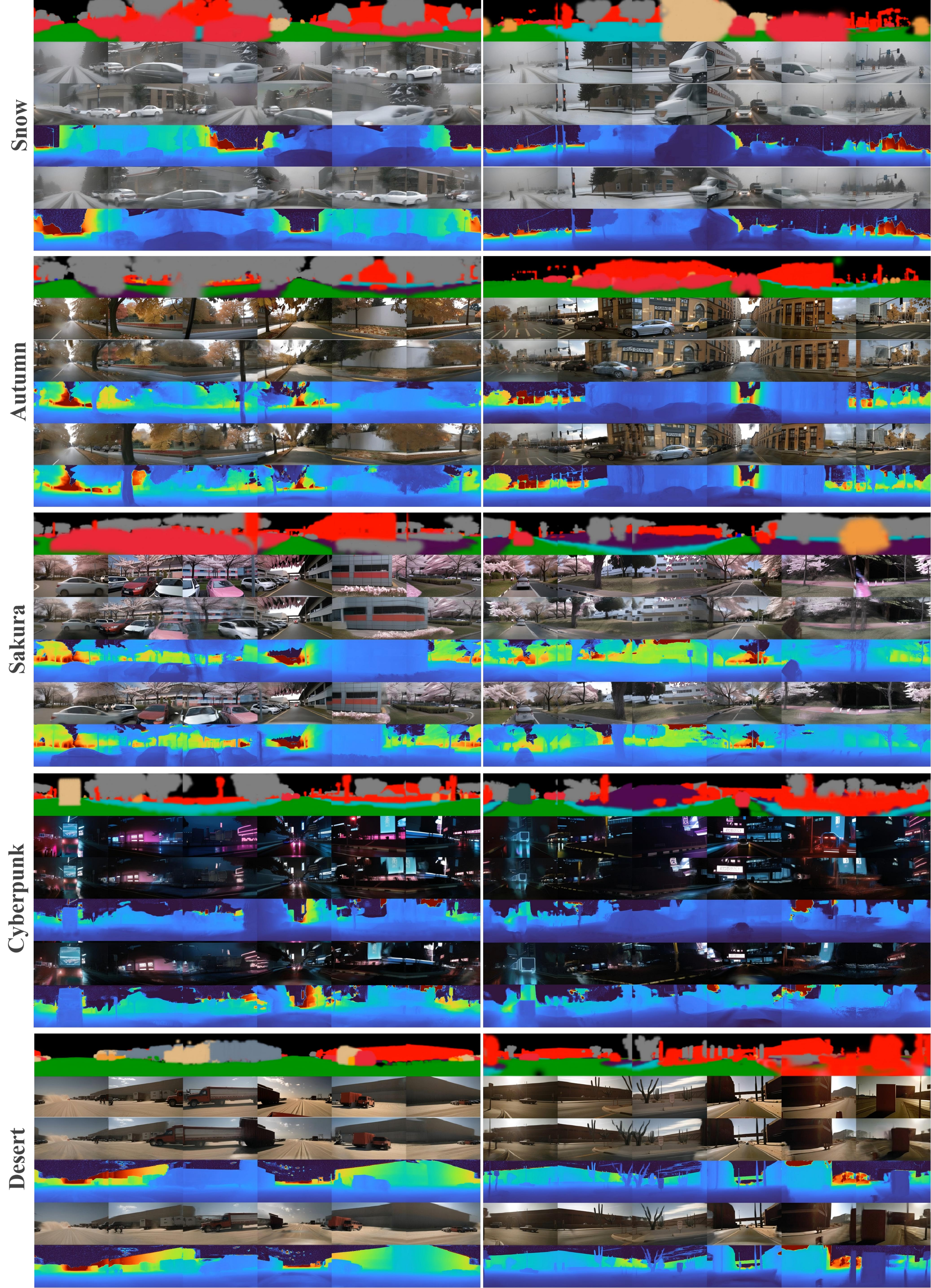}
	\caption{\textbf{Generalization Across Diverse Texture Domains.} We fine-tune SyntheOcc on SDXL to ensure superior style control.} 
	\label{texturedomain}
\end{figure*}

To further enhance the controllability of our generation process, we demonstrate that our model can be seamlessly integrated with layout-based generative models. We chose SyntheOcc \citep{syntheocc}, a voxel-conditioned scene generator, and fine-tuned our model to incorporate its Gaussian-rendered semantic occupancy as an additional control signal. This hybrid approach enables a powerful decoupling of control: users can define the high-level artistic style and global attributes via a text prompt, while simultaneously dictating the precise spatial layout of scene elements (e.g., vehicles, roads) through the semantic map. As shown in \Cref{chatsyntheocc} and \Cref{texturedomain}, this dual-guidance mechanism yields highly controllable results that adhere to both textual and structural constraints.

\clearpage
\newpage

\clearpage
\newpage

\newpage

\begin{figure*}[htb]
	\centering
	\includegraphics[width=0.8\linewidth]{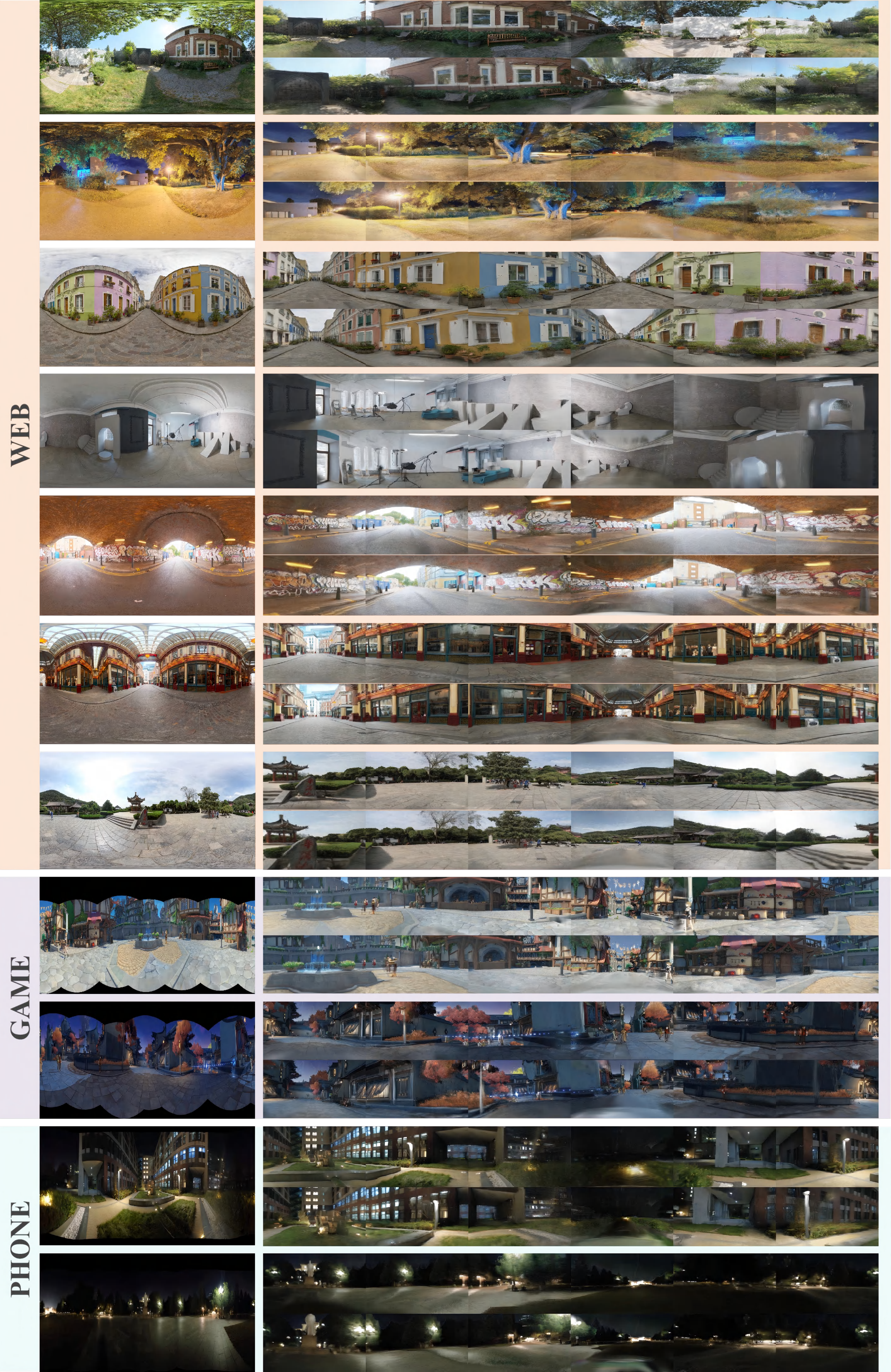}
	\caption{\textbf{Visualization results with panoramic images as input.}}
	\label{pano}
\end{figure*}

% \newpage

\end{document}